\documentclass[5p]{elsarticle}
\usepackage{amsmath,amsfonts,amssymb}
\usepackage{bm}
\usepackage{algorithm}
\usepackage{algorithmic}
\usepackage{url}
\usepackage[breaklinks=true,letterpaper=true,colorlinks,bookmarks=false]{hyperref}
\usepackage[table]{xcolor}
\usepackage{tabularx}
\usepackage{style}
\usepackage{todonotes}
\usepackage{paralist}
\usepackage[inline]{enumitem}
\usepackage{textcomp}
\usepackage{lineno}
\modulolinenumbers[5]
\usepackage{rotating}
\usepackage{hyphenat}
\usepackage{multirow}
\usepackage{xspace}
\usepackage{placeins}
\usepackage{multirow}
\usepackage{makecell}
\usepackage{ragged2e}
\usepackage{newfloat}
\usepackage{multicol}

\definecolor{deep_blue}{rgb}{0,.2,.5}
\definecolor{dark_blue}{rgb}{0,.15,.5}

\hypersetup{pdftex,  
  breaklinks=true,  
  colorlinks=true,
  linkcolor=dark_blue,
  citecolor=deep_blue,
  }

\newcolumntype{Y}{>{\centering\arraybackslash}X}

\newcommand{\B}[1]{\mathbf{#1}}
\def\trans{^\mathsf{T}}
\def\slantfrac#1#2{\kern.1em^{#1}\kern-.1em/\kern-.1em_{#2}}

\newcommand{\hairsp}{\hspace{1pt}}
\newcommand{\ie}{\textit{i.\hairsp{}e.}\xspace}
\newcommand{\eg}{\textit{e.\hairsp{}g.}\xspace}

\makeatletter

\newcommand\autorefs[1]{\@first@ref#1,@}
\def\@throw@dot#1.#2@{#1}
\def\@set@refname#1{
    \edef\@tmp{\getrefbykeydefault{#1}{anchor}{}}%
    \def\@refname{\@nameuse{\expandafter\@throw@dot\@tmp.@autorefname}s}%
}
\def\@first@ref#1,#2{%
  \ifx#2@\autoref{#1}\let\@nextref\@gobble
  \else%
    \@set@refname{#1}
    \@refname~\ref{#1}
    \let\@nextref\@next@ref
  \fi%
  \@nextref#2%
}
\def\@next@ref#1,#2{%
   \ifx#2@ and~\ref{#1}\let\@nextref\@gobble
   \else, \ref{#1}
   \fi%
   \@nextref#2%
}

\makeatother

\journal{NeuroImage}

\begin{document}

\begin{frontmatter}

\title{Deriving reproducible biomarkers from 
    multi-site resting-state data: An Autism-based example}


\newif\ifanonimize

\anonimizefalse

\ifanonimize

\author[parietaladdress, ceaaddress]{Xxxxxxxxx \textsc{Xxxxxxxxx}\corref{correspondingauthor}}
\cortext[correspondingauthor]{Corresponding author}
\ead{xxxxxxxxx.xxxxxxxxx@xxxxxxxxxx.xxx}
\author[parietaladdress, ceaaddress]{Xxxxxxxxx \textsc{Xxxxxxxxxx}}
\author[parietaladdress, ceaaddress]{Xxxxxxxxx \textsc{Xxxxxxxxxx}}
\author[stonybrookaddress, centraleaddress]{Xxxxxxxxx \textsc{Xxxxxxxxxx}}
\author[parietaladdress, ceaaddress]{Xxxxxxxxxxx \textsc{Xxxxxxxxxx}}

\address[parietaladdress]{Xxxxxxx Xxxxxxx Xxxxxxx Xxxxxxx Xxxxxxx}
\address[ceaaddress]{Xxxxxxx Xxxxxxx Xxxxxxx Xxxxxxx Xxxxxxx}
\address[stonybrookaddress]{Xxxxxxx Xxxxxxx Xxxxxxx Xxxxxxx Xxxxxxx}
\address[centraleaddress]{Xxxxxxx Xxxxxxx Xxxxxxx Xxxxxxx Xxxxxxx}

\else

\author[parietaladdress,ceaaddress]{Alexandre \textsc{Abraham}\corref{correspondingauthor}}
\cortext[correspondingauthor]{Corresponding author}
\ead{abraham.alexandre@gmail.com}
\author[cmiaddress,klineaddress]{Michael \textsc{Milham}}
\author[nyuaddress]{Adriana \textsc{Di Martino}}
\author[cmiaddress,klineaddress]{R. Cameron \textsc{Craddock}}
\author[stonybrookaddress,centraleaddress]{Dimitris \textsc{Samaras}}
\author[parietaladdress,ceaaddress]{Bertrand \textsc{Thirion}}
\author[parietaladdress,ceaaddress]{Gael \textsc{Varoquaux}}

\address[parietaladdress]{Parietal Team, INRIA Saclay-\^{I}le-de-France, Saclay, France}
\address[ceaaddress]{CEA, DSV, I\textsuperscript{2}BM, Neurospin b\^{a}t 145, 91191 Gif-Sur-Yvette, France}
\address[stonybrookaddress]{Stony Brook University, NY 11794, USA}
\address[centraleaddress]{Ecole Centrale, 92290 Ch\^atenay Malabry, France}
\address[cmiaddress]{Child Mind Institute, New York, USA}
\address[klineaddress]{Nathan S. Kline Institute for Psychiatric Research, Orangeburg, New York, USA}
\address[nyuaddress]{NYU Langone Medical Center, New York, USA}

\fi

\sloppy

\begin{abstract}

Resting-state functional Magnetic Resonance Imaging (R-fMRI) holds the promise
to reveal functional biomarkers of neuropsychiatric disorders.  However,
extracting such biomarkers is challenging for complex multi-faceted 
neuropathologies, such as autism spectrum disorders. Large multi-site datasets
increase sample sizes to compensate for this complexity, at the cost of
uncontrolled heterogeneity.
This heterogeneity raises new challenges, akin to those
face in realistic diagnostic applications.
Here, we demonstrate the feasibility of inter-site
classification of neuropsychiatric status, with an application to the Autism
Brain Imaging Data Exchange (ABIDE) database, a large (N=871) multi-site
autism dataset.
For this purpose, we investigate pipelines that
extract the most predictive biomarkers from the data.
These R-fMRI pipelines build participant-specific connectomes from
functionally-defined brain areas.
Connectomes are then compared across participants to learn patterns of
connectivity that differentiate typical controls from individuals with autism. 
We predict this neuropsychiatric status for participants from the same
acquisition sites or different, unseen, ones. Good choices of methods for
the various steps of the pipeline lead to 67\% prediction accuracy on the
full ABIDE data, which is significantly better than previously reported results.
We perform extensive validation on multiple subsets of the data defined 
by different inclusion criteria.  These enables detailed
analysis of the factors contributing to successful connectome-based
prediction. First, prediction accuracy improves as we include more
subjects, up to the maximum amount of subjects available. Second, the 
definition of functional brain areas is of paramount importance for biomarker
discovery: brain areas extracted from large R-fMRI datasets outperform
reference atlases in the classification tasks. 

\end{abstract}


\end{frontmatter}


\section{Introduction}

In psychiatry, as in other fields of medicine, both the standardized observation
of signs, as well as the symptom profile are critical for diagnosis. However,
compared to other fields of medicine, psychiatry lacks accompanying objective
markers that could lead to more refined diagnoses and targeted treatment
\cite{kapur2012}. Advances in non-invasive brain
imaging techniques and analyses (\eg \cite{craddock2013,vanessen2012})
are showing great promise for uncovering patterns of
brain structure and function that can be used as objective measures of mental
illness.
Such \emph{neurophenotypes} are important for clinical applications such as disease
staging, determination of risk prognosis, prediction and monitoring of
treatment response, and aid towards diagnosis (\eg \cite{castellanos2013}).

Among the many imaging techniques available, resting-state fMRI (R-fMRI) is a
promising candidate to define functional neurophenotypes
\cite{kelly2008,vanessen2012}.
In particular, it is non-invasive and, unlike conventional
task-based fMRI, it does not require a constrained experimental setup
nor the active and focused participation of the subject. It has been 
proven to capture interactions between brain regions that may lead
to neuropathology diagnostic biomarkers \cite{greicius2008b}.
Numerous studies have linked variations in brain
functional architecture measured from R-fMRI to behavioral traits and mental
health conditions  such as Alzheimer disease (\eg \cite{greicius2004,chen2011},
Schizophrenia (\eg \cite{garrity2007,zhou2007,jafri2008,calhoun2011}), ADHD,
autism (\eg \cite{plitt2014}) and others (\eg \cite{anderson2013}).
Extending these findings, predictive modeling approaches have
revealed patterns of brain functional connectivity that could serve as
biomarkers for classifying depression (\eg \cite{craddock2009}), ADHD
(\eg \cite{adhd2012}), autism (\eg \cite{anderson2011}),
and even age \cite{dosenbach2010}. This growing number of
studies has shown the feasibility of using R-fMRI to identify biomarkers.
However questions about the readiness of R-fMRI to detect clinically useful biomarkers
remain \cite{plitt2014}. In particular, the reproducibility and
generalizability of these approaches in research or clinical settings are
debatable. Given the modest sample size of most R-fMRI studies, the effect of
cross-study differences in data acquisition, image processing, and sampling
strategies \cite{desmond2002,murphy2004,thirion2007} has not been quantified.

Using larger datasets is commonly cited as a solution to
challenges in reproducibility and statistical power \cite{button2013}. They are
considered a prerequisite to 
R-fMRI-based classifiers for the detection of psychiatric illness.
Recent efforts have accelerated the generation
of large databases through sharing and aggregating independent data
samples\cite{fair2012,mennes2013,dimartino2014}.
However, a number of concerns must be addressed before
accepting the utility of this approach. Most notably, the many potential sources
of uncontrolled variation that can exist across studies and sites, which
range from MRI acquisition protocols (\eg scanner type, imaging sequence, see
\cite{friedman2008}), to
participant instructions (\eg eyes open vs. closed, see \cite{yan2013}),
to recruitment strategies
(age-group, IQ-range, level of impairment, treatment history and acceptable
comorbidities). Such variation in aggregate samples is often viewed as
dissuasive, as its effect on diagnosis and biomarker extraction is unknown.
It commonly motivates researchers to limit the number of sites
included in their analyses at the cost of sample size. 

Cross-validated results obtained from predictive models are more
robust to inhomogeneities: they measure model generalizability by applying it
to unseen data, \ie, data not used to train the model. In particular,
leave-out cross-validation strategies, which remove single individuals (or
random subsets), are common in biomarkers studies. However, these strategies
do not measure the effect of potential site-specific confounds. In the
present study we leverage aggregated R-fMRI samples to address this problem.
Instead of leaving out random subsamples as test sets, we left out entire
sites to measure performance in the presence of uncontrolled variability.

Beyond challenges due to inter-site data heterogeneity, choices in the
functional-connectivity data-processing pipeline further add to the
variability of results \cite{carp2012,yan2013,shirer2015}. While
preprocessing procedures are now standard, the different steps of the
prediction pipeline vary from one study to another. These entail
specifying regions of interest, extracting regional time courses,
computing connectivity between regions, and
identifying connections that relate to subject's phenotypes 
\cite{craddock2009,richiardi2010,shirer2012,eickhoff2015}.

Lack of ground truth for brain functional architecture
undermines the validation of R-fMRI data-processing pipelines.
The use of functional connectivity for individual prediction suggests
a natural figure of merit: prediction accuracy.
We contribute quantitative evaluations, to help 
settling down on a parameter-free pipeline for R-fMRI. 
Using efficient implementations, we were
able to evaluate many pipeline options and select the best
method to estimate atlases, extract connectivity matrices, and predict
phenotypes. 

To demonstrate that pipelines to extract R-fMRI neuro-phenotypes can
reliably learn inter-site biomarkers of
psychiatric status on inhomogeneous data, we analyzed R-fMRI
in the Autism Brain Imaging Data Exchange (ABIDE) \cite{dimartino2014}.
It compiles a dataset of 1112 R-fMRI participants by gathering data from 17
different sites. After preprocessing, we selected 871
to meet quality criteria for MRI and phenotypic information.
Our inter-site prediction methodology reproduced conditions found under most
clinical settings, by leaving
out whole sites and using them as newly seen test sets.
To validate the robustness of our approach, we performed nested cross-validation and
varied samples per inclusion criteria (\eg sex, age).
Finally, to assess losses in predictive power
associated with using a heterogeneous aggregate dataset instead of 
uniformly defined samples, we included a comparison of intra- and inter-site
prediction strategies. 

\section{Material and methods}

\begin{figure*}[tb]
    \includegraphics[width=\linewidth]{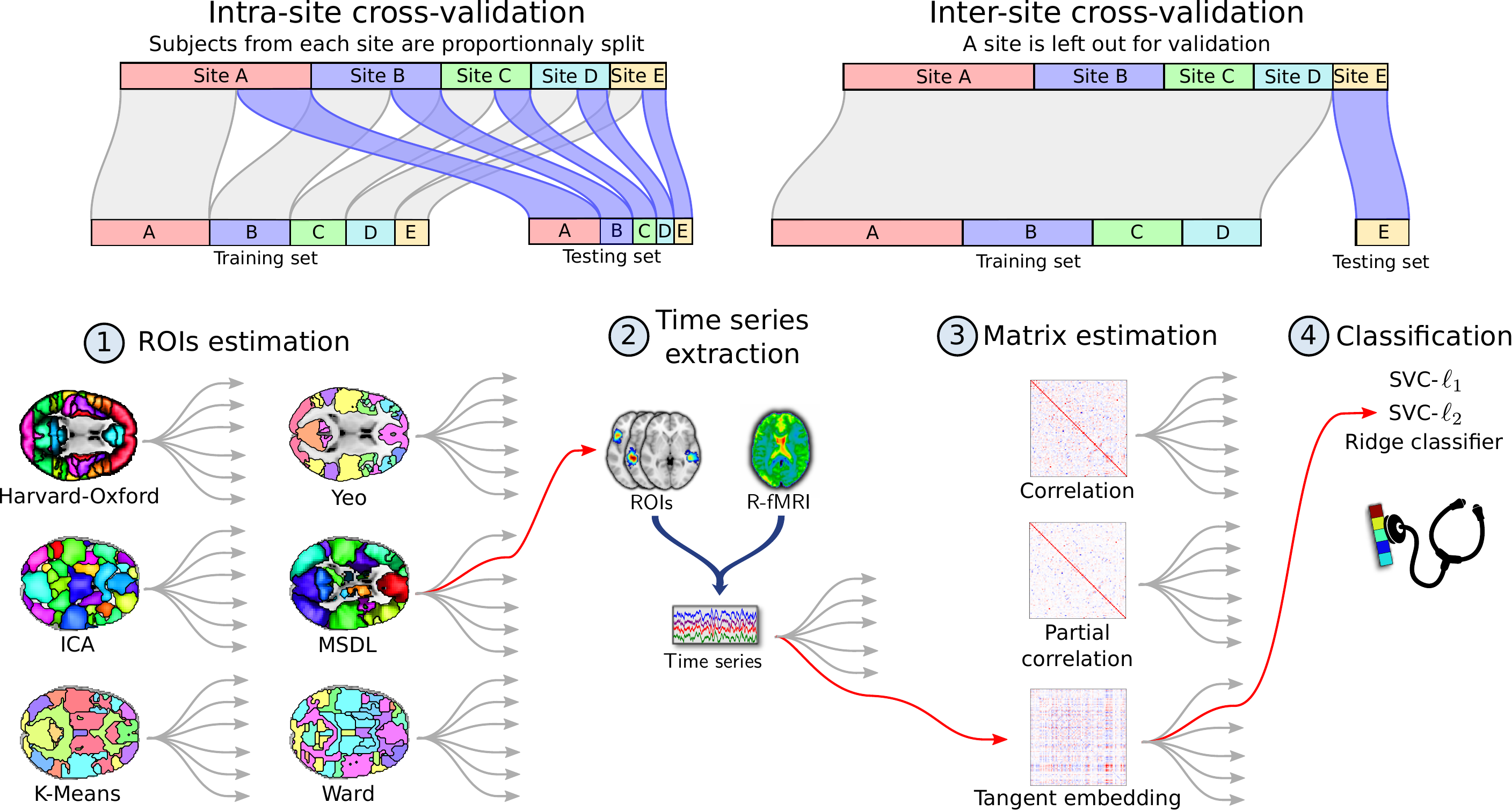}%
    \caption{\textbf{Functional MRI analysis pipeline.} Cross-validation
        schemes used to validate the pipeline are presented above.
        \textbf{Intra-site} cross-validation consists of randomly splitting
        the participants into training and testing sets while preserving the
        ratio of samples for each site and condition.
        \textbf{Inter-site} cross-validation consists of leaving out
        participants from an entire site as testing set. In the first step of
        the pipeline, regions of interest are estimated from the training set.
        The second step consists of extracting signals of interest from all
        the participants, which are turned into connectivity features via
        covariance estimation at the third step. These features are used in
        the fourth step to perform a supervised learning task and yield an
        accuracy score. An example of pipeline is highlighted in red. This
        pipeline is the one that gives best results for inter-site prediction.
        Each model is decribed in the section relative to material and
        methods.
    \label{fig:pipeline}}
\end{figure*}

A connectome is a functional-connectivity matrix between a set brain
regions of interest (ROIs). We call such a set of ROIs an atlas, even
though some of the methods we consider extract the regions from the data
rather than relying on a reference atlas (see \autoref{fig:pipeline}).
We investigate here pipelines that discriminate individuals based on the
connection strength of this connectome \cite{varoquaux2013}, 
with a classifier on the edge weights \cite{craddock2009}.

Specifically, we consider connectome-extraction pipelines composed of four steps:
\begin{enumerate*}[label=\arabic*)]
    \item{region estimation,}
    \item{time series extraction,}
    \item{matrix estimation, and}
    \item{classification}
\end{enumerate*}
--see \autoref{fig:pipeline}.
We investigate different options for the various steps:
brain parcellation method, connectivity matrix estimation, and
final classifier. 

In order to measure the impact of uncontrolled variability on prediction
performance, we designed a diagnosis task where the model classifies participants
coming from an acquisition site not seen during training. 

\subsection{Datasets: Preprocessing and Quality Assurance (QA)}

We ran our analyses on the ABIDE repository \cite{dimartino2014}, a sample
aggregated across 17 independent sites.
Since there was no prior coordination between sites,
the scan and diagnostic/assessment protocols
vary across sites \footnote{See
\url{http://fcon_1000.projects.nitrc.org/indi/abide/} for specific information}.
To easily replicate and extend our work,
we rely on a publicly available preprocessed version of this
dataset provided by the Preprocessed Connectome Project
\footnote{\url{http://preprocessed-connectomes-project.github.io/abide}}
initiative.
We specifically used
the data processed with the Configurable Pipeline for the Analysis of
Connectomes \cite{craddock2013b} (C-PAC). Data were selected based on the results of quality visual inspection by
three human experts who inspected for largely incomplete brain coverage,
high movement peaks, ghosting and other scanner artifacts.
This yielded 871 subjects out of the initial 1112.

Preprocessing included slice-timing correction, image
realignment to correct for motion, and intensity normalization.
Nuisance regression was performed to remove signal fluctuations induced by head motion,
respiration, cardiac pulsation, and scanner drift \cite{lund2005,fox2005}.
Head motion was modeled using 24 regressors derived from the
parameters estimated during motion realignment \cite{friston1995}, scanner drift
was modeled using a quadratic and linear term, and physiological noise was
modeled using the 5 principal components with highest variance from a decomposition of white
matter and CSF voxel time series (CompCor) \cite{behzadi2007}.
After regression of nuisance signals, fMRI  was coregistered on the anatomical
image with FSL BBreg, and the results where normalized to MNI space with the
non-linear registration from ANTS \cite{avants2009}.
Following time series extraction, data
were detrended and standardized (dividing by the standard deviation in each
voxel).

\subsection{Step 1: Region definition}

To reduce the dimensionality of
the estimated connectomes, and to improve the interpretability of results,
connectome nodes were defined from regions of interest (ROIs) rather than
single voxels. ROIs can be either selected from a priori atlases or directly
estimated from the correlation structure of the data being analyzed. Several
choices for both approaches were used and compared to evaluate the impact of
ROI definition on prediction accuracy. Note that the ROI were all
defined on the training set to avoid possible overfitting.

We considered the following predefined atlases:
\textbf{Harvard Oxford} \cite{desikan2006}, a structural atlas computed
from 36 individuals' T1 images, \textbf{Yeo} \cite{yeo2011}, a functional atlas
composed of 17 networks derived from clustering \cite{lashkari2010b} of
functional connectivity on a thousand individuals, and \textbf{Craddock}
\cite{craddock2012}, a multiscale functional atlas computed using constrained
spectral clustering, to study the impact of the number of regions. 

We derived data-driven atlases based on four strategies. We explored two
clustering methods: \textbf{K-Means}, a technique to cluster fMRI time series
\cite{goutte1999,thirion2014}, which is a top-down approach
minimizing cluster-level variance; and \textbf{Ward's clustering}, which also minimizes
a variance criterion using hierarchical agglomeration. Ward's algorithm admits
spatial constraints at no cost and has been extensively used to learn brain
parcellations \cite{thirion2014}. We also explored two
decomposition methods: \textbf{Independent component analysis (ICA)} and
\textbf{multi-subject dictionary learning (MSDL)}.  ICA is a widely-used method
for extracting brain
maps from R-fMRI \cite{calhoun2001a, beckmann2004} that maximizes the statistical
independence between spatial maps. We specifically used the ICA implementation
of \cite{varoquaux2010}. MSDL extracts a group atlas based on
dictionary learning and multi-subject modeling, and employs spatial
penalization to promote contiguous regions \cite{abraham2013}. Unlike
MSDL and Ward clustering, ICA and K-Means do not take into account the spatial
structure of the features and are not able to enforce a particular
spatial constraint on their components. We indirectly provide such structure by applying a
Gaussian smoothing on the data \cite{thirion2014} with a full-width half
maximum varying from 4mm to 12mm.
Differences in the number of ROIs in the various
atlases present a potential confound for our comparisons. As such,
connectomes were constructed using only the 84 largest ROIs from each atlas,
following \cite{abraham2013}\footnote{For Harvard-Oxford and Yeo reference
atlases, selecting 84 ROIs yields slightly better results than using the full
version as shown in \autoref{fig:regions_vs_full}. This is probably due to the fact
that these additional regions are smaller and less important regions, hence they
induce spurious correlations in the connectivity matrices.}.

\subsection{Step 2: Time-series extraction}

The common procedure to extract one representative time series per ROI is to
take the average of the voxel signals in each region (for non overlapping ROIs)
or the weighted average of non-overlapping regions
(for fuzzy maps).
For example, ICA and MSDL produce overlapping fuzzy maps. Each
voxel's intensity reflects the level of confidence that this voxel belongs to a
specific region in the atlas. 
In this case, we found that extracting time courses using a spatial regression
that models each volume of R-fMRI data as a mixture of spatial patterns gives
better results 
(details in \ref{app:time_series} and \autoref{fig:ts_extraction}).
For non-overlapping ROIs, this procedure is equivalent to weighted
averaging on the ROI maps.

After this extraction, we remove at the region level
the same nuisance regressors as at the voxel
level in the preprocessing.
Signals summarizing high-variance voxels (CompCor \cite{behzadi2007}) are
regressed out: the 5 first principal
components of the 2\% voxels with highest variance.
In addition, we extract the signals of ROIs corresponding to physiological or
acquisition noise\footnote{We identify these ROIs using an approach similar to
    FSL FIX (FMRIB's ICA-based Xnoiseifier). FIX extracts spatial and temporal descriptors from matrix
decomposition results and uses a classifier trained on manually labeled
components to determine if they correspond to noise.}
and regress them out to reject non-neural information \cite{varoquaux2013}.
Finally, we also regress out primary and secondary head motion artifacts 
using 24-regressors derived from the parameters estimated during motion
realignment \cite{friston1995}.

\subsection{Step 3: Connectivity matrix estimation}

Since scans in the ABIDE dataset do not include enough R-fMRI time
points to reliably estimate the true covariance matrix for a given individual,
we used the Ledoit-Wolf shrinkage estimator, a parameter-free regularized
covariance estimator \cite{ledoit2004}, to estimate relationships between
ROI time series (details are given in \ref{app:cov}).
For pipeline step 3, we studied three different connectivity
measures to capture interactions between brain regions:
\begin{enumerate*}[label=\textit{\roman*)}]
  \item the correlation itself,
  \item partial correlations from the inverse covariance matrix
        \cite{varoquaux2013,smith2011}, and
  \item the tangent embedding parameterization of the covariance matrix
        \cite{varoquaux2010,ng2014}.
\end{enumerate*}
We obtained one connectivity weight per pair of regions for
each subject.
At the group level, we ran a nuisance regression across the connectivity
coefficients to remove the effects of site, age and gender.

\subsection{Step 4: Supervised learning}

As ABIDE represents a post-hoc aggregation of data from several different sites,
the assessments used to measure autism severity vary somewhat
across sites. As a consequence, autism severity scores are not directly 
quantitatively comparable between sites.
However, the diagnosis between ASD and  Typical Control (TC) is more reliable and has
already been used in the context of classification
\cite{nielsen2013, plitt2014, chen2015}.

Discriminating between ASD and TC individuals is a
supervised-learning task.
We use the connectivity measures between all pairs of regions, extracted in the
previous step, as features to train a classifier across individuals to
discriminate ASD patients from TC. 

We explore different machine-learning methods\footnote{For the sake of
simplicity, we report only the 3 main methods here. However, we also
investigated random forests and
Gaussian naive Bayes, that gave low prediction performance.
Feature selection also led to worse results, both
with univariate ANOVA screening and with sparse models such as Lasso
or SVC-$\ell_1$ --see Figure~\ref{fig:with_or_wo_fs} for more details.
Logistic regression gave results similar to SVC. 
}. 
We first rely on the most commonly used approach, the $\ell_2$-penalized support
vector classification \textbf{SVC}. Given that we expect only a few connections
to be important for diagnosis, we also include the $\ell_1$-penalized
sparse \textbf{SVC}. Finally, we include \textbf{ridge regression}, which also uses an
$\ell_2$ penality but is faster than the \textbf{SVC}. For all models
we use the implementation of the scikit-learn library
\cite{pedregosa2011}\footnote{Software versions: python 2.7, scikit-learn 0.17.1, scipy
0.14.0, numpy 1.9.0, nilearn 0.1.5.}.
We chose to work only with linear models for their interpretability. In
particular, in \autoref{subsec:connections} we inspect classifier weights
to identify which connections are the most
important for prediction. 

\subsection{Exploring dataset heterogeneity}

Previous studies of prediction using the ABIDE sample
\cite{plitt2014,chen2015} have included less than 25\% of the
available data, most likely to limit heterogeneity -- not only due to
differences in imaging, but also to factors such as age, sex, and handedness.
We explored the effect of such choice on the results, by examining data
subsets of different heterogeneity (see \autoref{tab:subsets}). These included: subsample \#1
- \textbf{all participants} (871 individuals, 727 males and 144 females, 6 to 64 years old) is the full set of individuals that
passed QA, subsample \#2 - \textbf{largest sites} (736 individuals, 613 males
and 123 females) 6 sites with
less than 30 participants are removed from subsample \#1, subsample \#3 - \textbf{right
handed males} (639 individuals) consists of the right-handed males from
subsample \#1, subsample \#4 - \textbf{right handed males, 9-18 years} (420 individuals) is the
restriction of subsample \#3 to participants between 9 and 18 years old;
subsample \#5 - \textbf{right handed males, 9-18 yo, 3 sites} (226 individuals) are the
individuals from subsample \#4 belonging to the three largest sites (see
\autoref{tab:full_subsets} for more details)\footnote{\autoref{tab:subject_ids}
    lists subjects of each subsample. The lists of subjects for each cross-validation fold
    are available at \url{https://team.inria.fr/parietal/files/2016/04/cv_abide.zip}}.

\begin{table}[tb]
    \footnotesize%
    \rowcolors{2}{gray!15}{white}%
    \begin{tabularx}{\linewidth}{p{.001cm}p{2.3cm}|ccX}
        \multicolumn{2}{l}{Subsample} & Card. & Sites & Selection criteria \\
        \hline
        1 &All subjects & 871   & 16    & All subjects after QC \\
        2 & Biggest sites & 736   & 11    & Remove 6 smallest sites \\
        3 & Right handed males & 639   & 14    & All subjects without left handed and women \\
        4 & Right handed males, 9-18 yo & 420   & 14    & Right handed males between 9 and 18 years old \\
        5 & Right handed males, 9-18 yo, 3 sites & 226   & 3     & Young right
        handed males from 3 biggest sites \\
\end{tabularx}\\
\caption{\textbf{Subsets of ABIDE used in evaluation.} We explore several
subsets of ABIDE with different homogeneity. \emph{Card.} stands for the
cardinality of the dataset, \ie the number of subjects. More details about
these subsets are presented in the appendices --\autoref{tab:full_subsets}.}
\label{tab:subsets}
\end{table}

\subsection{Experiments: prediction on heterogeneous data}

ABIDE datasets come from 17 international sites with no prior coordination,
which is a typical clinical application setting.
In this setting, we want to:
\begin{enumerate*}[label=\textit{\roman*)}]
    \item measure the performance of different prediction pipelines,
    \item use these measures to find the best options for each pipeline step
        (\autoref{fig:pipeline}) and,
    \item extract ASD biomarkers from the best pipeline.
\end{enumerate*}

\paragraph{Cross-validation}
In order to measure the prediction accuracy of a pipeline, we use
10-fold cross-validation by training it exclusively on training data
(including atlas estimation) and predicting on the testing set.
Any hyper-parameter of the various methods used inside a pipeline is set 
internally in a nested cross-validation.

We used two validation approaches.
\textbf{Inter-site prediction} addresses the challenges associated with
aggregate datasets by using whole sites as testing sets. This scenario also
closely simulates the real world clinical challenge of not being able to sample
data from every possible site. Note that this cross-validation can only be
applied on a dataset with at least 10 acquisition sites, in order to leave out a different
site in each fold. We do not apply it on subsample \#5 that has been restricted
to 3 sites to reduce site-specific variability.
\textbf{Intra-site prediction} builds training and testing sets are as
homogeneous as possible, reducing site-related variability. It is based on
stratified shuffle split cross-validation, which splits participants into
training and test sets while preserving the ratio of samples for each site and
condition. We used 80\% of the dataset for training and the remaining for
testing.

\paragraph{Learning curves}

A learning curve measures the impact of the number of training
samples on the performance of a predictive model.
For a given prediction task, the learning curve is constructed by
varying the number of samples in the training set with a constant
test set.
Typically, an increasing curve
indicates that the model
can still gain additional useful information from more data.
It is expected that the performance of a model will plateau at some point.

\paragraph{Summarizing prediction scores for a pipeline choice}

To quantify the impact of the different options on the prediction accuracy,
we select representative prediction scores for each model. 
Indeed, for ICA, K-Means, and MSDL, we explore several values for each parameters
and thus obtain a range of prediction values. 
For these methods,
we use the 10\% best-performing pipelines for post-hoc analysis
\footnote{Note that the criterion used to select
the data-point used for the post-hoc analysis, \ie the 10\% best
performing, is independent from the factors studied in this analysis,
as we select the 10\% best performing for each value of these factors.
Hence, the danger of inflating effects by a selection \cite{vul2009} does not
apply to the post hoc analysis.}.
We do not rely only on the
top-performing pipelines, as they may result from overfitting, nor the
complete set of pipelines as some pipelines yield bad scores because of bad
parametrization and may artificially lower the performance of these methods.

\paragraph{Impact of pipeline choices on prediction accuracy}

In order to determine the best pipeline option at each step, we want to measure
the impact of these choices on the prediction scores.
In a full-factorial ANOVA, we then fit a linear model to explain prediction
accuracy from the choice of pipeline steps, modeled as categorical
variables. 
We report the corresponding effect size and its 95\% confidence interval. 

\paragraph{Pairwise comparison of pipeline choice}

Two options can also be compared by comparing the scores between pairs of
pipelines that differ only by that option using a Wilcoxon test that
does not rely on Gaussian assumptions.

\paragraph{Extracting biomarkers}
\label{par:biomarkers}
For a given pipeline, the final classifier yields a biomarker
to diagnose between HC and ASD. To characterize this biomarker, we measure $p$-values on the
classifier weights.
We obtain the null distribution of the weights by permuting the prediction
labels.

\subsection{Computational aspects}

For data-driven brain atlas models (\eg ICA), region-extraction methods
entail feature learning on up to 300 GB of
data, too big to fit in memory. Systematic study requires many
cross-validations and subsamples and thus careful computational optimizations.
Hence, for K-Means and ICA, we reduced the length of the time series by applying PCA
dimensionality reduction with efficient randomized SVD \cite{martinsson2011}.
MSDL uses algorithmic optimizations \cite{abraham2013}
to process a large number of subjects in a reasonable time
\footnote{Despite the optimizations, the whole study presented here is
very computationally costly. Indeed, the nested cross-validation, the 
various pipeline options, and the various subsets, would lead to a computation
time of 10
years on a computer with 8 cores and 32GB RAM. We used a computer grid to 
reduce it to two months.} (2 hours
for 871 subjects). However, it is still the most costly method because of the
number of runs required to find the optimal value for its 3 parameters.

For reproducibility, we rely on the publicly
available implementations of
the scikit-learn package \cite{pedregosa2011} for all the clustering algorithms,
connectivity matrix estimators and predictors.

\section{Results}

Here we present the most notable trends emerging from our analyses.
More details are provided in supplementary materials.
First, we compared inter- and intra-site prediction of
ASD diagnostic labels while varying the number of subjects in the training
set.
Second, in a post-hoc analysis, we identified the pipeline
choices most relevant for prediction and proposed a good choice of
pipeline steps for prediction.
Finally, by analyzing the weights of the classifiers, we highlighted functional
connections that best discriminate between ASD and TC.

\subsection{Overall prediction results}

\begin{figure}[tb]
    \includegraphics[width=\linewidth]{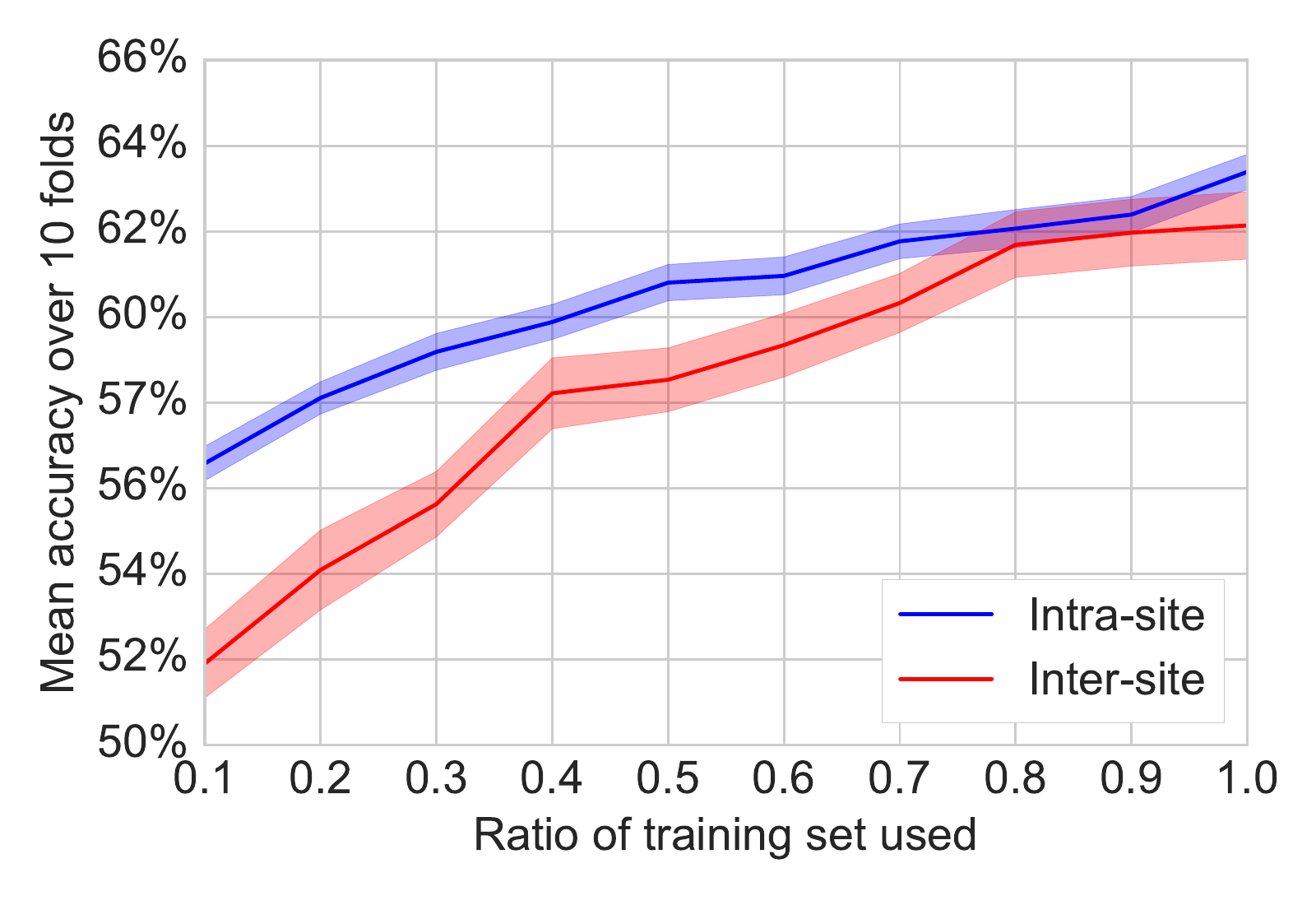}%
    \caption{\textbf{Learning curve.}
Classification results obtained by varying the ratio of the
training set using to train the classifier while keeping a fixed testing set.
The colored bands represent the standard error of the prediction. A score
increasing with the number of subjects, \ie a positive slope indicates that the
addition of new subjects improves performance. This curve is an average of the
results obtained on several subsamples. Detailed results per subsamples are presented in
the appendices
--\autoref{fig:full_learning_curve}}
    \label{fig:learning_curve}
\end{figure}

For inter-site prediction, the highest accuracy obtained on the whole dataset is
$66.8\%$ (see \autoref{tab:scores}) which exceeds previously published ABIDE
findings \cite{nielsen2013,plitt2014,chen2015} and chance at
53.7\%\footnote{Chance is calculated by taking the highest score obtained by a
dummy classifier that predicts random labels following the distribution of the
two classes.}.  Importantly, performance
increases steadily with training set size, for all subsamples
(\autorefs{fig:learning_curve,fig:full_learning_curve}).  This increase implies
that optimal classification performance is not reached even for the largest
dataset tested.  

The main difference between intra-site and inter-site settings is that the
variability of inter-site prediction is higher (\autoref{tab:scores}). However, this difference
disappears with
a large enough training set (\autoref{fig:learning_curve}).
This is encouraging for clinical applications.

\begin{figure}[tb]
    \centerline{%
    \includegraphics[width=.9\linewidth]{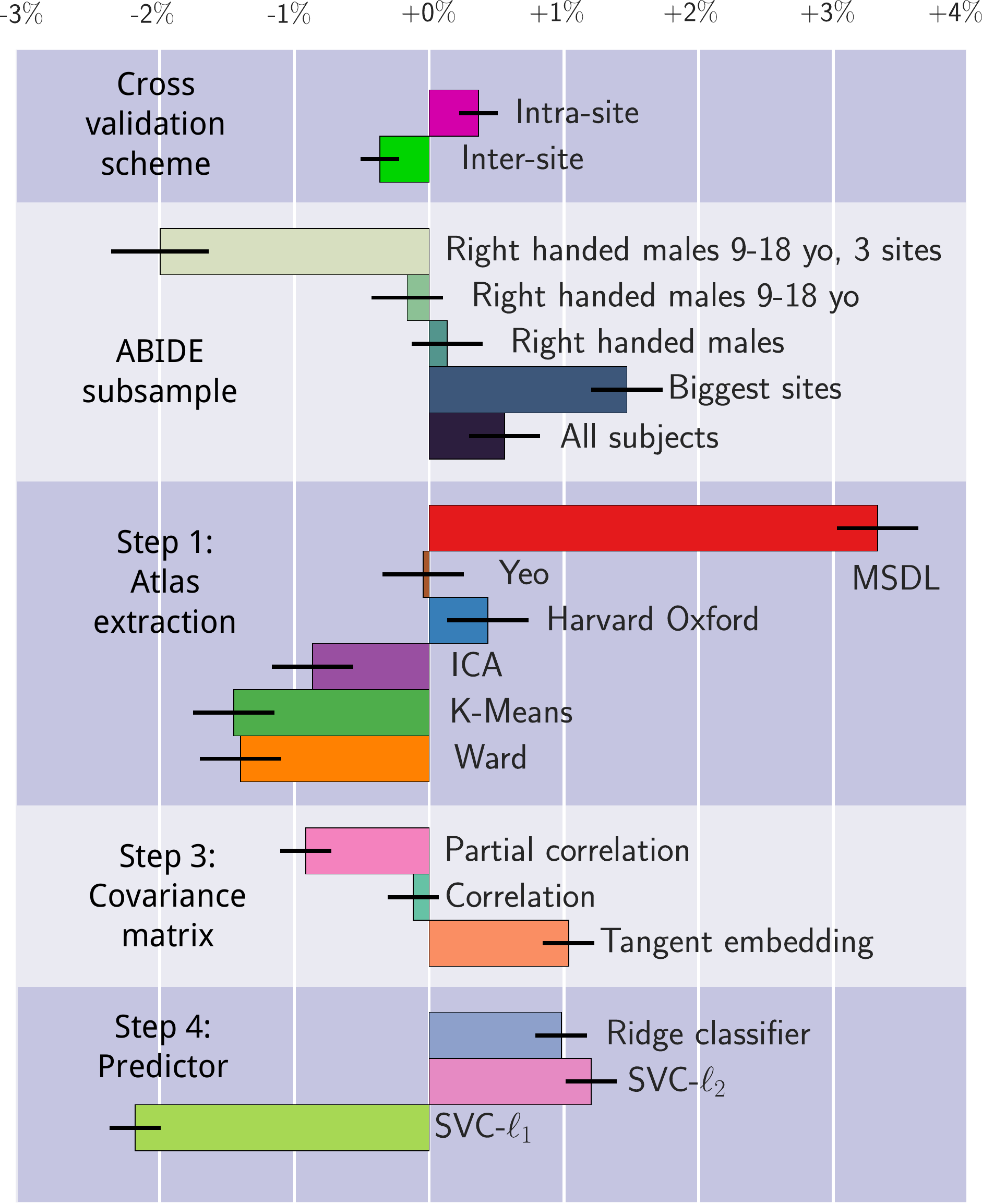}%
    }
    \caption{\textbf{Impact of pipeline steps on prediction.}
        Each block of bars represents a step of the pipeline (namely step 1, 3
        and 4). We also prepended two steps corresponding to the
        cross-validation schemes and ABIDE subsamples. Each bar represents the
        impact of the corresponding option on the prediction accuracy,
        relative to the mean prediction. This effect is measured via a
        full-factorial analysis of variance (ANOVA), analyzing the
        contribution of each step in a linear model. Each step of the pipeline
        is considered as a categorical variable. Error bars give the 95\%
        confidence interval. Multi Subject Dictionary Learning (MSDL) atlas
        extraction method gives significantly better results while reference
        atlases are slightly better than the mean. Among all matrix types,
        tangent embedding is the best on all ABIDE subsets. Finally,
        $l_2$-regularized classifiers perform
	better than $l_1$-regularized ones.
        Results for each subsamples are presented in the appendices
	--\autoref{fig:full_pipeline_steps}.
    \label{fig:pipeline_steps}}
\end{figure}

\begin{figure}[tb]
    \includegraphics[width=\linewidth]{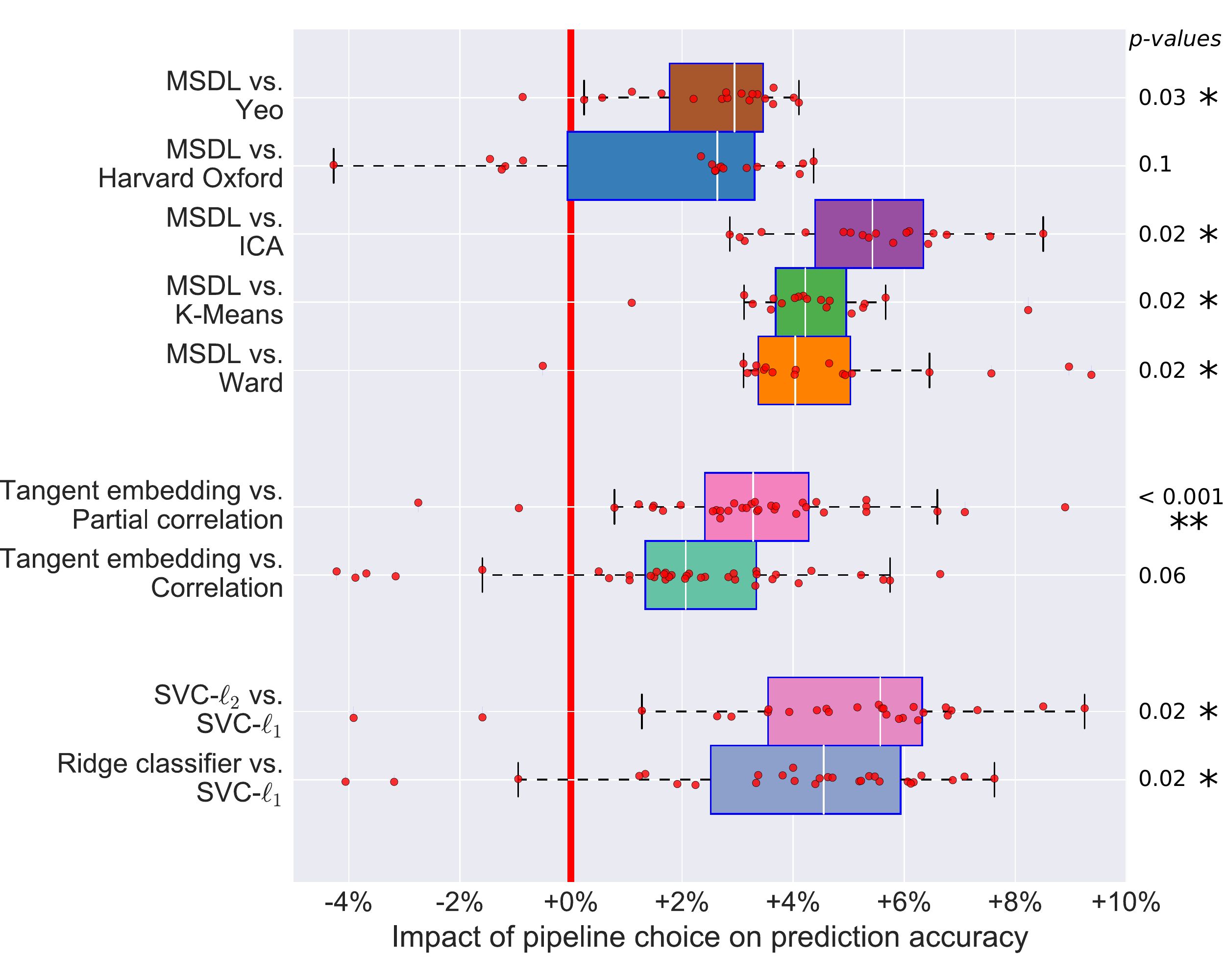}%

    \caption{\textbf{Comparison of pipeline options.}
        Each plot compares the classification accuracy
    scores if one parameter is changed and the others kept fixed.
    We measured statistical signficance using a Wilcoxon signed rank test and
    corrected for multiple comparisons.
    Detailed one-to-one comparison of the scores are shown in the appendices
    --\autoref{fig:full_comparison_plots}}
    \label{fig:comparison_plots}
\end{figure}

As shown in \autoref{fig:pipeline_steps}, the best predicting
pipeline is MSDL, tangent space embedding and $l_2$-regularized classifiers
(SVC-$l_2$ and ridge classifier). All effects are observed with $p < 0.01$.
Detailed pairwise comparisons (\autoref{fig:comparison_plots}) confirm these trends by showing the
effect of each option compared to the best ones. Results of intra-site
prediction have smaller standard error, confirming higher variability of
inter-site prediction results (\autoref{fig:full_comparison_plots}).

\begin{table}[tb]

\small%
\begin{tabularx}{\linewidth}{p{1.4cm}|Y|Y|Y|Y|Y}
    Subsample & \#5 & \#4 & \#3 & \#2 & \#1 \\
    
\hline
Inter-site& & $69.7\%$ & $65.1\%$ & $68.7\%$ & $66.8\%$\\
Accuracy& & $ \pm 8.9\%$ & $ \pm 5.8\%$ & $ \pm 9.3\%$ & $ \pm
5.4\%$\\[.1ex]
\rowcolor{gray!15}%
\footnotesize Inter-site& & \footnotesize $74.0\%$ & \footnotesize $69.1\%$ & \footnotesize $73.9\%$ & \footnotesize $72.3\%$\\
\rowcolor{gray!15}%
\footnotesize Specificity& & \footnotesize $ \pm 12.9\%$ & \footnotesize $ \pm 7.8\%$ & \footnotesize $ \pm 11.7\%$ & \footnotesize $ \pm 12.1\%$\\[.1ex]
\footnotesize Inter-site& & \footnotesize $65.4\%$ & \footnotesize $61.3\%$ & \footnotesize $62.8\%$ & \footnotesize $61.0\%$\\
\footnotesize Sensitivity& & \footnotesize $ \pm 13.7\%$ & \footnotesize $ \pm 7.7\%$ & \footnotesize $ \pm 13.6\%$ & \footnotesize $ \pm 17.0\%$\\
\hline\\[-1em]
Intra-site & $66.6\%$ & $65.8\%$ & $65.7\%$ & $67.9\%$ & $66.9\%$\\
Accuracy & $ \pm 5.4\%$ & $ \pm 5.9\%$ & $ \pm 4.9\%$ & $ \pm 1.9\%$ & $
\pm 2.7\%$\\[.1ex]
\rowcolor{gray!15}%
\footnotesize Intra-site & \footnotesize $78.4\%$ & \footnotesize $72.3\%$ & \footnotesize $77.5\%$ & \footnotesize $76.6\%$ & \footnotesize $78.3\%$\\
\rowcolor{gray!15}%
\footnotesize Specificity & \footnotesize $ \pm 9.7\%$ & \footnotesize $ \pm 9.1\%$ & \footnotesize $ \pm 8.1\%$ & \footnotesize $ \pm 5.0\%$ & \footnotesize $ \pm 4.1\%$\\[.1ex]
\footnotesize Intra-site & \footnotesize $53.2\%$ & \footnotesize $58.1\%$ & \footnotesize $52.2\%$ & \footnotesize $58.1\%$ & \footnotesize $53.2\%$\\
\footnotesize Sensitivity & \footnotesize $ \pm 10.9\%$ & \footnotesize $ \pm 9.2\%$ & \footnotesize $ \pm 9.1\%$ & \footnotesize $ \pm 5.0\%$ & \footnotesize $ \pm 5.8\%$\\
\hline
\end{tabularx}

    \caption{\textbf{
        Average accuracy, specificity and sensitivity scores (and standard deviation) for top performing
    pipelines.}
        Accuracy is the fraction of well-classified individuals (chance level
	is at 53.7\%).
        Specificity is the fraction of well classified healthy individuals among
        all healthy individuals (percentage of well-classified negatives).
        Sensitivity is the ratio of ASD individuals among all subjects
        classified as ASD (percentage of positives that are true positives).
        Results per atlas are available in the appendices
	--\autoref{tab:full_scores}.}
    \label{tab:scores}
\end{table}

\subsection{Effect of the choice of atlas}

The choice of atlas appears to have the greatest impact on prediction
accuracy (\autoref{fig:pipeline_steps}). Estimation of an atlas from the
data with MSDL, or using reference atlases leads to
maximal performance. Of note, for smaller samples, data-driven
atlas-extraction methods other than MSDL perform poorly
(\autoref{fig:full_pipeline_steps}). In this regime, 
noise can be
problematic and limit the generalizability of atlases derived from
data.

MSDL performs well regardless of sample size, and the performance gain is
significant compared to all other strategies apart from the Harvard-Oxford
atlas (\autoref{fig:comparison_plots}). This is likely attributable to
MSDL's strong spatially-structured regularization that increases its robustness to
noise.

\subsection{Effect of the covariance matrix estimator}

While correlation and partial correlation
approaches are currently the two dominant approaches to connectome
estimation in the current R-fMRI literature, our results highlight the
value of tangent-space projection that outperforms the other approaches in all
settings (\autoref{fig:pipeline_steps} and
\autoref{fig:comparison_plots}), though the difference is significant
only compared with correlation matrices (\autoref{fig:comparison_plots}).
This gain in prediction accuracy is not without a
cost, as the matrices derived cannot be read as correlation matrices.
Yet, the differences in connectivity that they capture
can be directly interpreted \cite{varoquaux2010b}.

Among the correlation-based approaches, full correlations outperformed
partial correlations, most notably for intra-site prediction. These
limitations of partial correlations may reflect estimation
challenges\footnote{Note that we experimented with a variety of
more sophisticated covariance estimators, including GraphLasso, though without
improved results.}
with the relatively short time series included in ABIDE datasets
(typically 5-6 minutes per participant).

\subsection{Effect of the final classifier}

The results show that $\ell_2$-regularized classifiers perform best
in all settings (\autoref{fig:pipeline_steps} and
\autoref{fig:comparison_plots}). This may be due to the fact that a global
hypoconnectivity, as previously observed in ASD patients, is not well captured
by $\ell_1$-regularized classifiers, which induce sparsity.
In addition $\ell_2$-penalization is rotationally-invariant, which means that it is not
sensitive to an arbitrary mixing of features. Such mixing may happen if a 
functional
brain module sits astride several regions. Thus, a possible explanation for the
good performance of $\ell_2$-penalization is that it does not need an
optimal choice of regions, perfectly aligned with the underlying
functional modules.

\subsection{Effect of the dataset size and heterogeneity}

While not a property of the data-processing pipeline itself, a large
number of training subjects is the most important factor of success for
prediction. Indeed, we observe that prediction results improve with the
number of subjects, even for large number of subjects
(\autoref{fig:pipeline_steps}). 

Despite that increase, we note that prediction on the largest
subsample (subsample \#1) gives lower scores than prediction on subsample \#2,
which may be due to the variance introduced by small sites.
We also note that the importance of the
choice between each option of the pipeline decreases with the number of
subjects (\autoref{fig:full_pipeline_steps}).

\subsection{Effect of the number of regions}

\begin{figure}[tb]
    \centerline{%
    \includegraphics[width=.9\linewidth]{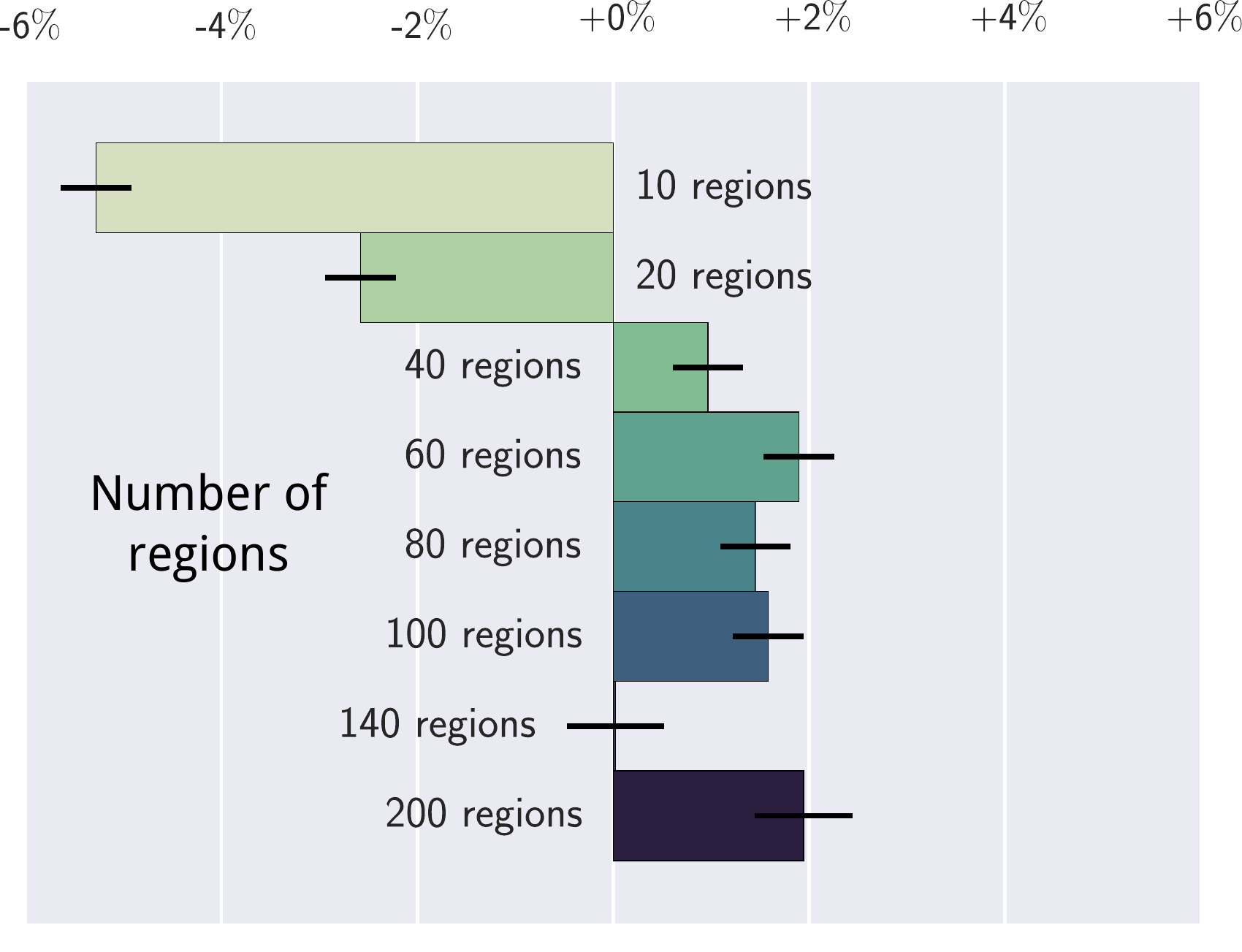}%
    }%
    \caption{\textbf{Impact of the number of regions on prediction}: each
        bar indicates the impact of the number of regions on the prediction accuracy
        relative to the mean of the prediction. These values are coefficients in a linear
        model explaining the best classification scores as function of
        the number of regions. Error bars give the 95\% confidence interval,
        computed by a full factorial ANOVA. Atlases containing more than 40 ROIs give
        better results in all settings. Results per
        subsamples are presented in the appendices
	--\autoref{fig:full_region_count}.}
    \label{fig:region_count}
\end{figure}

In order to compare models of similar complexity, the previous experiments 
were restricted to
84 regions as in \cite{abraham2013}. However, this choice is based on previous
studies and may not be the optimal choice for our problem.
To validate this aspect of the study, we also varied
the number of regions and analyzed its impact on the classification results in
\autoref{fig:region_count}, similarly to
\autoref{fig:pipeline_steps}. In order to avoid computational cost\footnote{
As MSDL is much more computationally expensive than Ward, 
we have not explored the
effect of modifying of the number of regions extracted with MSDL.
However, we expect that the optimal number of
regions is not finely dependent on the region-extraction method.}, we
explored only two atlases: the data-driven Ward's hierarchical clustering and
spectral clustering atlases computed in \cite{craddock2012}.

Across all settings, we distinguish 3 different regimes. First, below
20 ROIs, performance is bad. Indeed very large regions
average out signals of different functional areas.
Above 140 ROIs, the results seem to be unstable, though
this trend is less clear in intra-site prediction. Finally, 
the best results are between 40 and 100 ROIs. Our choice of
84 ROIs is within the range of
best performing models.

\subsection{Inspecting discriminative connections}
\label{subsec:connections}
\begin{figure*}[tb]
    \begin{minipage}[t]{.3\linewidth}
	\includegraphics[width=\linewidth]{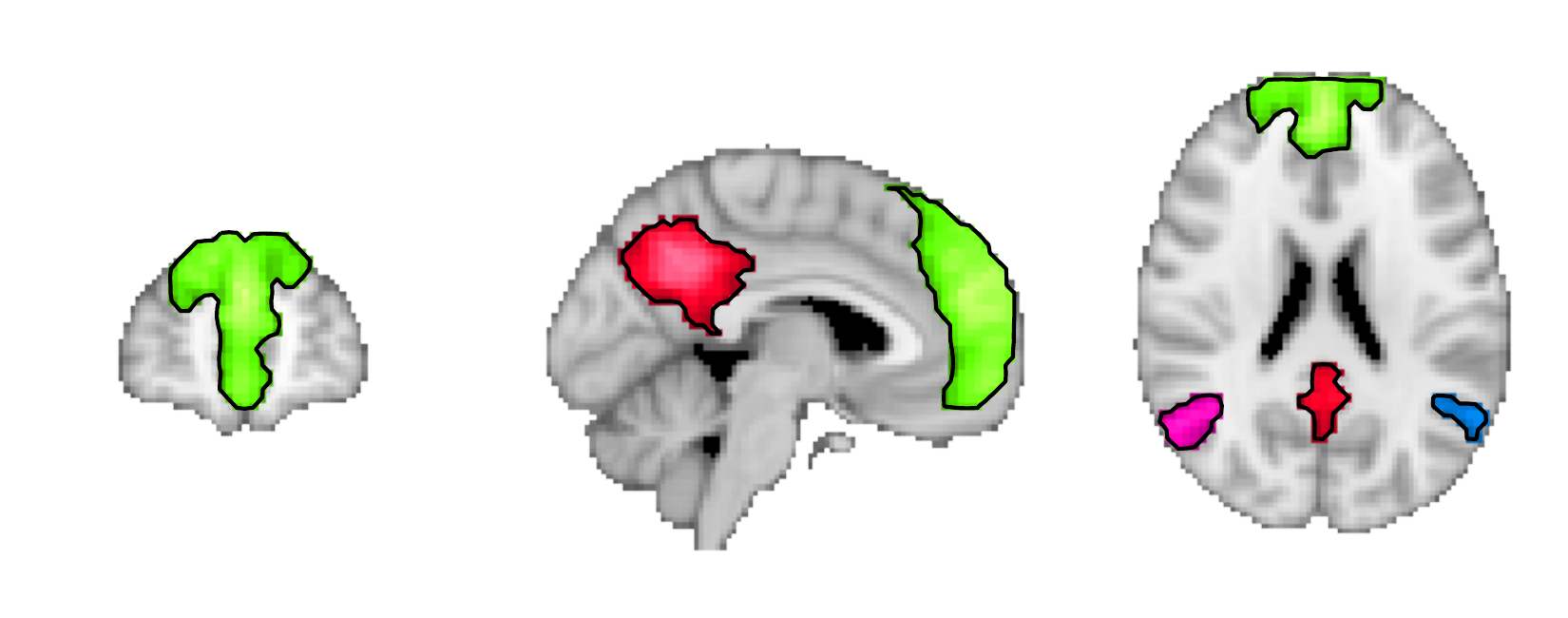}%
	\llap{%
	\raisebox{.32\linewidth}{\rlap{\textbf{\sffamily
		Default Mode Network}}%
	\hspace*{\linewidth}}}

	\includegraphics[width=\linewidth]{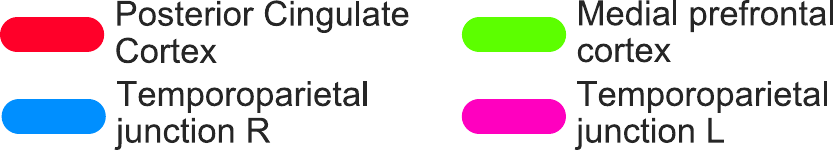}
	\includegraphics[width=\linewidth]{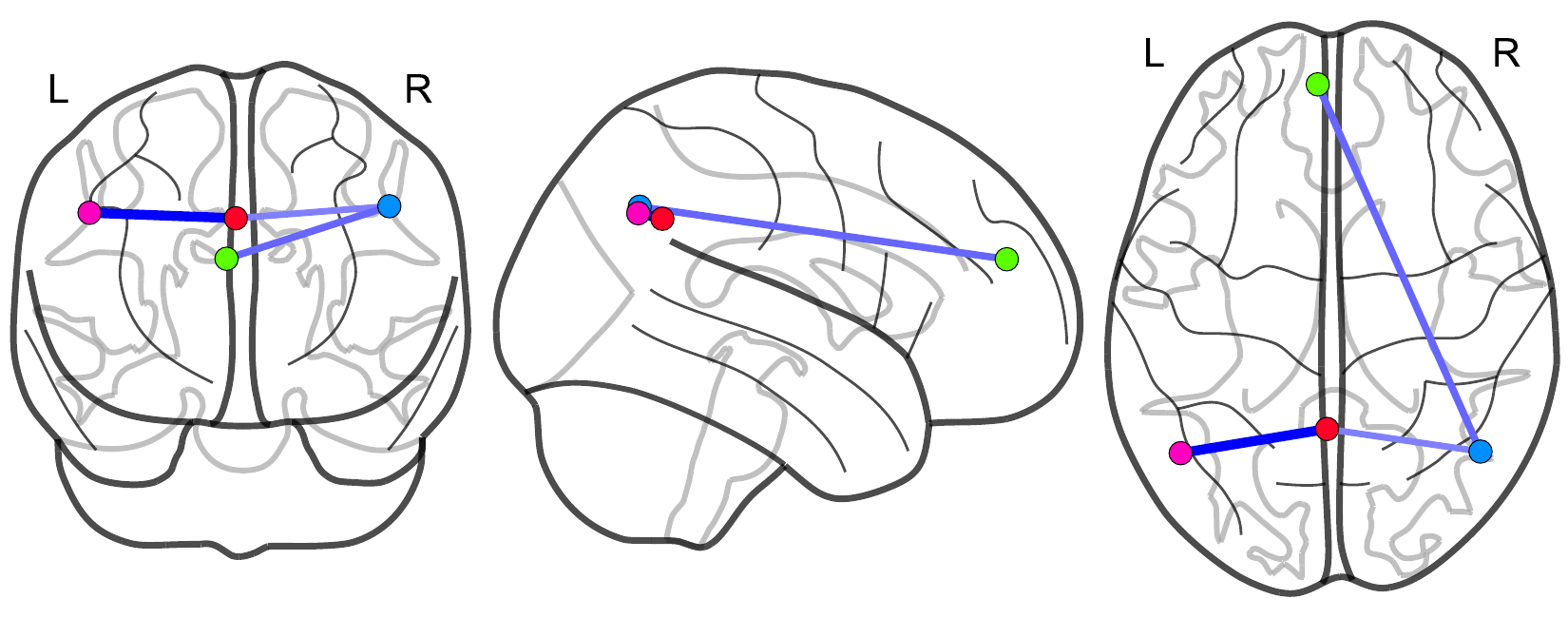}%
	\vspace*{-1.4em}

	\textbf{\emph{a}}
    \end{minipage}\hfill%
    \begin{minipage}[t]{.3\linewidth}
	\includegraphics[width=\linewidth]{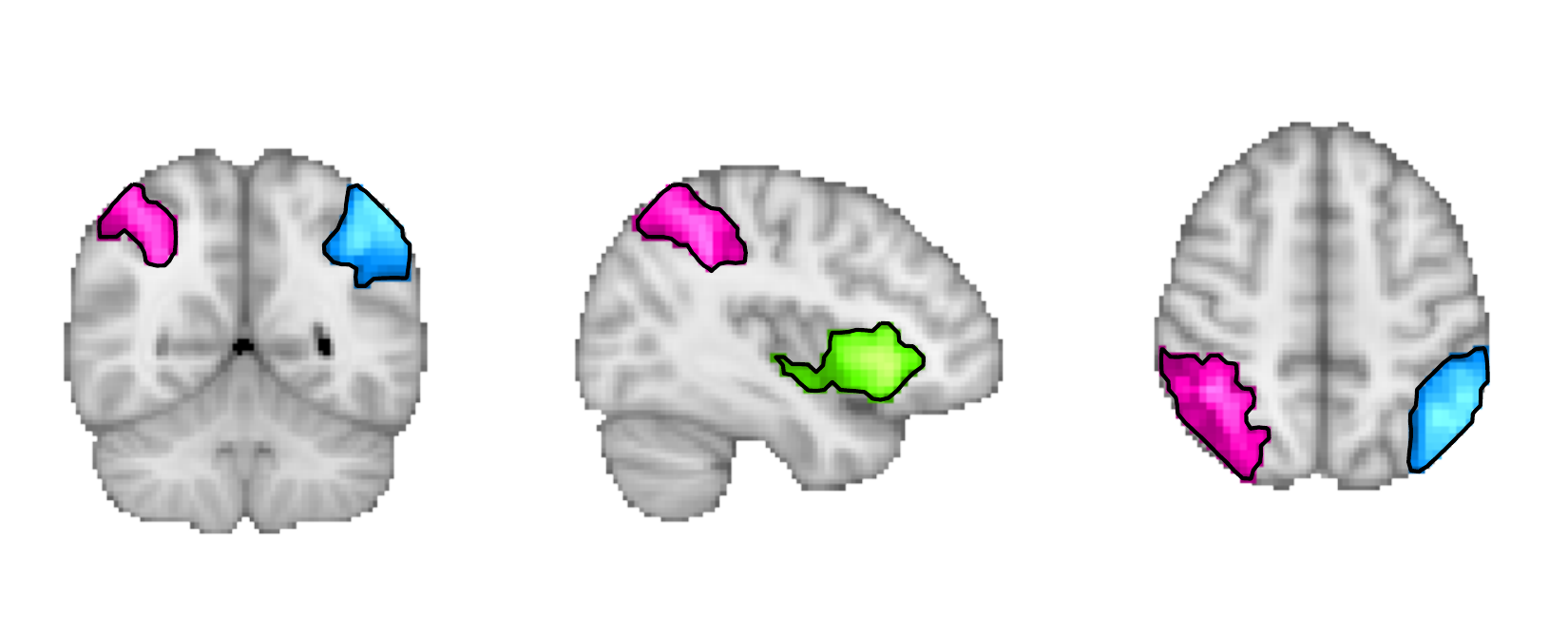}%
	\llap{%
	\raisebox{.32\linewidth}{\rlap{\textbf{\sffamily
		Pareto-insular network}}%
	\hspace*{\linewidth}}}

	\includegraphics[width=\linewidth]{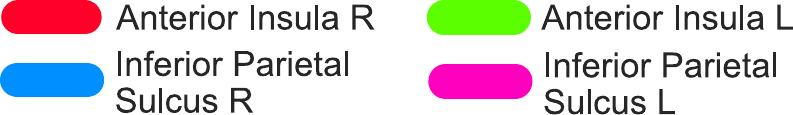}
	\includegraphics[width=\linewidth]{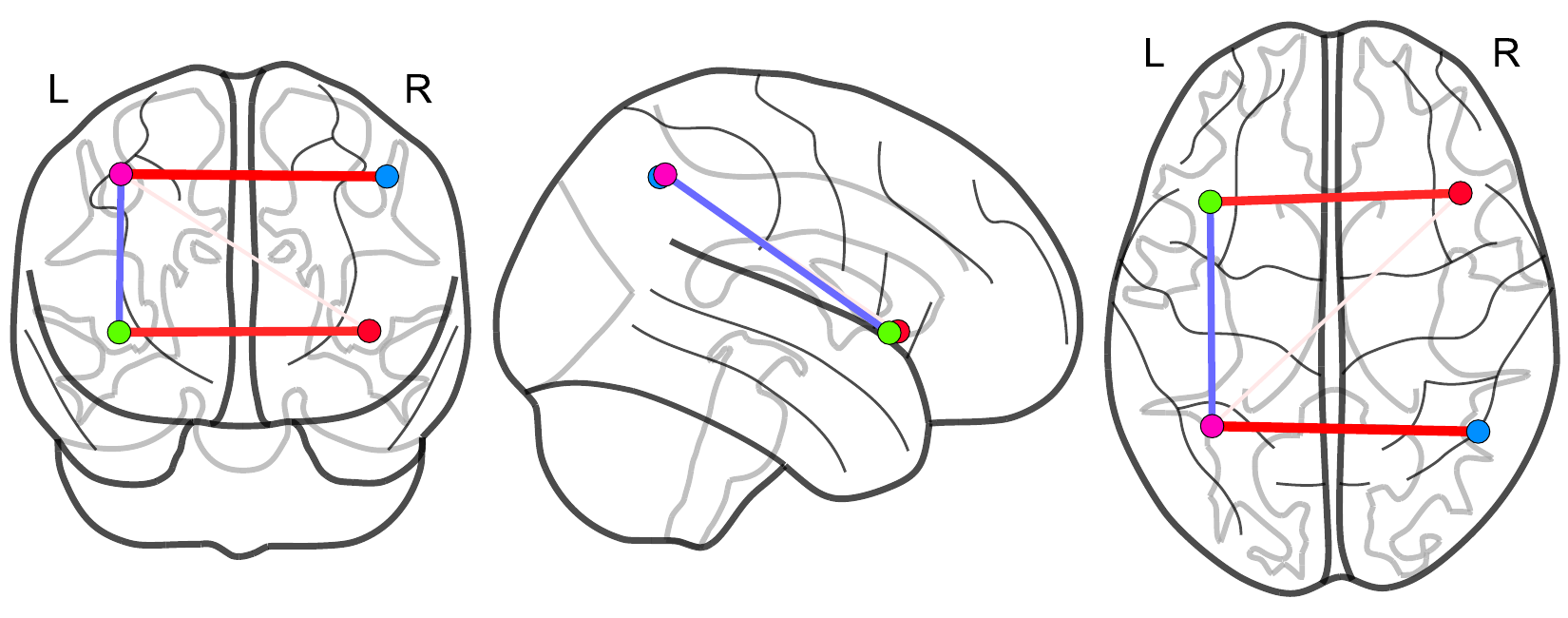}%
	\vspace*{-1.4em}

	\textbf{\emph{b}}
    \end{minipage}\hfill%
    \begin{minipage}[t]{.3\linewidth}
	\includegraphics[width=\linewidth]{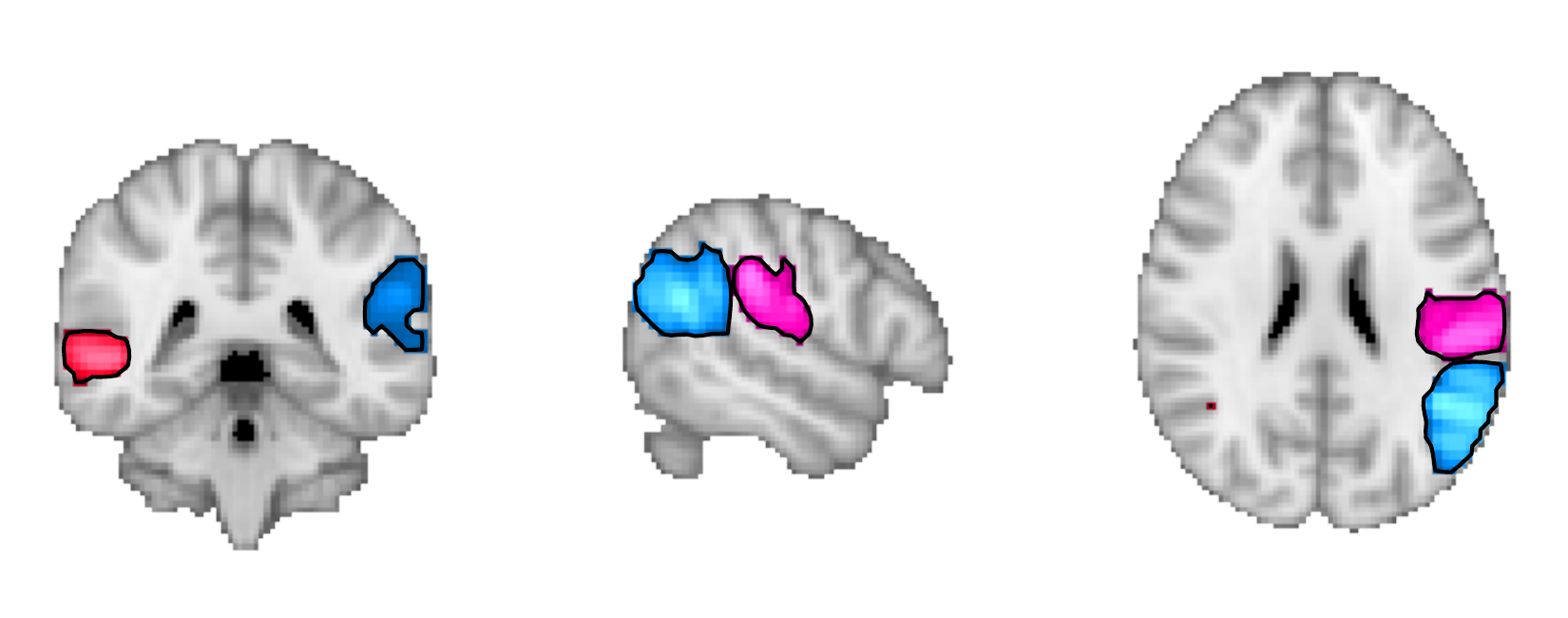}%
	\llap{%
	\raisebox{.32\linewidth}{\rlap{\textbf{\sffamily Semantic ROIs}}%
	\hspace*{\linewidth}}}

	\includegraphics[width=\linewidth]{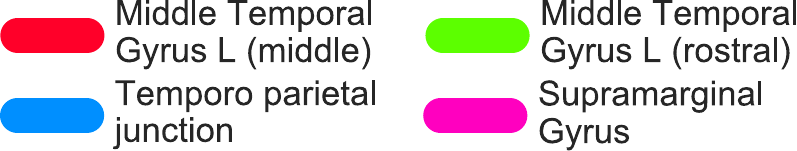}
	\includegraphics[width=\linewidth]{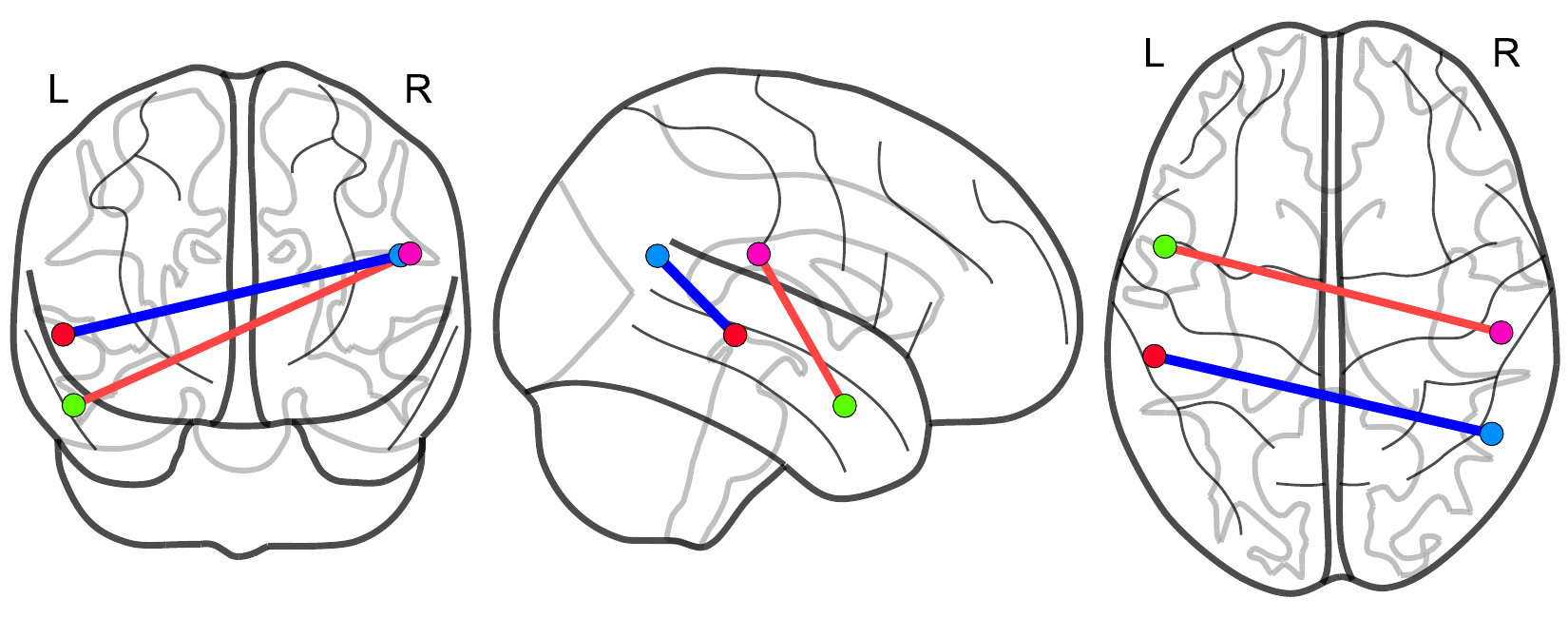}%
	\vspace*{-1.4em}

	\textbf{\emph{c}}
    \end{minipage}\\
    \vspace{14pt}
    \begin{minipage}{.48\linewidth}
        \includegraphics[width=\linewidth]{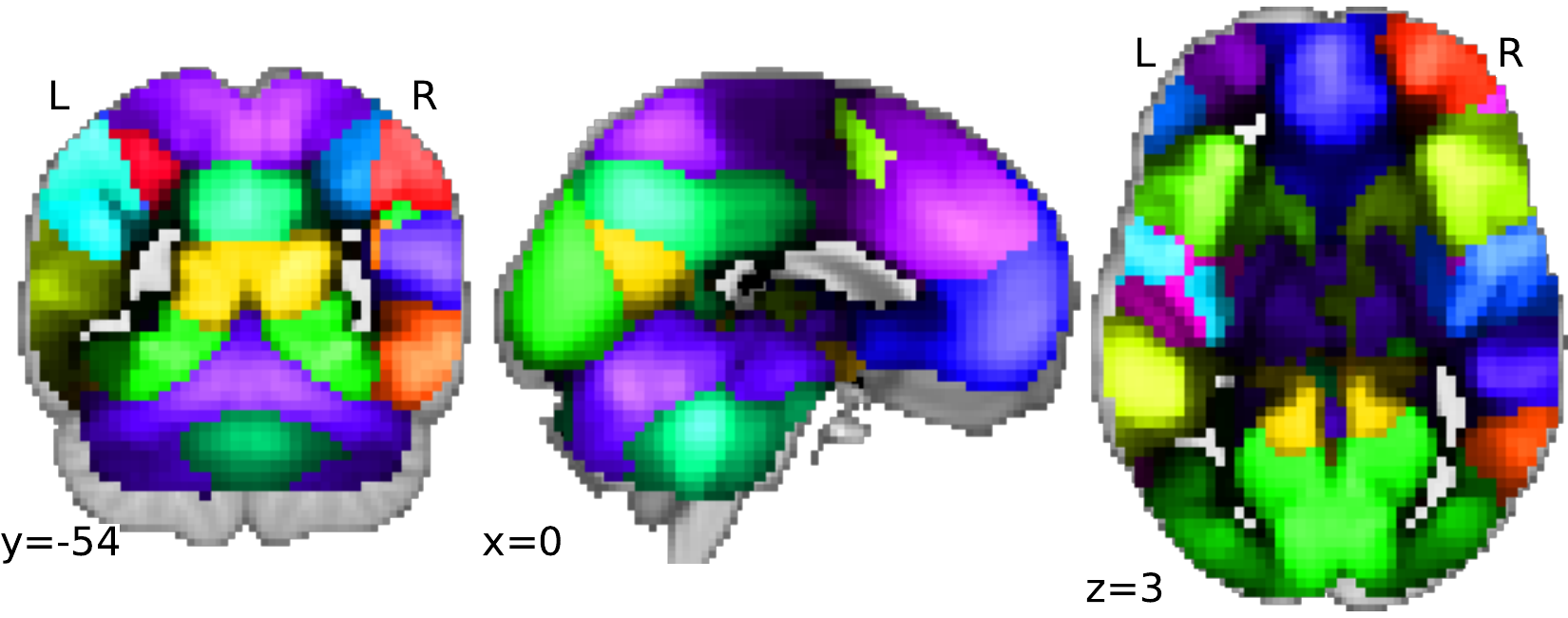}
    \end{minipage}\hfill%
    \begin{minipage}{.48\linewidth}
    \caption{\textbf{Significant non-zero connections in the predictive
        biomarkers distinguishing controls from ASD patients.}
            Red connections are stronger in controls and blue connections
            are stronger in ASD patients. Subfigures $a$, $b$, and $c$ reported
            intra-network difference. On the left is the consensus atlas
            obtained by selecting regions consistently extracted on 10
            subsets of ABIDE. Colors are randomly assigned. Results obtained on other
        networks are shown in the appendices --\autoref{fig:full_networks}.}
    \label{fig:networks}
    \vspace{32pt}
    \end{minipage}
\end{figure*}

A connection between two regions is considered discriminative if its value helps
diagnosing ASD from TC. To understand what is driving the best prediction
pipeline, we extract its most discriminative
connections as described in \autoref{par:biomarkers}.
Given that the 10-fold cross-validation yields 10 different atlases, we
start by building a consensus atlas by taking all the regions with a
DICE similarity score above $0.9$. We obtain an atlas composed of 37 regions
(see \autoref{fig:networks}). These regions and the discriminative connections
form the neurophenotypes. Note that discriminative
connections should be interpreted with caution. Indeed, our study covers a
wide range of age and diagnostic criteria. Additionally, predictive models
cannot be interpreted as edge-level tests.

\paragraph{Default Mode Network (\autoref{fig:networks}.a)}
We observe a lower connectivity between left and right temporo-parietal
junctions. Both hypoconnectivity between regions on adolescent and adult
patients \cite{cherkassky2006,kennedy2006} and
hyperconnectivity in children \cite{supekar2013} have been reported. Our
findings are concordant with \cite{monk2009} that observed lower
fronto-parietal connectivity and stronger connectivity between temporal ROIs.

\paragraph{Pareto-insular network (\autoref{fig:networks}.b)}
We observe interhemispheric hypoconnectivity
between left and right anterior insulae, and left and right inferior parietal
lobes. Those regions are part of a frontoparietal control network related to
learning and memory \cite{iidaka2006}. Such hypoconnectivity has already
been widely evoked in ASD \cite{anderson2010,dimartino2011}
and previously found in the ABIDE dataset \cite{plitt2014,ray2014}.
These ROIs also match anatomical regions of
increased cortical density in ASD \cite{haar2014}.

\paragraph{Semantic ROIs (\autoref{fig:networks}.c)} We observe a more complicated pattern
in ROIs related to language.
Connectivity is stronger between the right supramarginal gyrus and the left Middle
Temporal Gyrus (MTG) while a lower connectivity is observed between the right MTG
and the left temporo-parietal junction. Differences of laterality in language
networks implicating the middle temporal gyri have already been observed
\cite{kleinhans2008}. Although this symptom is often considered as decorrelated from
ASD, we find that these regions are relevant for diagnosis.

\section{Discussion}



We studied pipelines that extract neurophenotypes from aggregate R-fMRI
datasets through the following steps: 
\begin{enumerate*}[label=\arabic*)]
    \item region definition from R-fMRI,
    \item extraction of regions activity time series
    \item estimation of functional interactions between regions, and
    \item construction of a discriminant model for brain-based
    diagnostic classification.
\end{enumerate*}
The proposed pipelines can be built with the Nilearn
neuroimaging-analysis software and the atlas computed with MSDL is available for download
\footnote{\url{http://team.inria.fr/parietal/files/2015/07/MSDL_ABIDE.zip}}.
We have demonstrated that they can successfully predict diagnostic
status across new acquisition sites in a real-world situation, a large
Autism database.
This validation with a leave-one-site-out
cross-validation strategy reduces
the risk of overfitting data from a single
site, as opposed to the commonly used leave-one-subject-out
validation approach \cite{nielsen2013}. It enabled us
to measure the impact of methodological choices on prediction.
In the ABIDE dataset, this approach classified the label of
autism versus control with an accuracy of 67\% -- a rate superior to prior work
using the larger ABIDE sample.

The heterogeneity of weakly-controlled datasets formed by aggregation of
multiple sites poses great challenges to develop brain-based
classifiers for psychiatric illnesses.
Among those most commonly cited are:
\begin{enumerate*}[label=\textit{\roman*)}]
    \item the many sources of uncontrolled variation that can arise across
          studies and sites (\eg scanner type, pulse sequence, recruitment
          strategies, sample composition),
    \item the potential for developing classifiers that are reflective of only
          the largest sites included, and
    \item the danger of extracting biomarkers that will be useful only on sites included
          in the training sample.
\end{enumerate*}
Counter to these concerns,
our results demonstrated the feasibility of using weakly-controlled
heterogeneous datasets to identify imaging biomarkers that are robust to
differences among sites. Indeed, 
our prediction accuracy of 67\% is the highest reported to-date
on the whole ABIDE dataset (compared to 60\% in \cite{nielsen2013}), 
though better prediction has been reported 
on curated more homogeneous subsets \cite{plitt2014, chen2015}.
Importantly, we found that increasing sample size helped tackling
heterogeneity as
inter-site prediction accuracy approached that of intra-site prediction.
This undoubtedly increases the perceived value that can be derived from
aggregate datasets, such as the ABIDE and the ADHD-200 \cite{adhd2012}.

Our results suggest directions to improve prediction accuracy. 
First, more data is needed to optimize the classifier. Indeed, our experiments 
show that with a training set of 690
participants aggregated across multiple sites (80\% of 871 available subjects), the pipelines have not reached their optimal performance. 
Second, a broader diversity of data will be needed to more definitively assess
prediction for sites not included in the training
set. Both of these needs motivate more data sharing.

Methodologically, our systematic comparison of choices in the pipeline steps
can give indications to establish standard
processing pipelines. Given
that processing pipelines differ greatly between studies \cite{carp2012}, comparing
data-analysis methods is challenging. On the ABIDE dataset, we found that 
a good choice of processing steps can strongly improve prediction
accuracy.
In particular, the choice of regions is most important. We found that
extracting these regions with MSDL gives best results. However, the
benefit of this method is not always clear cut -- on small datasets,
references atlases are also good performers. For the rest of the pipeline,
tangent embedding and $\ell_2$-regularized classifier are overall the
best choice to maximize prediction accuracy.

While the primary goals of this study were methodological, it is worth
noting that biomarkers identified were largely consistent with the current
understanding
of the neural basis of ASD. The most informative features to
predict ASD concentrated in the
intrinsic functional connectivity of 
three main functional systems: the default-mode, parieto-insular, and language
networks, previously involved in ASD \cite{vissers2012, rane2015, hernandez2015}. 
Specifically, decreased homotopic connectivity between temporo-parietal
junction, insula and inferior parietal lobes appeared to characterize ASD.
Decreases in homotopic functional connectivity have been previously reported in
R-fMRI studies with moderately \cite{anderson2010, dinstein2011} or large
samples such as ABIDE \cite{dimartino2014}.
Here our findings point toward a set of
inter-hemispheric connections involving heteromodal association cortex subserving
social cognition (\ie temporo-parietal junction \cite{saxe2003}), and cognitive
control (anterior insula and right inferior parietal cortex \cite{menon2010}), commonly
affected in ASD \cite{anderson2010,dimartino2011, plitt2014,ray2014}.
Limitations of our biomarkers may arise from the heterogeneity of ASD as
a neuropsychiatric disorder \cite{geschwind2007}.

Beyond forming a neural correlate of the disorder, predictive biomarkers
can help define diagnosis subgroups -- the next logical endeavor in
classification studies \cite{loth2016}.

As a psychiatric study, the present work has a number of limitations. First,
although composed of samples from geographically distinct sites, the
representativeness of the ABIDE sample has not been established. As such, the
features identified may be biased and necessitate further replication;
though their consistency with prior findings alleviates this concern to some degree.
Further work calls for characterizing this prediction pipeline on more
participants (\eg ABIDE II) and different
pathologies (\eg ADHD-200 dataset).
Second, while the present work primarily relied on prediction accuracy to
assess the classifiers derived, a variety of alternative measures exist
(\eg sensitivity, specificity, J-statistic). This decision was largely
motivated to facilitate comparison of our findings with those from
prior studies using ABIDE, which most frequently used prediction accuracy.
Depending on the specific application for a
classifier, the prediction metric to optimize may vary. For
example screening tools would need to favor sensitivity over accuracy or
specificity; in contrast with medical diagnostic tests that require specificity over
sensitivity or accuracy \cite{castellanos2013}.
The versatility of the pipeline studied allows us to maximize
either of these scores depending on the experimenter's needs. Finally,
while the present work focused on the ABIDE dataset, 
it is likely that its findings regarding optimal pipeline decisions will
carry over to other aggregate samples, as well as more homogeneous samples.
With that said, future applications will be required to verify this point.

\medskip

In sum, we experimented with a large number of different pipelines and
parametrizations on the task of predicting ASD diagnosis on the ABIDE dataset.
From an extensive study of the results,
we have found that R-fMRI from a large amount of participants sampled
from many different sites could lead to functional-connectivity biomarkers of
ASD that are robust, including to inter-site variability. Their predictive power
markedly increases with the number of participants included. This increase holds
promise for optimal biomarkers forming good probes of disorders.

\bigskip

\textbf{Acknowledgments.} We acknowledge funding from the NiConnect project
and the SUBSample project from the DIGITEO Institute, France. The effort of
Adriana Di Martino was partially supported by the NIMH grant 1R21MH107045-01A1.
Finally, we thank the anonymous reviewers for their feedback, 
as well as all sites and investigators who have worked to share their
data through ABIDE.

\FloatBarrier

\section*{References}

\bibliography{biblio}
\newpage
\appendix

\renewcommand{\thefigure}{A\arabic{figure}}
\setcounter{figure}{0}
\renewcommand{\thetable}{A\arabic{table}}
\setcounter{table}{0}

\section{Time series extraction}

\label{app:time_series}

In order to extract time series from overlapping brain maps, we use an ordinary
least square approach (\textbf{OLS}). Let $\B{Y} \in \mathbb{R}^{n \times p}$ be the original
subject signals of $p$ voxels over $n$ temporal scans and $\B{V} \in
\mathbb{R}^{k \times p}$ the atlas of $k$ maps containing $p$ voxels. We
estimate $\B{U} \in \mathbb{R}^{n \times k}$, the set of time series proper to
each region, using an \textbf{OLS}:

\[\hat{\B{U}} = \arg\min_{\B{U}} \|\B{Y} - \B{UV}\|\]

In order to remove the confounds, we orthogonalize our signals with regards to
them. We do this by removing the the signals projected on an orthonormal basis of
the confounds. Let $\B{C} \in \mathbb{R}^{n \times c}$ be a matrix of $c$
confounds:

\[\B{C} = \B{QR} \;\;\;\text{with}\; \B{Q}\; \text{orthonormal}\]
\[\hat{\B{Y}} = \B{Y} - \B{Q}\B{Q}\trans\B{Y}\]

\section{Covariance matrix estimation}

\label{app:cov}

The empirical covariance matrix is not a good estimator when $n < p$. In
particular, it can leads to the estimation of matrices that cannot be inverted.
This problem can be avoided by using shrinkage, a method to pull the extreme
values of the covariance matrix toward the central ones. It is done by finding
the $\ell_2$-penalized maximum likelihood estimator of the covariance matrix:

\[\Sigma_{\text{shrunk}} = (1 - \alpha) \hat{\Sigma} + \alpha
    \frac{\text{Tr}(\hat{\Sigma})}{p}Id\]

We use the Ledoit-Wolf approach \cite{ledoit2004} to set parameter $\alpha$ to minimize the Mean
Square Error between the real and
the estimated covariance matrix. 

\section{Classifiers}

\label{app:classifiers}

Linear classifiers try to find an hyperplane splitting the sample in the input
space in order to classify them into two categories.

\subsection*{Support Vector Classification \textbf{SVC}}

\textbf{SVC} aims at maximizing the margin, \ie the distance to the closest
training sample and the hyperplane.
We use the classical squared hinge loss and thus minimize:

\[\min_{\B{w}} \frac{1}{2}|\B{w}|_p + C\sum_{i=1}^{n} \|\B{y}_i - \B{w}\trans\B{x}_i\|^2_2  \]

With $\B{w} \in \mathbb{R}^{n}$, $\B{x} \in \mathbb{R}^{n \times m}$, and
$\B{y} \in \mathbb{R}^m$. In our case, $n$, the number of features, is the
number of connections between pairs of brain regions and $m$, the number of
samples, is the number of subjects.
We explore the sparse regularization $\ell_1$ and $\ell_2$-penalized SVC
(resp. $p=1$ and $p=2$).
We use the implementation of scikit-learn \cite{pedregosa2011} that relies
on the implementation provided by LibLinear \cite{fan2008}.

\subsection*{Ridge classifier}

Ridge regression can be used in the context of classification by attributing
binary values to each of the classes (typically $-1$ and $1$). Classifying an
sample boils down to looking at the sign of the prediction of the regressor.

The Ridge regression minimizes the square distance to the hyperplane, like an
ordinary least squares, but it adds an $\ell_2$ penalization on the weights in order
to reward coefficients close to 0 and thus obtain the \textit{smallest} possible
model. This prevents colinear feature to skew the coefficients of the model. The
minimized function is:

\[\min_{\B{w}} \|\B{y} - \B{w}\trans\B{x}\|_2 + \|\B{w}\|_2 \]

We use the implementation of scikit-learn \cite{pedregosa2011}.

\section{Effect of movements}

\label{app:movements}

Participant's movements during fMRI acquisition is known to induce spurious
correlations in the connectome estimation. In our particular case, we expect
healthy individuals and ASD ones to exhibit different movement patterns because
of the behavioral symptoms of the latter.

In our study, we use standard methods to remove the effect of movement as much
as possible. We also performed an analysis to measure the amount of
information contained in the movement patterns and ensure that our predictions
are not based on potential movement residuals.

\subsection*{Movement regression}

\begin{figure}[bth]
    \includegraphics[width=\linewidth]{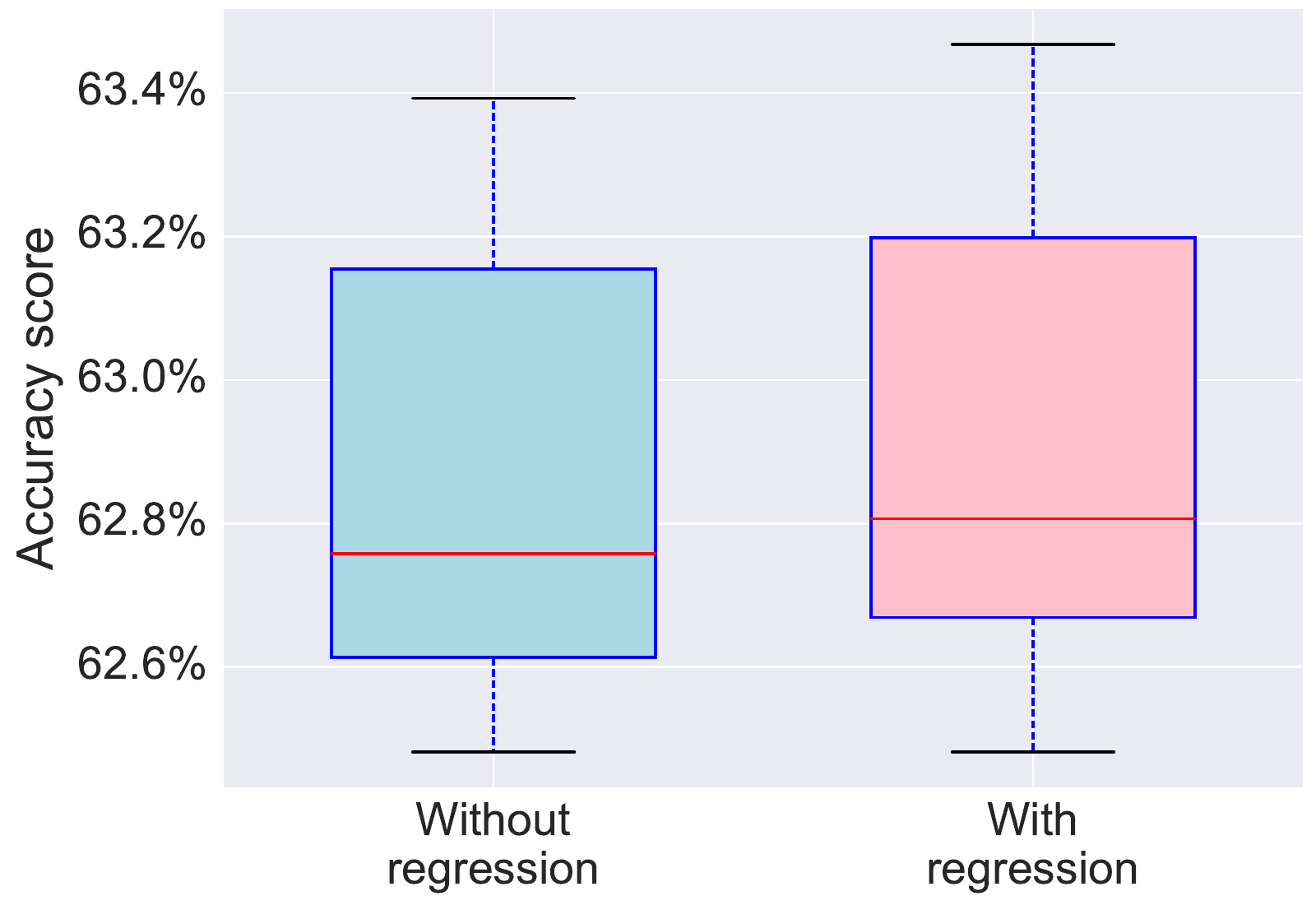}
    \caption{\textbf{Prediction accuracy scores with and without performing movement
        regression during time series extraction.} No significant difference
    is observed.}
    \label{fig:movement_effect}
\end{figure}

In the experiments, during time series extraction (step 2), we regress out all
possible confounding factors, which includes movement estimation. In order to
quantify the information contained in the movements, we try predicting ASD
diagnosis with and without movement regression. Figure~\ref{fig:movement_effect}
shows that no significant difference is observed with and without movement
regression, which means that movement effect does not affect strongly our connectome
estimation.

\subsection*{Movement-based prediction}

Going further, we also perform the diagnosis directly based on movement
estimations. Because movement parameters are time-series, it is not possible to
use them directly as features in a prediction task. For this purpose, we extract
temporal descriptors from these series. These descriptors are similar to the ones
extracted in FSL FIX \cite{salimi2014}, a tool to identify noisy ICA components:
coefficients of autoregressive models of several orders, kurtosis, skewness,
entropy, difference between the mean and the median, and Fourier transform
coefficients.

As for the prediction pipelines, we tried a wide range of strategies to predict
from movement parameters. We extracted the descriptors on the movements, the
gradient of the movements, the gradient squared, and the framewise displacement.
We tried several predictors (with and without feature extraction),
logistic regression, SVC, Ridge classifier,
Gaussian Naive Bayes and random forests.

We present the results obtained on the best prediction procedure.
For each participant, 56 descriptors are extracted from the movement estimations
and their gradient, and are used as features in an SVC for diagnosis. 
We observe that the scores obtained using movements are at chance level.
Note that we see a trend
already observed on prediction based on connectomes: The
variance of prediction scores is much larger for inter-site prediction.

\begin{table}[bth]

    \small%
\begin{tabularx}{\linewidth}{p{1.4cm}|Y|Y|Y|Y|Y}
 &
\multicolumn{3}{c|}{Right handed males} &
Biggest &
All \\
 & \multicolumn{2}{c|}{9-18 yo}& & sites & subjects \\
& 3 sites & & & & \\
\hline
Intra-site & $53.2\%$ & $53.9\%$ & $53.5\%$ & $53.0\%$ & $54.3\%$\\
Accuracy & $ \pm 0.1\%$ & $ \pm 0.6\%$ & $ \pm 0.2\%$ & $ \pm 0.2\%$ & $
\pm 0.1\%$\\[.1ex]
\hline
Inter-site& & $53.8\%$ & $54.3\%$ & $51.5\%$ & $52.8\%$\\
Accuracy& & $ \pm 8.1\%$ & $ \pm 8.3\%$ & $ \pm 7.3\%$ & $ \pm
8.2\%$\\[.1ex]
\end{tabularx}
    \caption{\textbf{Prediction accuracy scores based on movements estimations.}
        Results are similar across folds but variance of inter-site prediction
    is larger.}
    \label{tab:movement_prediction}
\end{table}
\vspace{-.5cm}
\begin{figure}[h]
    \centering
    \includegraphics[width=\linewidth]{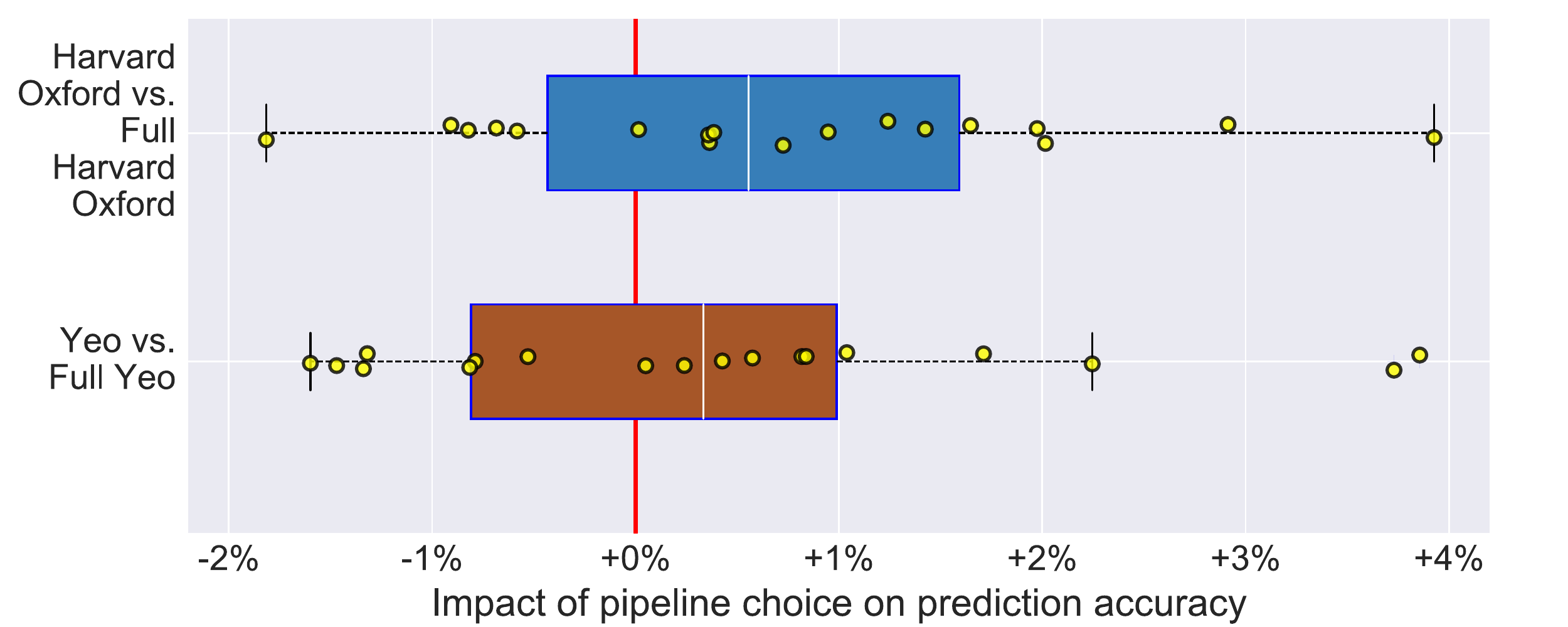}
    \caption{\textbf{Comparison of prediction results obtained using 84 regions
    extracted from Harvard Oxford and Yeo atlases against the set of
    all regions extracted from them (labelled as \textit{full}).}
    Results obtained using 84 regions are better than full atlases. This is probably
    due to the fact that small regions can induce spurious correlations in the
    connectivity matrices.}
    \label{fig:regions_vs_full}
\end{figure}

\begin{figure}[h]
        \includegraphics[width=\linewidth]{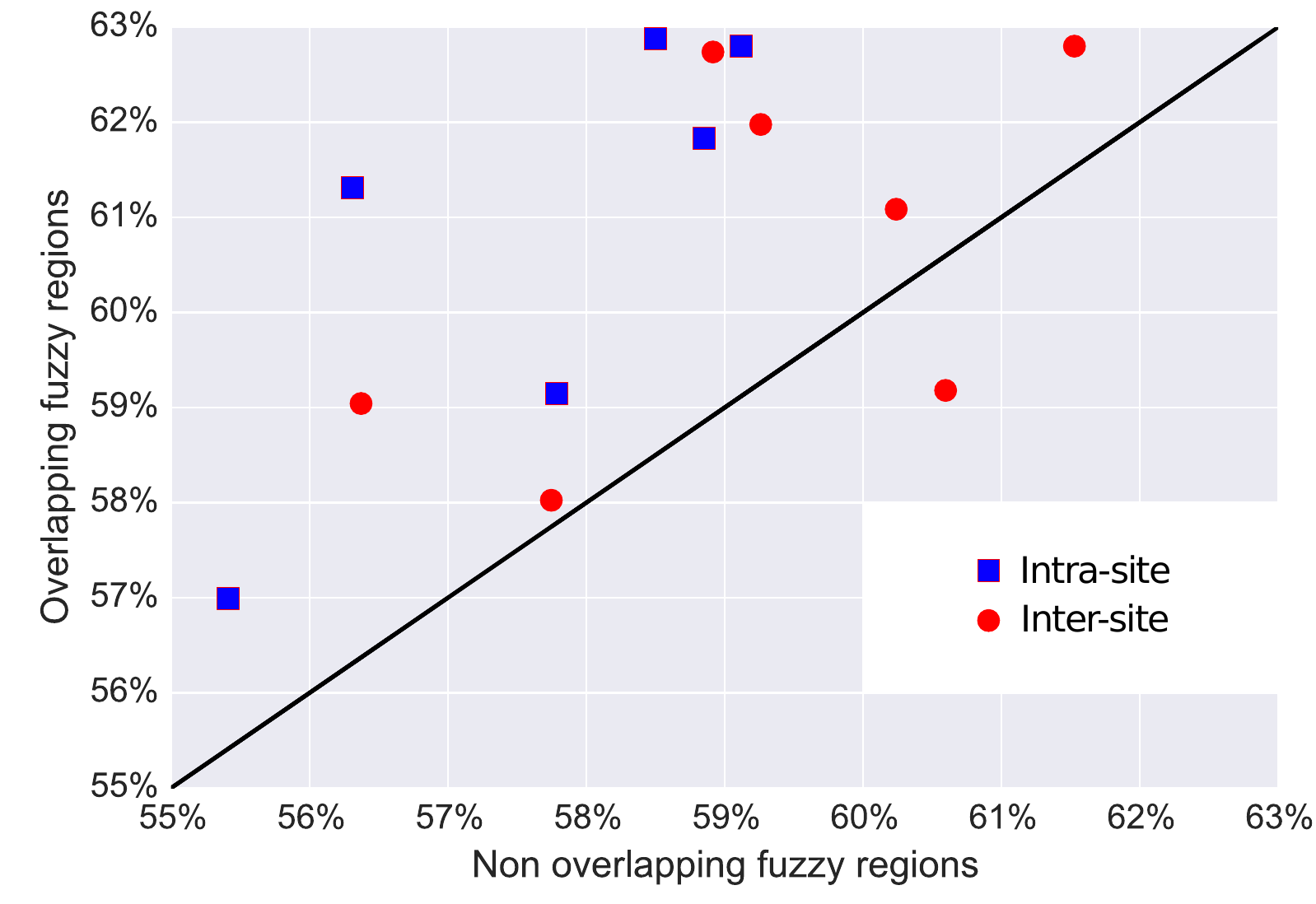}
    \caption{\textbf{Classification accuracy scores using overlapping atlases
        and their non overlapping counterpart.}
        Point above the identity line support the idea that fuzzy overlapping
        maps are better at predicting behavioral variable than
        non-overlapping.}
    \label{fig:ts_extraction}
\end{figure}

\begin{figure}[h]
    \centering
    \includegraphics[width=\linewidth]{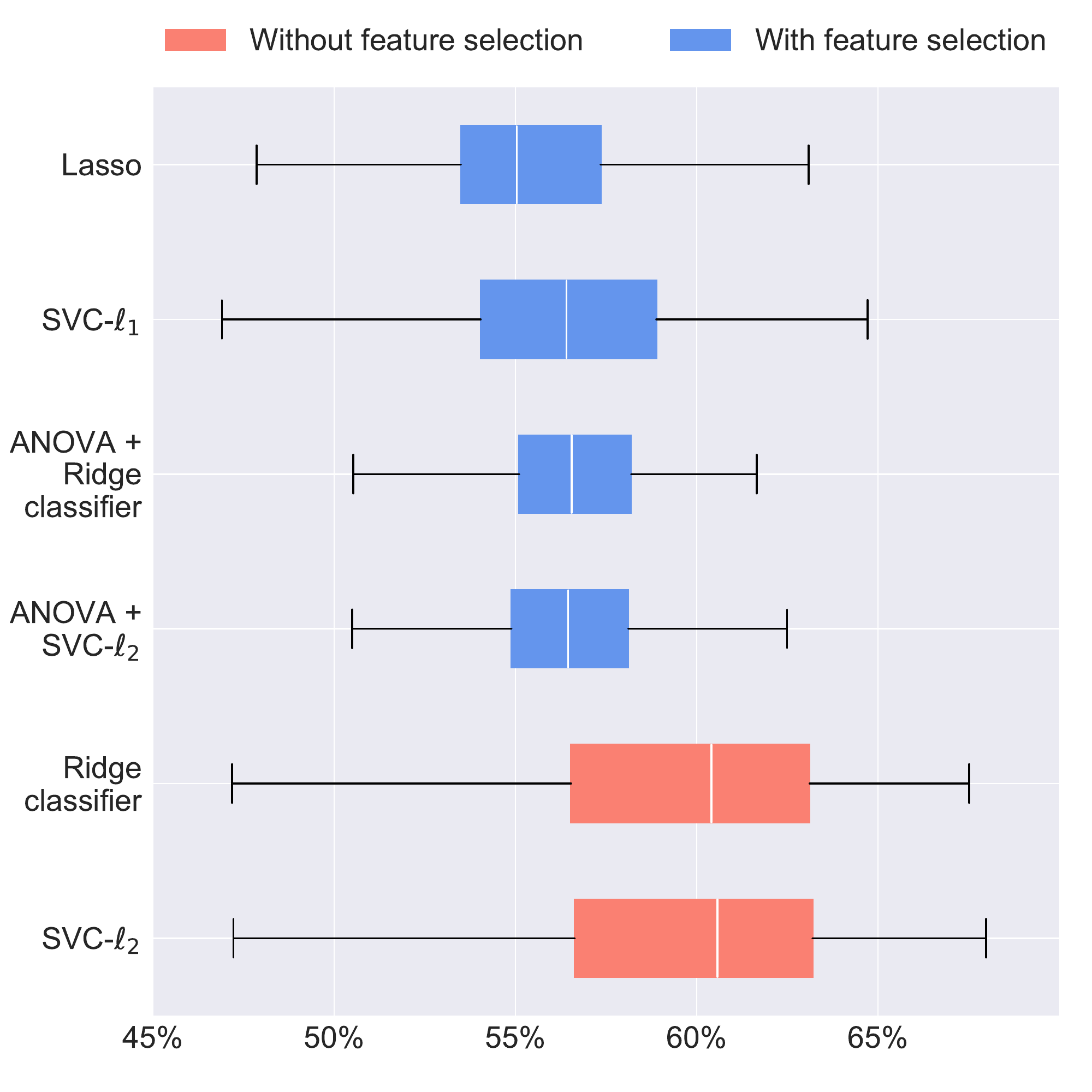}
    \caption{\textbf{Comparison of prediction results obtained using feature
        selection or sparse methods.}
    When selecting a reduced number of features, using either ANOVA to select
    10\% of the features, or sparsity inducing methods, we observe that the
    prediction accuracy drops compared to a classification using all features.
    This can be due to the fact that the connectivity matrices represent mixed
    features that should be handled together rather than seaprately. This may also be
    due to global effects in the connectivity: A global hypoconnectivity,
    as already observed in ASD patients, cannot be captured by $\ell_1$ 
    regularized classifiers.
\label{fig:with_or_wo_fs}}
\end{figure}

\begin{figure}[h]
    \includegraphics[width=\linewidth]{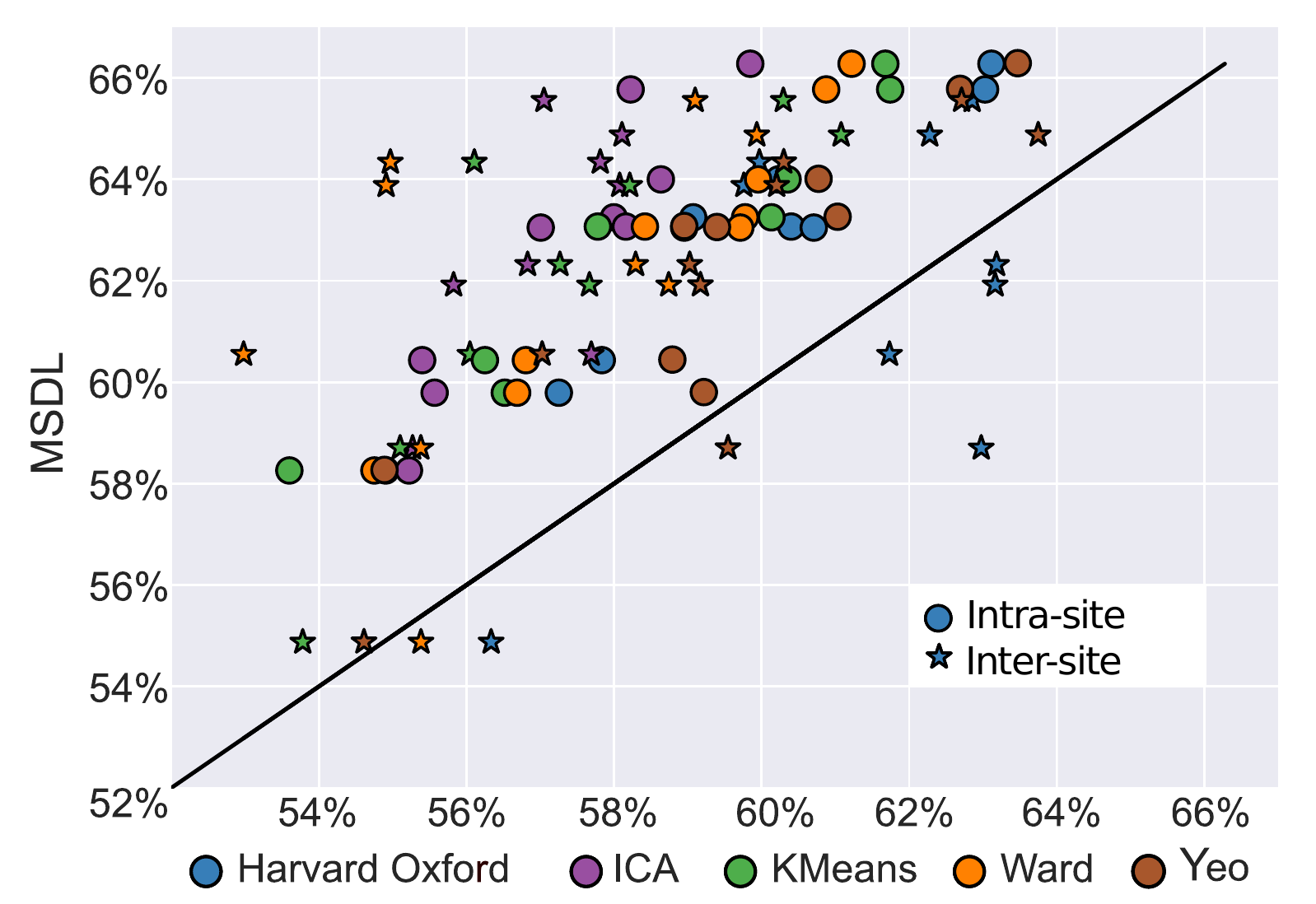}\\[.7cm]
    \includegraphics[width=\linewidth]{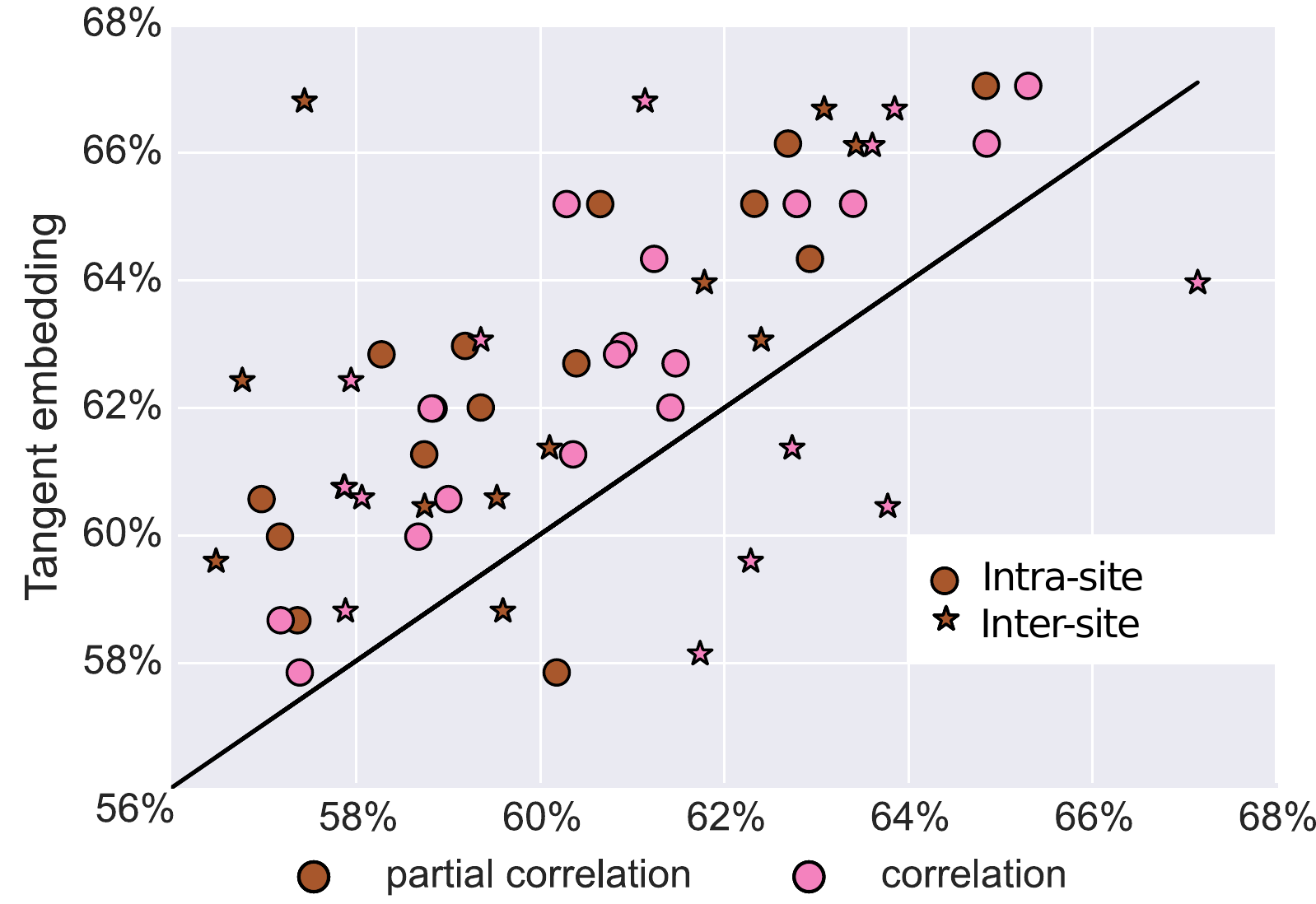}\\[.7cm]
    \includegraphics[width=\linewidth]{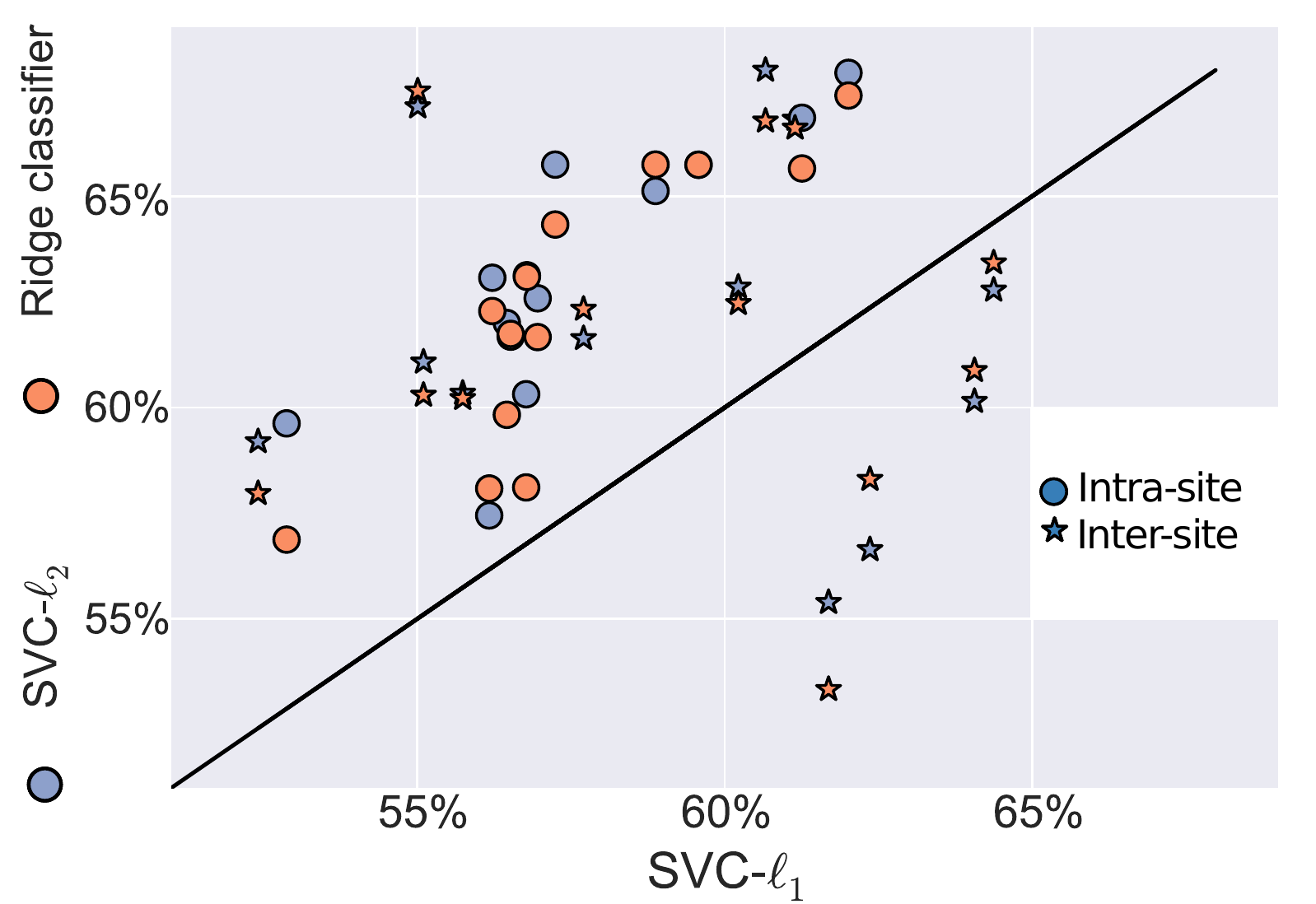}\\[.1cm]

    \caption{\textbf{Comparison of pipeline options.}
        Each plot compares the results of the best option against the other
        for each step. Points above the line means that the option show on
        y-axis is better than option on the x-axis. The 2 cross validation
        schemes are represented by circles and stars.
    \label{fig:full_comparison_plots}}
\end{figure}

\newcolumntype{R}{>{\raggedleft\arraybackslash}X}
\newcolumntype{G}{>{\raggedleft\arraybackslash\columncolor{gray!20}}X}
\newcolumntype{C}{>{\centering\arraybackslash}X}
\newcommand{\cc}[1]{\multicolumn{1}{c|}{#1}}
\newcommand{\cg}[1]{\multicolumn{1}{>{\columncolor{gray!20}}c|}{#1}}
\begin{table*}[h!]
    \footnotesize%
    \begin{tabularx}{\linewidth}{r|R|R|G|G|R|R|G|G|R|R|}
        & \multicolumn{2}{c|}{All subjects}
        & \multicolumn{2}{c|}{\makecell{Biggest \\ sites}}
        & \multicolumn{2}{c|}{\makecell{Right-handed \\ males}}
        & \multicolumn{2}{c|}{\makecell{Right-hand \\ males, 9-18}}
        & \multicolumn{2}{c|}{\makecell{Right-hand \\ males, 9-18, 3 sites}}
        \\\hline
        \cc{Sites} & \cc{ASD} & \cc{TC} & \cg{ASD} & \cg{TC} & \cc{ASD} &
        \cc{TC} & \cg{ASD} & \cg{TC} & \cc{ASD} & \cc{TC} \\\hline
    CALTECH    &   5 &  10 &     &     &     &     &     &     &     &     \\\hline
    CMU        &   6 &   5 &     &     &     &     &     &     &     &     \\\hline
    KKI        &  12 &  21 &  12 &  21 &   7 &  15 &   6 &  12 &     &     \\\hline
    LEUVEN\_1  &  14 &  14 &  14 &  14 &  13 &  13 &   1 &   1 &     &     \\\hline
    LEUVEN\_2  &  12 &  16 &  12 &  16 &   7 &  12 &   7 &  12 &     &     \\\hline
    MAX\_MUN   &  19 &  27 &  19 &  27 &  16 &  25 &   5 &   3 &     &     \\\hline
    NYU        &  74 &  98 &  74 &  98 &  60 &  70 &  36 &  40 &  36 &  40 \\\hline
    OHSU       &  12 &  13 &     &     &  10 &  13 &   9 &  11 &     &     \\\hline
    OLIN       &  14 &  14 &  14 &  14 &   8 &  11 &   7 &   7 &     &     \\\hline
    PITT       &  24 &  26 &  24 &  26 &  20 &  18 &  11 &  10 &     &     \\\hline
    SBL        &  12 &  14 &     &     &     &     &     &     &     &     \\\hline
    SDSU       &   8 &  19 &   8 &  19 &   7 &  10 &   7 &  10 &     &     \\\hline
    STANFORD   &  12 &  13 &     &     &   8 &   9 &   7 &   6 &     &     \\\hline
    TRINITY    &  19 &  25 &  19 &  25 &  18 &  24 &  12 &  14 &     &     \\\hline
    UCLA\_1    &  37 &  27 &  37 &  27 &  29 &  24 &  28 &  24 &  28 &  24 \\\hline
    UCLA\_2    &  11 &  10 &  11 &  10 &   9 &   8 &   9 &   8 &   9 &   8 \\\hline
    UM\_1      &  34 &  52 &  34 &  52 &  23 &  34 &  22 &  30 &  22 &  30 \\\hline
    UM\_2      &  13 &  21 &  13 &  21 &  12 &  19 &  12 &  17 &  12 &  17 \\\hline
    USM        &  43 &  24 &  43 &  24 &  37 &  22 &  12 &   7 &     &     \\\hline
    YALE       &  22 &  19 &  22 &  19 &   8 &  10 &   7 &  10 &     &     \\\hline
    Total      & 403 & 468 & 356 & 413 & 292 & 337 & 198 & 222 & 107 & 119 \\\hline
\end{tabularx}
\caption{Count of subjects per site and per subsamples.}
\label{tab:full_subsets}
\end{table*}

\begin{table*}[h!]
	\center
\footnotesize%
\rowcolors{2}{gray!15}{white}%
\begin{tabularx}{\linewidth}{p{1.2cm}|c|Y|Y|Y|Y|Y}
Cross\newline validation & Atlas
estimator&Right handed males 9-18 yo, 3 sites&Right handed males 9-18 yo&Right handed males&Biggest sites&All subjects\\
\hline
\cellcolor{white}&MSDL &$\mathbf{66.6\% \pm 5.4\%}$ &$\mathbf{65.8\% \pm 5.9\%}$ &$\mathbf{65.7\% \pm 4.9\%}$ &$\mathbf{67.9\% \pm 1.9\%}$ &$\mathbf{66.9\% \pm 2.7\%}$\\
\cellcolor{white}&Yeo &$60.9\% \pm 7.5\%$ &$62.3\% \pm 3.6\%$ &$64.7\% \pm 3.1\%$ &$64.7\% \pm 2.9\%$ &$66.9\% \pm 3.0\%$\\
\cellcolor{white}&Harvard Oxford &$63.6\% \pm 2.7\%$ &$62.1\% \pm 4.6\%$ &$64.8\% \pm 5.1\%$ &$64.8\% \pm 2.5\%$ &$66.4\% \pm 2.7\%$\\
\cellcolor{white}&ICA &$61.3\% \pm 8.7\%$ &$62.3\% \pm 3.0\%$ &$61.6\% \pm 4.3\%$ &$65.2\% \pm 3.1\%$ &$62.0\% \pm 3.4\%$\\
\cellcolor{white}&K-Means &$62.6\% \pm 5.9\%$ &$61.4\% \pm 5.4\%$ &$61.5\% \pm 3.0\%$ &$61.3\% \pm 3.4\%$ &$65.1\% \pm 1.8\%$\\
\multirow{-6}{*}{Intra-site}\cellcolor{white}&Ward &$63.2\% \pm 6.3\%$ &$60.1\% \pm 6.0\%$ &$62.2\% \pm 4.9\%$ &$64.6\% \pm 2.7\%$ &$63.7\% \pm 3.4\%$\\
\hline
\cellcolor{white}&MSDL& &$68.3\% \pm 7.6\%$ &$63.4\% \pm 6.3\%$ &$\mathbf{68.7\% \pm 9.3\%}$ &$\mathbf{66.8\% \pm 5.4\%}$\\
\cellcolor{white}&Yeo& &$\mathbf{69.7\% \pm 8.9\%}$ &$64.5\% \pm 10.3\%$ &$61.4\% \pm 7.9\%$ &$61.3\% \pm 7.2\%$\\
\cellcolor{white}&Harvard Oxford& &$68.1\% \pm 9.0\%$ &$\mathbf{65.1\% \pm 5.8\%}$ &$62.4\% \pm 5.4\%$ &$63.6\% \pm 6.2\%$\\
\cellcolor{white}&ICA& &$63.1\% \pm 9.9\%$ &$62.5\% \pm 7.8\%$ &$65.0\% \pm 4.8\%$ &$60.9\% \pm 5.2\%$\\
\cellcolor{white}&K-Means& &$62.8\% \pm 13.9\%$ &$61.5\% \pm 8.1\%$ &$61.9\% \pm 10.1\%$ &$60.3\% \pm 4.8\%$\\
\multirow{-6}{*}{Inter-site}\cellcolor{white}&Ward& &$62.4\% \pm 11.7\%$ &$59.8\% \pm 6.9\%$ &$63.4\% \pm 5.4\%$ &$63.1\% \pm 4.0\%$\\
\end{tabularx}
    \caption{\textbf{Average accuracy scores (along with their standard
        deviations) for top performing pipelines}.
        This table summarizes the scores for pipelines with best prediction
        accuracy for each atlas and subset using
        intra-site or inter-site prediction. Best results are shown in bold.}
    \label{tab:full_scores}
\end{table*}

\begin{table*}[h!]
	\center
\footnotesize%
\rowcolors{2}{gray!15}{white}%
\begin{tabularx}{\linewidth}{p{1.2cm}|c|Y|Y|Y|Y|Y}
Cross\newline validation & Atlas
estimator&Right handed males 9-18 yo, 3 sites&Right handed males 9-18 yo&Right handed males&Biggest sites&All subjects\\
\hline
\cellcolor{white}&MSDL &$\mathbf{73.6\% \pm 11.7\%}$ &$\mathbf{76.5\% \pm 25.1\%}$ &$62.4\% \pm 21.3\%$ &$\mathbf{59.3\% \pm 8.0\%}$ &$\mathbf{61.3\% \pm 4.8\%}$\\
\cellcolor{white}&Yeo &$68.2\% \pm 36.5\%$ &$63.0\% \pm 20.2\%$ &$58.5\% \pm 4.7\%$ &$59.0\% \pm 5.1\%$ &$59.6\% \pm 5.6\%$\\
\cellcolor{white}&Harvard Oxford &$67.7\% \pm 23.3\%$ &$61.4\% \pm 10.8\%$ &$59.5\% \pm 5.3\%$ &$57.4\% \pm 7.8\%$ &$59.5\% \pm 6.0\%$\\
\cellcolor{white}&ICA &$69.1\% \pm 12.1\%$ &$65.7\% \pm 24.5\%$ &$\mathbf{64.4\% \pm 7.9\%}$ &$57.3\% \pm 4.2\%$ &$\mathbf{61.3\% \pm 6.5\%}$\\
\cellcolor{white}&K-Means &$69.1\% \pm 12.3\%$ &$68.9\% \pm 22.7\%$ &$61.4\% \pm 5.3\%$ &$53.3\% \pm 5.9\%$ &$60.4\% \pm 6.6\%$\\
\multirow{-6}{*}{Intra-site}\cellcolor{white}&Ward &$66.4\% \pm 20.9\%$ &$63.2\% \pm 21.4\%$ &$62.7\% \pm 5.8\%$ &$53.3\% \pm 4.9\%$ &$59.8\% \pm 5.9\%$\\
\hline
\cellcolor{white}&MSDL& &$\mathbf{72.7\% \pm 19.0\%}$ &$67.5\% \pm 10.8\%$ &$\mathbf{67.9\% \pm 9.0\%}$ &$66.6\% \pm 14.7\%$\\
\cellcolor{white}&Yeo& &$65.4\% \pm 13.7\%$ &$67.1\% \pm 14.7\%$ &$57.2\% \pm 9.4\%$ &$59.1\% \pm 8.8\%$\\
\cellcolor{white}&Harvard Oxford& &$71.9\% \pm 22.5\%$ &$66.7\% \pm 21.2\%$ &$60.7\% \pm 12.1\%$ &$64.2\% \pm 9.0\%$\\
\cellcolor{white}&ICA& &$69.9\% \pm 17.1\%$ &$65.8\% \pm 13.0\%$ &$63.4\% \pm 12.2\%$ &$\mathbf{68.9\% \pm 6.7\%}$\\
\cellcolor{white}&K-Means& &$69.7\% \pm 13.2\%$ &$\mathbf{67.7\% \pm 13.7\%}$ &$60.3\% \pm 15.2\%$ &$67.7\% \pm 6.2\%$\\
\multirow{-6}{*}{Inter-site}\cellcolor{white}&Ward& &$66.0\% \pm 12.1\%$ &$63.3\% \pm 16.9\%$ &$59.5\% \pm 9.3\%$ &$65.4\% \pm 9.7\%$\\
\end{tabularx}
    \caption{\textbf{Average specificity scores
            (along with their standard deviations) for top performing pipelines}.
            This table summarizes the scores for pipelines with best prediction
            accuracy for each atlas and subset using
            intra-site or inter-site prediction. Best results are shown in bold.}
    \label{tab:full_scores}
\end{table*}

\begin{table*}[h!]
	\center
\footnotesize%
\rowcolors{2}{gray!15}{white}%
\begin{tabularx}{\linewidth}{p{1.2cm}|c|Y|Y|Y|Y|Y}
Cross\newline validation & Atlas
estimator&Right handed males 9-18 yo, 3 sites&Right handed males 9-18 yo&Right handed males&Biggest sites&All subjects\\
\hline
\cellcolor{white}&MSDL &$\mathbf{85.2\% \pm 14.6\%}$ &$\mathbf{86.0\% \pm 12.9\%}$ &$88.1\% \pm 12.6\%$ &$\mathbf{84.9\% \pm 11.7\%}$ &$\mathbf{90.9\% \pm 7.5\%}$\\
\cellcolor{white}&Yeo &$82.0\% \pm 13.2\%$ &$82.1\% \pm 8.4\%$ &$84.3\% \pm 12.6\%$ &$81.1\% \pm 6.1\%$ &$79.7\% \pm 3.4\%$\\
\cellcolor{white}&Harvard Oxford &$83.6\% \pm 14.3\%$ &$79.1\% \pm 20.4\%$ &$85.9\% \pm 10.6\%$ &$84.4\% \pm 3.0\%$ &$82.3\% \pm 2.7\%$\\
\cellcolor{white}&ICA &$80.4\% \pm 16.4\%$ &$77.7\% \pm 11.8\%$ &$\mathbf{88.4\% \pm 8.2\%}$ &$81.5\% \pm 8.2\%$ &$85.1\% \pm 4.4\%$\\
\cellcolor{white}&K-Means &$74.0\% \pm 13.1\%$ &$78.8\% \pm 20.3\%$ &$79.4\% \pm 19.0\%$ &$83.2\% \pm 5.7\%$ &$80.0\% \pm 12.2\%$\\
\multirow{-6}{*}{Intra-site}\cellcolor{white}&Ward &$78.0\% \pm 11.0\%$ &$77.0\% \pm 11.3\%$ &$80.0\% \pm 16.2\%$ &$80.4\% \pm 3.6\%$ &$80.4\% \pm 4.3\%$\\
\hline
\cellcolor{white}&MSDL& &$\mathbf{84.2\% \pm 10.5\%}$ &$\mathbf{85.4\% \pm 12.5\%}$ &$\mathbf{85.5\% \pm 13.6\%}$ &$\mathbf{100.0\% \pm 0.0\%}$\\
\cellcolor{white}&Yeo& &$80.4\% \pm 12.3\%$ &$81.5\% \pm 16.0\%$ &$76.5\% \pm 16.3\%$ &$80.0\% \pm 15.1\%$\\
\cellcolor{white}&Harvard Oxford& &$69.7\% \pm 16.5\%$ &$71.9\% \pm 10.9\%$ &$73.3\% \pm 13.6\%$ &$88.7\% \pm 18.9\%$\\
\cellcolor{white}&ICA& &$71.2\% \pm 16.2\%$ &$72.5\% \pm 7.6\%$ &$75.5\% \pm 12.9\%$ &$93.1\% \pm 15.0\%$\\
\cellcolor{white}&K-Means& &$75.4\% \pm 18.3\%$ &$67.2\% \pm 14.3\%$ &$68.4\% \pm 16.8\%$ &$94.3\% \pm 18.1\%$\\
\multirow{-6}{*}{Inter-site}\cellcolor{white}&Ward& &$70.3\% \pm 17.0\%$ &$63.2\% \pm 11.8\%$ &$74.1\% \pm 10.8\%$ &$82.6\% \pm 22.9\%$\\
\end{tabularx}
    \caption{\textbf{Average sensitivity scores
            (along with their standard deviations) for top performing pipelines}.
            This table summarizes the scores for pipelines with best prediction
            accuracy for each atlas and subset using
            intra-site or inter-site prediction. Best results are shown in bold.}
    \label{tab:full_scores}
\end{table*}

\begin{figure*}[h!]
    \begin{minipage}{.25\linewidth}
    \caption{\textbf{Learning curves.}
        Classification results obtained by varying the number of subjects in
        the training set while keeping a fixed testing set. The colored band
        represent the standard error of the prediction accuracy. A score increasing
        with the number of subjects, \ie an increasing curve, indicates that
        the addition of new subjects brings information.
    \label{fig:full_learning_curve}}
    \bigskip
    \end{minipage}%
    \hfill%
    \begin{minipage}{.7\linewidth}
    \includegraphics[width=\linewidth]{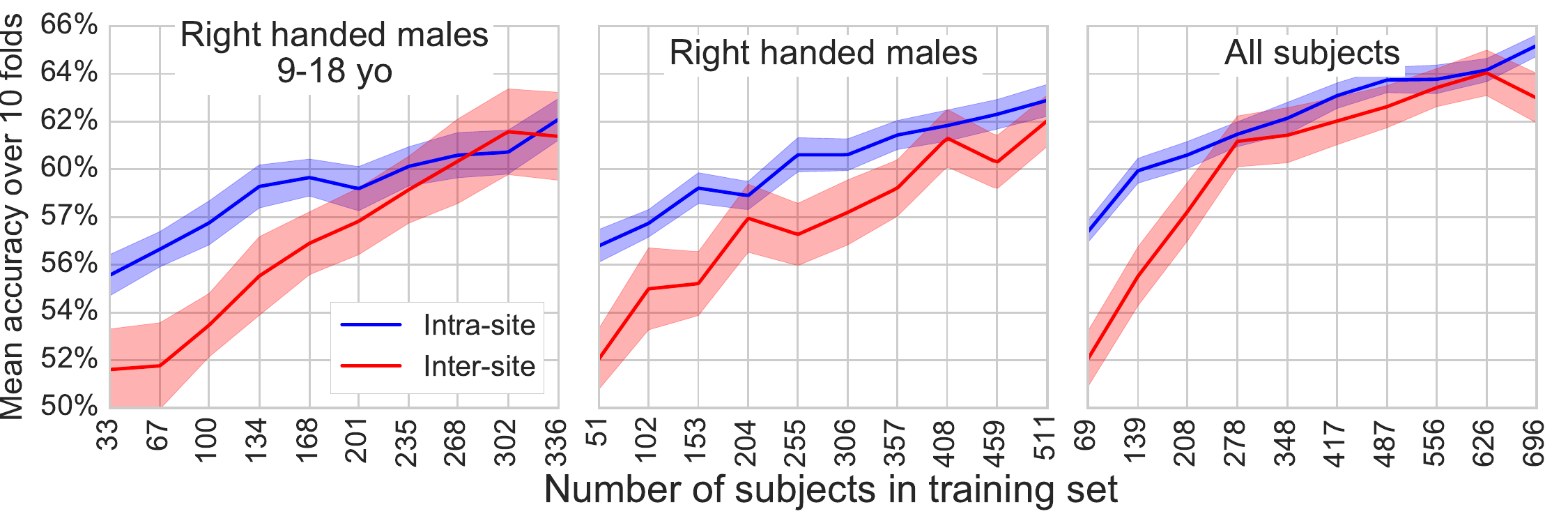}%
    \end{minipage}%
\end{figure*}

\begin{figure*}[tb]
    \begin{minipage}{.3\linewidth}
	\includegraphics[width=\linewidth]{img_2015/biomarkers/dmn.pdf}%
	\llap{%
	\raisebox{.38\linewidth}{\rlap{\textbf{\small\sffamily
		Default Mode Network}}%
	\hspace*{\linewidth}}}

	\includegraphics[width=\linewidth]{img_2015/biomarkers/dmn_legend.pdf}
	\includegraphics[width=\linewidth]{img_2015/biomarkers/dmn_conn.pdf}%
	\vspace*{-1.4em}

	\textbf{\emph{a}}
    \end{minipage}\hfill
    \begin{minipage}{.3\linewidth}
	\includegraphics[width=\linewidth]{img_2015/biomarkers/emo.pdf}%
	\llap{%
	\raisebox{.38\linewidth}{\rlap{\textbf{\small\sffamily
		Self-awareness ROIs}}%
	\hspace*{\linewidth}}}

	\includegraphics[width=\linewidth]{img_2015/biomarkers/emo_legend.pdf}
	\includegraphics[width=\linewidth]{img_2015/biomarkers/emo_conn.pdf}%
	\vspace*{-1.4em}

	\textbf{\emph{b}}
    \end{minipage}\hfill
    \begin{minipage}{.3\linewidth}
	\includegraphics[width=\linewidth]{img_2015/biomarkers/lang.pdf}%
	\llap{%
	\raisebox{.38\linewidth}{\rlap{\textbf{\small\sffamily Semantic ROIs}}%
	\hspace*{\linewidth}}}

	\includegraphics[width=\linewidth]{img_2015/biomarkers/lang_legend.pdf}
	\includegraphics[width=\linewidth]{img_2015/biomarkers/lang_conn.pdf}%
	\vspace*{-1.4em}

	\textbf{\emph{c}}
    \end{minipage}\\
    
    \vspace{14pt}
    \begin{minipage}{.3\linewidth}
        \vspace{-12pt}

    \caption{\textbf{Connections significantly non-zero in the predictive
        biomarkers distinguishing controls from ASD patients.}
            Red connections are stronger in controls and blue connections
            are stronger in ASD patients. Subfigures $a$, $b$, $c$, $d$ and $e$ reported
            intra-network difference. below is the consensus atlas
            extracted by selecting regions consistently extracted on 10
            subsets of ABIDE. Colors are random.
    \label{fig:full_networks}}
    \includegraphics[width=\linewidth]{img_2015/biomarkers/atlas.pdf}
    \end{minipage}\hfill
    \begin{minipage}{.3\linewidth}
	\includegraphics[width=\linewidth]{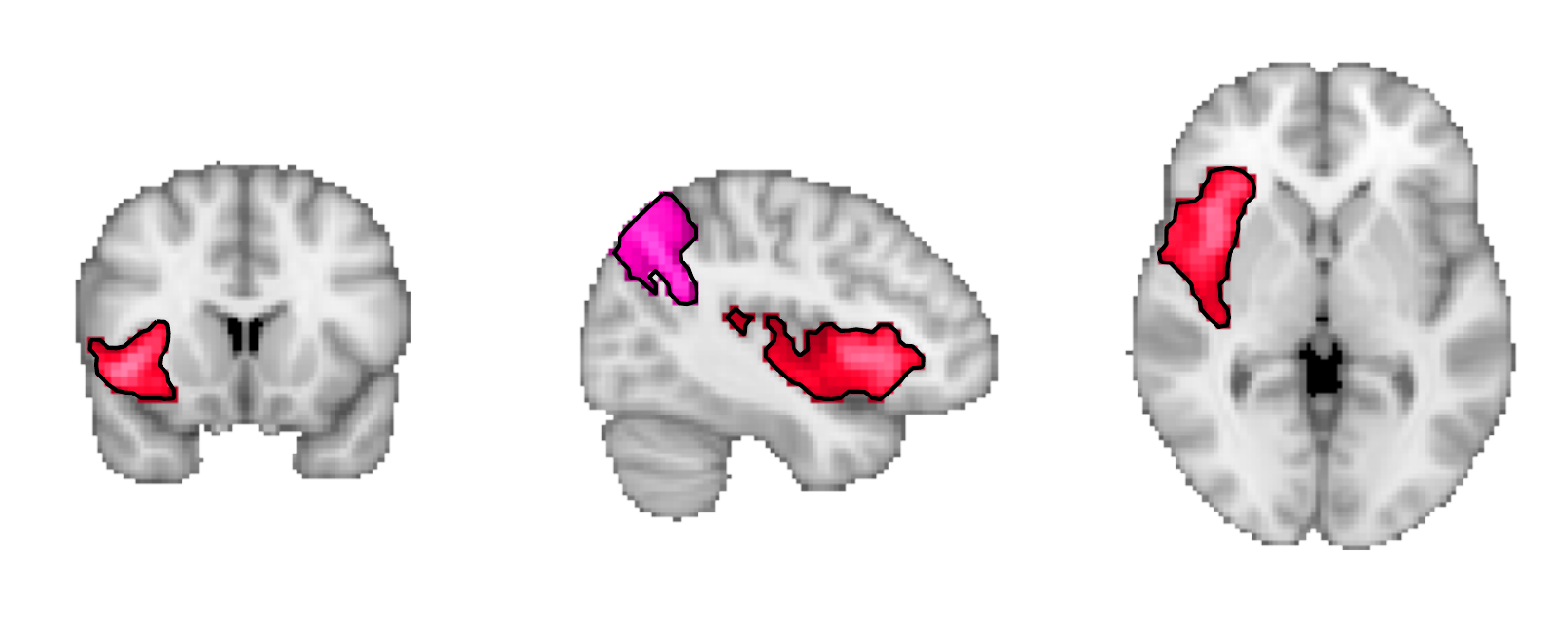}%
	\llap{%
	\raisebox{.38\linewidth}{\rlap{\textbf{\small\sffamily Broca's area and
        DMN}}%
	\hspace*{\linewidth}}}

	\includegraphics[width=\linewidth]{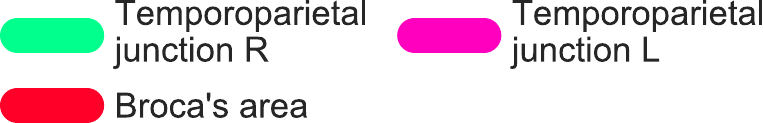}
	\includegraphics[width=\linewidth]{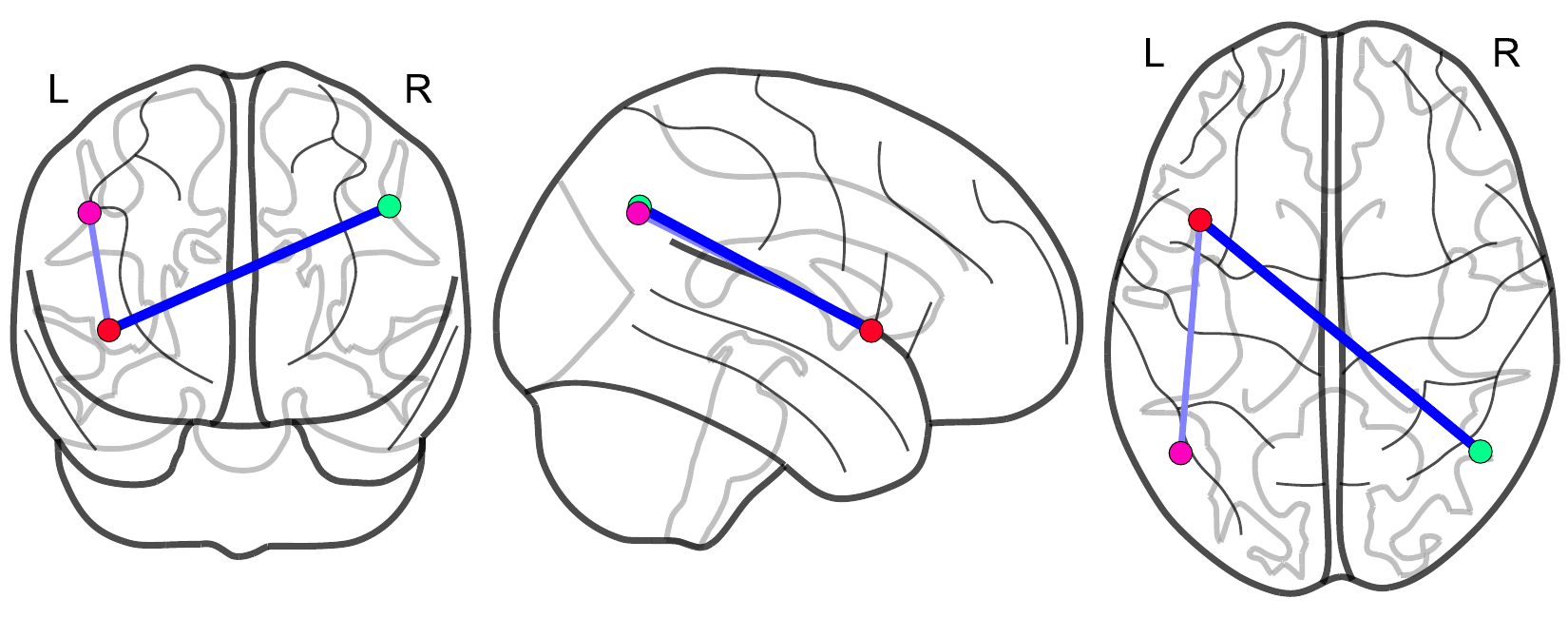}%
	\vspace*{-1.4em}
	\textbf{\emph{d}}
    \end{minipage}\hfill
    \begin{minipage}{.3\linewidth}
	\includegraphics[width=\linewidth]{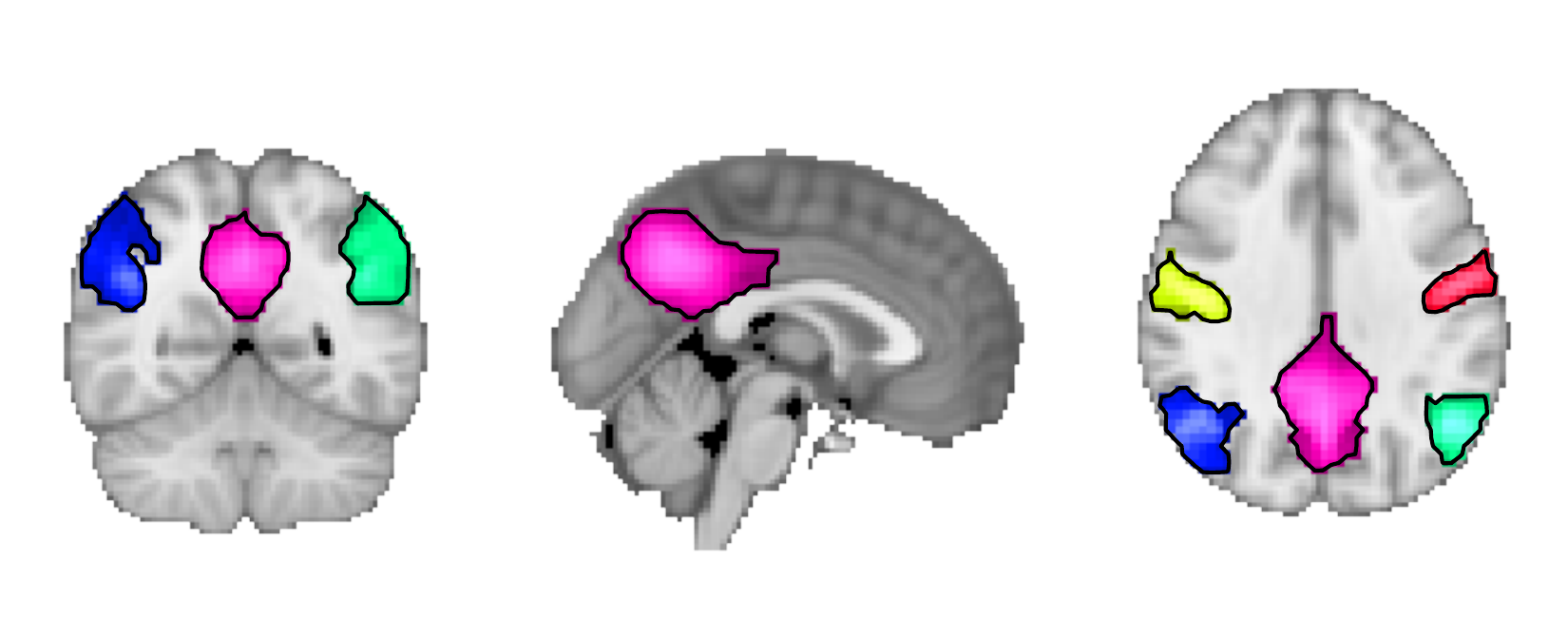}%
	\llap{%
	\raisebox{.38\linewidth}{\rlap{\textbf{\small\sffamily Primary Motor
        Cortex and DMN}}%
	\hspace*{\linewidth}}}

	\includegraphics[width=\linewidth]{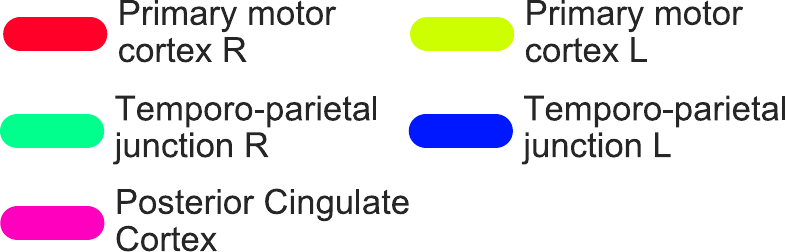}
	\includegraphics[width=\linewidth]{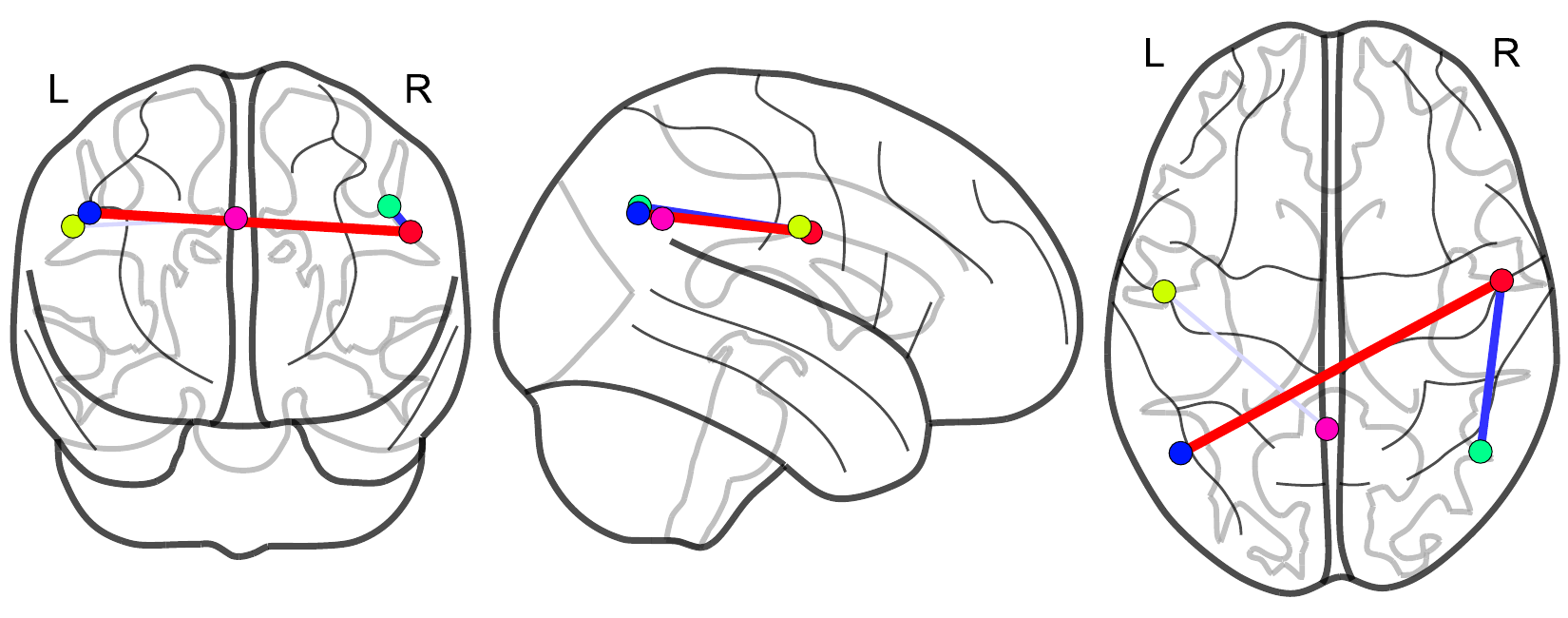}%
	\vspace*{-1.4em}
	\textbf{\emph{e}}
    \end{minipage}\\
\end{figure*}

%

\begin{figure*}[h!]
    \includegraphics[width=\linewidth]{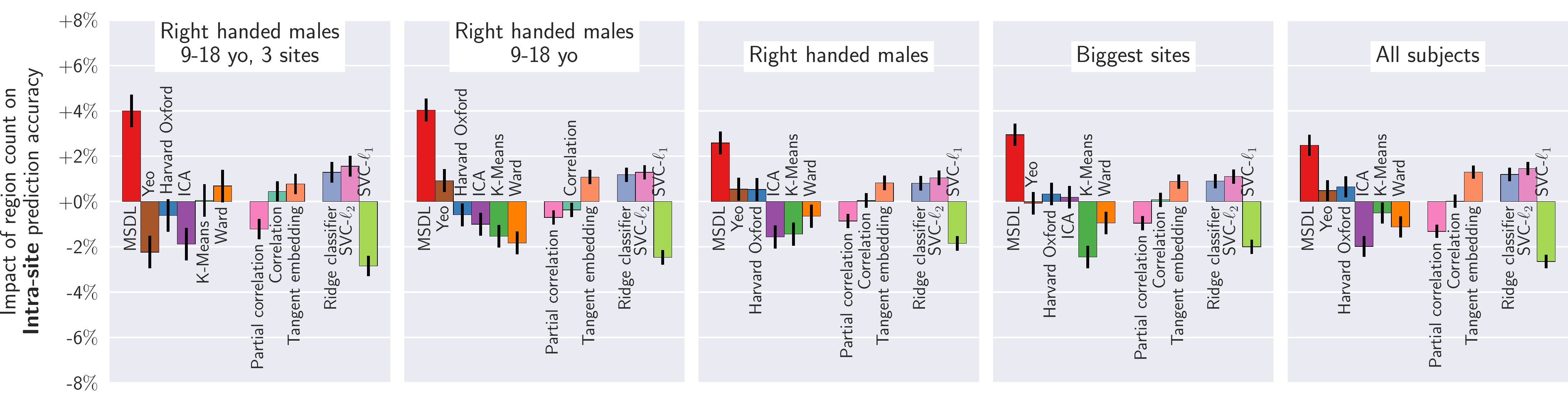}
    \begin{flushright}%
    \includegraphics[width=.808\linewidth]{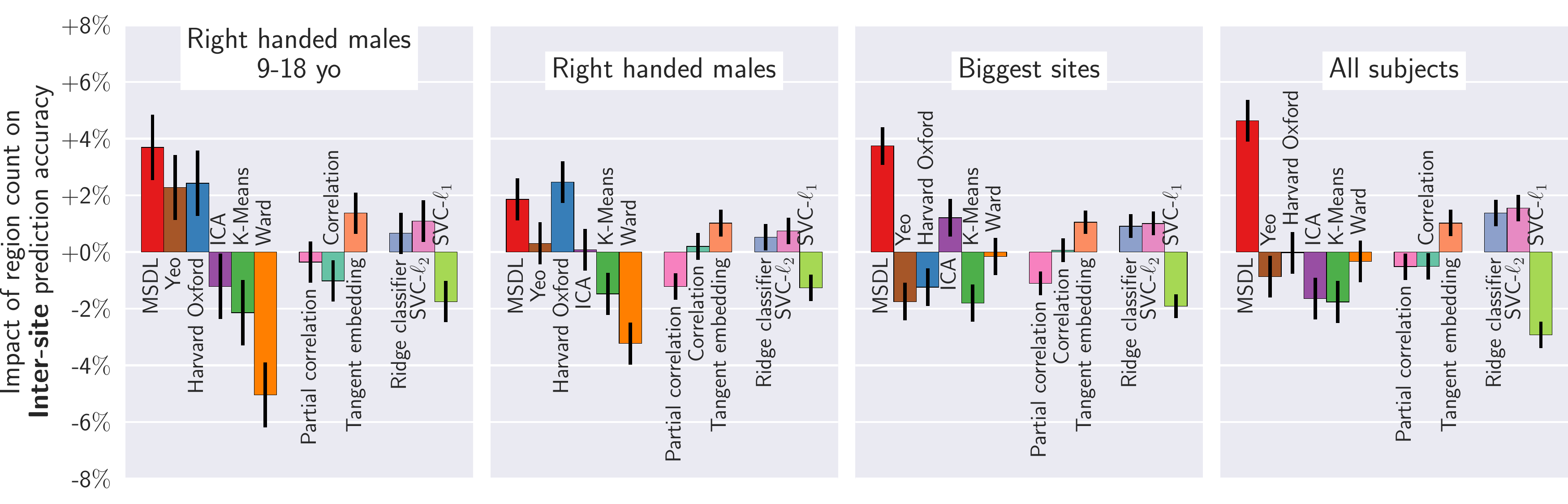}%
    \end{flushright}
    \caption{\textbf{Impact of pipeline steps on prediction.}
        Each plot represents the impact of each step of the pipeline for every
        subsets of ABIDE. In each figure, a block of bars represents a step of
        the pipeline (namely step 1, 3 and 4). Each bar represents the impact
        of the corresponding option on the prediction accuracy, relatively to
        the mean prediction. This effect is measured via a full-factorial
        ANOVA, analyzing the contribution of each step in a linear model. Each
        step of the pipeline is considered as a categorical variable. Error
        bars give the 95\% confidence interval. MSDL atlas extraction method
        gives significantly better results while reference atlases are
        slightly better than the mean. Among all matrix types, tangent
        embedding is the best on all ABIDE subsets. Finally, $l_2$ regularized
        classifiers are better than $l_1$ regularized.
    \label{fig:full_pipeline_steps}}
\end{figure*}

\begin{figure*}[h!]
    \begin{flushright}
    \begin{minipage}{.87\linewidth}%
    \includegraphics[width=\linewidth]{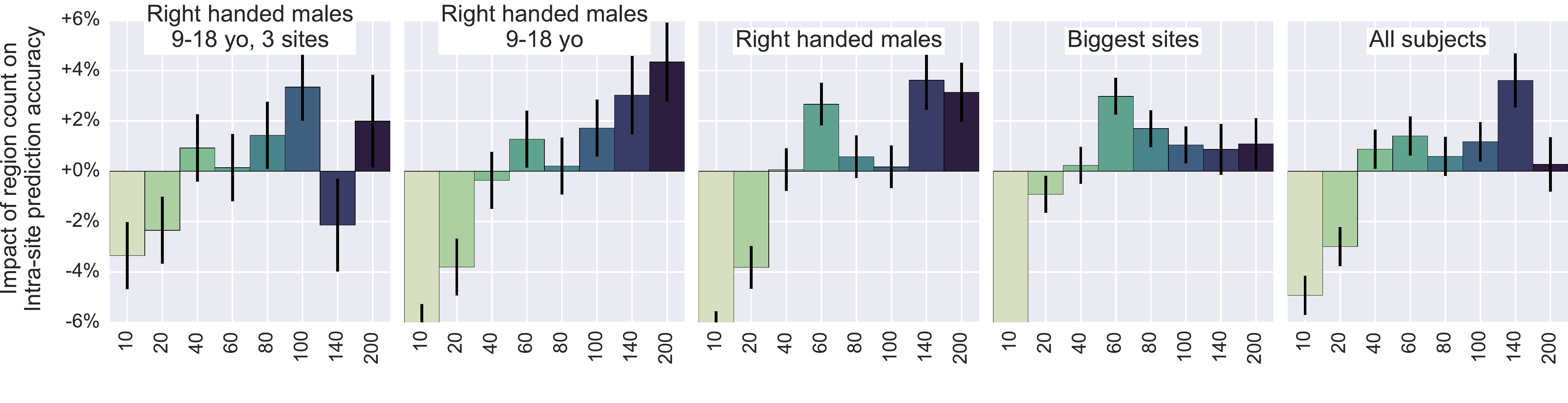}%
    \end{minipage}%
    \end{flushright}%
    \vspace{-.5cm}%
    \begin{minipage}{.28\linewidth}%
    \caption{\textbf{Impact of region number on prediction}: each
        bar indicates the impact of the number of regions on the prediction accuracy
        relatively to the mean of the prediction. These values are coefficients in a linear
        model explaining the best classification scores as function of
        the number of regions. Error bars give the 95\% confidence interval,
        computed by a full factorial ANOVA. Atlases containing more than 40 ROIs give
        better results in all settings.}
    \label{fig:full_region_count}
    \end{minipage}
    \hfill%
    \begin{minipage}{.7\linewidth}%
    \includegraphics[width=\linewidth]{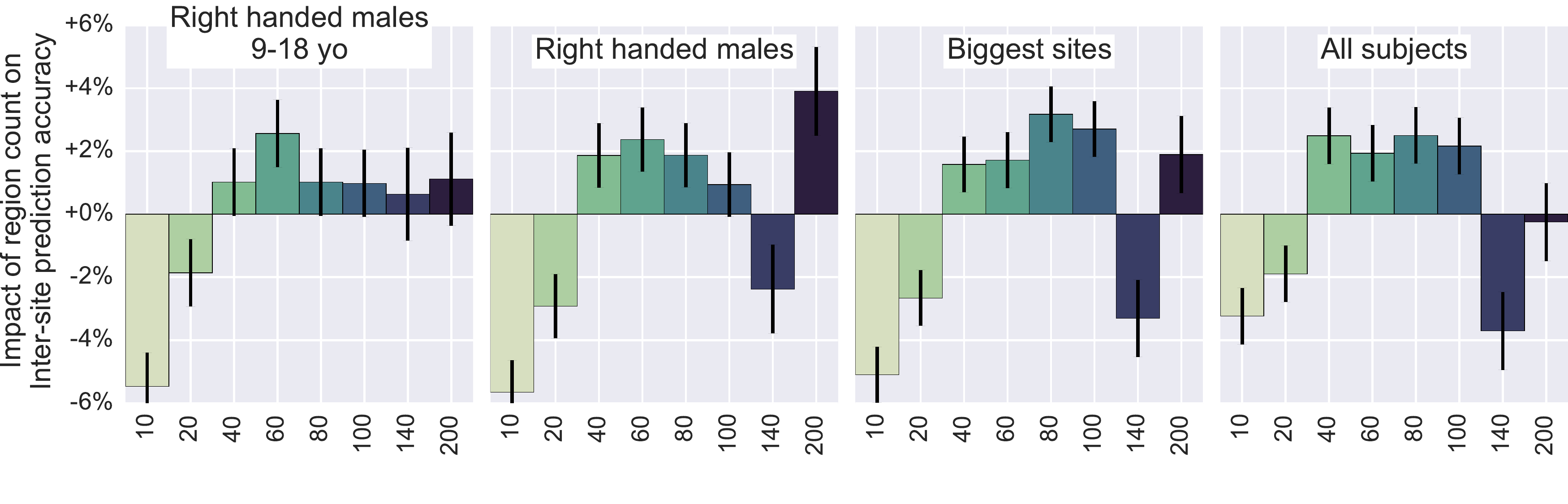}%
    \end{minipage}
\end{figure*}

\clearpage
\thispagestyle{empty}
\begin{table*}[h]
    \scriptsize
    \begin{tabular}{c@{\hskip 2pt}c@{\hskip 2pt}c@{\hskip 2pt}c@{\hskip 2pt}c@{\hskip 3pt}|@{\hskip 3pt}c@{\hskip 2pt}c@{\hskip 2pt}c@{\hskip 2pt}c@{\hskip 2pt}c@{\hskip 3pt}|@{\hskip 3pt}c@{\hskip 2pt}c@{\hskip 2pt}c@{\hskip 2pt}c@{\hskip 2pt}c@{\hskip 3pt}|@{\hskip 3pt}c@{\hskip 2pt}c@{\hskip 2pt}c@{\hskip 2pt}c@{\hskip 2pt}c@{\hskip 3pt}|@{\hskip 3pt}c@{\hskip 2pt}c@{\hskip 2pt}c@{\hskip 2pt}c@{\hskip 2pt}c@{\hskip 3pt}|@{\hskip 3pt}c@{\hskip 2pt}c@{\hskip 2pt}c@{\hskip 2pt}c@{\hskip 2pt}c@{\hskip 3pt}|@{\hskip 3pt}c@{\hskip 2pt}c@{\hskip 2pt}c@{\hskip 2pt}c@{\hskip 2pt}c@{\hskip 3pt}|@{\hskip 3pt}c@{\hskip 2pt}c@{\hskip 2pt}c@{\hskip 2pt}c@{\hskip 2pt}c@{\hskip 3pt}|@{\hskip 3pt}c@{\hskip 2pt}c@{\hskip 2pt}c@{\hskip 2pt}c@{\hskip 2pt}c@{\hskip 3pt}|@{\hskip 3pt}c@{\hskip 2pt}c@{\hskip 2pt}c@{\hskip 2pt}c@{\hskip 2pt}c} 
ID & 2 & 3 & 4 & 5 &ID & 2 & 3 & 4 & 5 &ID & 2 & 3 & 4 & 5 &ID & 2 & 3 & 4 & 5 &ID & 2 & 3 & 4 & 5 &ID & 2 & 3 & 4 & 5 &ID & 2 & 3 & 4 & 5 &ID & 2 & 3 & 4 & 5 &ID & 2 & 3 & 4 & 5 &ID & 2 & 3 & 4 & 5 \\
50003& $\bullet$& $\bullet$&  &  &50152&  & $\bullet$& $\bullet$&  &50290& $\bullet$& $\bullet$& $\bullet$& $\bullet$&50404& $\bullet$& $\bullet$& $\bullet$& $\bullet$&50531& $\bullet$&  &  &  &50728& $\bullet$& $\bullet$& $\bullet$&  &50988& $\bullet$& $\bullet$& $\bullet$& $\bullet$&51083& $\bullet$& $\bullet$&  &  &51194&  & $\bullet$& $\bullet$&  &51315& $\bullet$& $\bullet$& $\bullet$& $\bullet$\\
50004& $\bullet$& $\bullet$&  &  &50153&  & $\bullet$& $\bullet$&  &50291& $\bullet$& $\bullet$& $\bullet$& $\bullet$&50405& $\bullet$& $\bullet$& $\bullet$& $\bullet$&50532& $\bullet$& $\bullet$& $\bullet$&  &50730& $\bullet$&  &  &  &50989& $\bullet$& $\bullet$& $\bullet$& $\bullet$&51084& $\bullet$& $\bullet$& $\bullet$& $\bullet$&51197&  & $\bullet$& $\bullet$&  &51318& $\bullet$& $\bullet$&  &  \\
50005& $\bullet$&  &  &  &50156&  & $\bullet$& $\bullet$&  &50292& $\bullet$& $\bullet$& $\bullet$& $\bullet$&50406& $\bullet$& $\bullet$& $\bullet$& $\bullet$&50551& $\bullet$& $\bullet$& $\bullet$&  &50731& $\bullet$& $\bullet$& $\bullet$&  &50990& $\bullet$& $\bullet$& $\bullet$& $\bullet$&51085& $\bullet$& $\bullet$& $\bullet$& $\bullet$&51198&  & $\bullet$& $\bullet$&  &51319& $\bullet$& $\bullet$&  &  \\
50006& $\bullet$& $\bullet$& $\bullet$&  &50157&  & $\bullet$& $\bullet$&  &50293& $\bullet$& $\bullet$& $\bullet$& $\bullet$&50407& $\bullet$&  &  &  &50552& $\bullet$& $\bullet$& $\bullet$&  &50733& $\bullet$& $\bullet$& $\bullet$&  &50991& $\bullet$& $\bullet$& $\bullet$& $\bullet$&51086& $\bullet$& $\bullet$& $\bullet$& $\bullet$&51201& $\bullet$& $\bullet$& $\bullet$& $\bullet$&51320& $\bullet$& $\bullet$&  &  \\
50007& $\bullet$& $\bullet$& $\bullet$&  &50158&  & $\bullet$& $\bullet$&  &50294& $\bullet$& $\bullet$& $\bullet$& $\bullet$&50408& $\bullet$& $\bullet$& $\bullet$& $\bullet$&50555& $\bullet$&  &  &  &50735& $\bullet$&  &  &  &50992& $\bullet$& $\bullet$& $\bullet$& $\bullet$&51087& $\bullet$& $\bullet$& $\bullet$& $\bullet$&51202& $\bullet$& $\bullet$& $\bullet$& $\bullet$&51321& $\bullet$&  &  &  \\
50008& $\bullet$& $\bullet$&  &  &50159&  & $\bullet$& $\bullet$&  &50295& $\bullet$& $\bullet$& $\bullet$& $\bullet$&50410& $\bullet$& $\bullet$& $\bullet$& $\bullet$&50557& $\bullet$&  &  &  &50737& $\bullet$& $\bullet$& $\bullet$&  &50993& $\bullet$& $\bullet$& $\bullet$& $\bullet$&51088& $\bullet$& $\bullet$& $\bullet$& $\bullet$&51203& $\bullet$&  &  &  &51322& $\bullet$& $\bullet$&  &  \\
50010& $\bullet$& $\bullet$&  &  &50160&  & $\bullet$& $\bullet$&  &50297& $\bullet$& $\bullet$& $\bullet$& $\bullet$&50411& $\bullet$& $\bullet$& $\bullet$& $\bullet$&50558& $\bullet$&  &  &  &50738& $\bullet$& $\bullet$& $\bullet$&  &50994& $\bullet$& $\bullet$& $\bullet$& $\bullet$&51089& $\bullet$& $\bullet$& $\bullet$& $\bullet$&51204& $\bullet$& $\bullet$& $\bullet$& $\bullet$&51323& $\bullet$& $\bullet$&  &  \\
50011& $\bullet$&  &  &  &50161&  & $\bullet$& $\bullet$&  &50298& $\bullet$& $\bullet$& $\bullet$& $\bullet$&50412& $\bullet$& $\bullet$& $\bullet$& $\bullet$&50561& $\bullet$& $\bullet$& $\bullet$&  &50739& $\bullet$& $\bullet$& $\bullet$&  &50995& $\bullet$& $\bullet$& $\bullet$& $\bullet$&51090& $\bullet$& $\bullet$& $\bullet$& $\bullet$&51205& $\bullet$& $\bullet$& $\bullet$& $\bullet$&51325& $\bullet$&  &  &  \\
50012& $\bullet$& $\bullet$&  &  &50162&  & $\bullet$&  &  &50300& $\bullet$&  &  &  &50413& $\bullet$& $\bullet$& $\bullet$& $\bullet$&50563& $\bullet$&  &  &  &50740& $\bullet$& $\bullet$& $\bullet$&  &50996& $\bullet$& $\bullet$& $\bullet$& $\bullet$&51091& $\bullet$&  &  &  &51206& $\bullet$& $\bullet$& $\bullet$& $\bullet$&51326& $\bullet$& $\bullet$&  &  \\
50013& $\bullet$& $\bullet$& $\bullet$&  &50163&  & $\bullet$& $\bullet$&  &50301& $\bullet$& $\bullet$& $\bullet$& $\bullet$&50414& $\bullet$&  &  &  &50565& $\bullet$&  &  &  &50741& $\bullet$& $\bullet$& $\bullet$&  &50997& $\bullet$& $\bullet$& $\bullet$& $\bullet$&51093& $\bullet$& $\bullet$& $\bullet$& $\bullet$&51207& $\bullet$&  &  &  &51327& $\bullet$& $\bullet$&  &  \\
50014& $\bullet$& $\bullet$& $\bullet$&  &50164&  & $\bullet$&  &  &50302& $\bullet$&  &  &  &50415& $\bullet$& $\bullet$& $\bullet$& $\bullet$&50568& $\bullet$& $\bullet$& $\bullet$&  &50742& $\bullet$& $\bullet$& $\bullet$&  &50999& $\bullet$& $\bullet$& $\bullet$& $\bullet$&51094& $\bullet$& $\bullet$& $\bullet$& $\bullet$&51208& $\bullet$& $\bullet$& $\bullet$& $\bullet$&51328& $\bullet$& $\bullet$&  &  \\
50015& $\bullet$& $\bullet$& $\bullet$&  &50167&  & $\bullet$& $\bullet$&  &50304& $\bullet$& $\bullet$& $\bullet$& $\bullet$&50416& $\bullet$& $\bullet$& $\bullet$& $\bullet$&50569& $\bullet$&  &  &  &50743& $\bullet$&  &  &  &51000& $\bullet$& $\bullet$& $\bullet$& $\bullet$&51095& $\bullet$& $\bullet$& $\bullet$& $\bullet$&51210& $\bullet$& $\bullet$& $\bullet$& $\bullet$&51329& $\bullet$& $\bullet$& $\bullet$&  \\
50016& $\bullet$& $\bullet$&  &  &50168&  & $\bullet$& $\bullet$&  &50308& $\bullet$&  &  &  &50417& $\bullet$& $\bullet$& $\bullet$& $\bullet$&50570& $\bullet$& $\bullet$& $\bullet$&  &50744& $\bullet$&  &  &  &51001& $\bullet$& $\bullet$& $\bullet$& $\bullet$&51096& $\bullet$& $\bullet$& $\bullet$& $\bullet$&51211& $\bullet$& $\bullet$& $\bullet$& $\bullet$&51330& $\bullet$& $\bullet$&  &  \\
50020& $\bullet$& $\bullet$&  &  &50169&  & $\bullet$& $\bullet$&  &50310& $\bullet$& $\bullet$& $\bullet$& $\bullet$&50418& $\bullet$& $\bullet$& $\bullet$& $\bullet$&50571& $\bullet$& $\bullet$& $\bullet$&  &50745& $\bullet$& $\bullet$& $\bullet$&  &51002& $\bullet$& $\bullet$&  &  &51097& $\bullet$& $\bullet$& $\bullet$& $\bullet$&51212& $\bullet$& $\bullet$& $\bullet$& $\bullet$&51331& $\bullet$& $\bullet$&  &  \\
50022& $\bullet$& $\bullet$& $\bullet$&  &50170&  & $\bullet$& $\bullet$&  &50312& $\bullet$& $\bullet$& $\bullet$& $\bullet$&50419& $\bullet$& $\bullet$& $\bullet$& $\bullet$&50572& $\bullet$&  &  &  &50748& $\bullet$& $\bullet$& $\bullet$&  &51003& $\bullet$& $\bullet$&  &  &51098& $\bullet$& $\bullet$& $\bullet$& $\bullet$&51214& $\bullet$& $\bullet$& $\bullet$& $\bullet$&51332& $\bullet$& $\bullet$&  &  \\
50023& $\bullet$&  &  &  &50171&  & $\bullet$& $\bullet$&  &50314& $\bullet$& $\bullet$& $\bullet$& $\bullet$&50421& $\bullet$& $\bullet$& $\bullet$& $\bullet$&50573& $\bullet$& $\bullet$& $\bullet$&  &50749& $\bullet$&  &  &  &51006& $\bullet$& $\bullet$& $\bullet$& $\bullet$&51099& $\bullet$& $\bullet$& $\bullet$& $\bullet$&51215& $\bullet$&  &  &  &51333& $\bullet$& $\bullet$&  &  \\
50024& $\bullet$& $\bullet$&  &  &50182& $\bullet$& $\bullet$& $\bullet$&  &50315& $\bullet$& $\bullet$& $\bullet$& $\bullet$&50422& $\bullet$& $\bullet$& $\bullet$& $\bullet$&50574& $\bullet$& $\bullet$& $\bullet$&  &50750& $\bullet$&  &  &  &51007& $\bullet$& $\bullet$& $\bullet$& $\bullet$&51100& $\bullet$& $\bullet$& $\bullet$& $\bullet$&51216& $\bullet$& $\bullet$& $\bullet$& $\bullet$&51334& $\bullet$& $\bullet$&  &  \\
50025& $\bullet$& $\bullet$&  &  &50183& $\bullet$& $\bullet$& $\bullet$&  &50318& $\bullet$&  &  &  &50424& $\bullet$& $\bullet$& $\bullet$& $\bullet$&50575& $\bullet$& $\bullet$& $\bullet$&  &50751& $\bullet$& $\bullet$& $\bullet$&  &51008& $\bullet$& $\bullet$& $\bullet$& $\bullet$&51101& $\bullet$& $\bullet$& $\bullet$& $\bullet$&51217& $\bullet$& $\bullet$& $\bullet$& $\bullet$&51335& $\bullet$& $\bullet$&  &  \\
50026& $\bullet$& $\bullet$& $\bullet$&  &50184& $\bullet$&  &  &  &50319& $\bullet$&  &  &  &50425& $\bullet$& $\bullet$& $\bullet$& $\bullet$&50576& $\bullet$&  &  &  &50752& $\bullet$& $\bullet$& $\bullet$&  &51009& $\bullet$& $\bullet$& $\bullet$& $\bullet$&51102& $\bullet$& $\bullet$& $\bullet$& $\bullet$&51218& $\bullet$& $\bullet$& $\bullet$& $\bullet$&51336& $\bullet$& $\bullet$&  &  \\
50027& $\bullet$& $\bullet$& $\bullet$&  &50186& $\bullet$& $\bullet$& $\bullet$&  &50320& $\bullet$& $\bullet$&  &  &50426& $\bullet$& $\bullet$& $\bullet$& $\bullet$&50577& $\bullet$&  &  &  &50754& $\bullet$& $\bullet$& $\bullet$&  &51010& $\bullet$&  &  &  &51103& $\bullet$& $\bullet$& $\bullet$& $\bullet$&51219& $\bullet$&  &  &  &51338& $\bullet$& $\bullet$&  &  \\
50028& $\bullet$& $\bullet$& $\bullet$&  &50187& $\bullet$& $\bullet$& $\bullet$&  &50321& $\bullet$&  &  &  &50427& $\bullet$& $\bullet$& $\bullet$& $\bullet$&50578& $\bullet$& $\bullet$& $\bullet$&  &50755& $\bullet$& $\bullet$& $\bullet$&  &51011& $\bullet$& $\bullet$&  &  &51104& $\bullet$& $\bullet$& $\bullet$& $\bullet$&51220& $\bullet$& $\bullet$& $\bullet$& $\bullet$&51339& $\bullet$& $\bullet$&  &  \\
50030& $\bullet$& $\bullet$&  &  &50188& $\bullet$& $\bullet$& $\bullet$&  &50324& $\bullet$& $\bullet$& $\bullet$& $\bullet$&50428& $\bullet$& $\bullet$& $\bullet$& $\bullet$&50601& $\bullet$& $\bullet$& $\bullet$&  &50756& $\bullet$& $\bullet$& $\bullet$&  &51012& $\bullet$& $\bullet$& $\bullet$& $\bullet$&51105& $\bullet$& $\bullet$& $\bullet$& $\bullet$&51221& $\bullet$&  &  &  &51340& $\bullet$& $\bullet$&  &  \\
50031& $\bullet$& $\bullet$& $\bullet$&  &50189& $\bullet$& $\bullet$& $\bullet$&  &50325& $\bullet$&  &  &  &50433& $\bullet$& $\bullet$&  &  &50602& $\bullet$& $\bullet$& $\bullet$&  &50757& $\bullet$&  &  &  &51013& $\bullet$& $\bullet$&  &  &51106& $\bullet$& $\bullet$& $\bullet$& $\bullet$&51222& $\bullet$& $\bullet$& $\bullet$& $\bullet$&51341& $\bullet$& $\bullet$&  &  \\
50032& $\bullet$& $\bullet$&  &  &50190& $\bullet$& $\bullet$& $\bullet$&  &50327& $\bullet$& $\bullet$& $\bullet$& $\bullet$&50434& $\bullet$& $\bullet$&  &  &50603& $\bullet$&  &  &  &50772& $\bullet$& $\bullet$& $\bullet$&  &51014& $\bullet$& $\bullet$& $\bullet$& $\bullet$&51107& $\bullet$& $\bullet$& $\bullet$& $\bullet$&51223& $\bullet$& $\bullet$& $\bullet$& $\bullet$&51342& $\bullet$& $\bullet$&  &  \\
50033& $\bullet$& $\bullet$& $\bullet$&  &50193& $\bullet$& $\bullet$& $\bullet$&  &50329& $\bullet$& $\bullet$& $\bullet$& $\bullet$&50435& $\bullet$& $\bullet$& $\bullet$&  &50604& $\bullet$&  &  &  &50773& $\bullet$& $\bullet$& $\bullet$&  &51015& $\bullet$& $\bullet$&  &  &51109& $\bullet$& $\bullet$& $\bullet$& $\bullet$&51224& $\bullet$& $\bullet$& $\bullet$& $\bullet$&51343& $\bullet$& $\bullet$&  &  \\
50034& $\bullet$& $\bullet$& $\bullet$&  &50194& $\bullet$& $\bullet$& $\bullet$&  &50330& $\bullet$& $\bullet$& $\bullet$& $\bullet$&50436& $\bullet$& $\bullet$& $\bullet$&  &50606& $\bullet$&  &  &  &50774& $\bullet$& $\bullet$& $\bullet$&  &51016& $\bullet$& $\bullet$&  &  &51110& $\bullet$& $\bullet$& $\bullet$& $\bullet$&51225& $\bullet$& $\bullet$& $\bullet$& $\bullet$&51344& $\bullet$& $\bullet$&  &  \\
50035& $\bullet$&  &  &  &50195& $\bullet$&  &  &  &50331& $\bullet$& $\bullet$& $\bullet$& $\bullet$&50437& $\bullet$& $\bullet$& $\bullet$&  &50607& $\bullet$&  &  &  &50775& $\bullet$& $\bullet$& $\bullet$&  &51017& $\bullet$& $\bullet$&  &  &51111& $\bullet$& $\bullet$& $\bullet$& $\bullet$&51226& $\bullet$&  &  &  &51345& $\bullet$& $\bullet$&  &  \\
50036& $\bullet$&  &  &  &50196& $\bullet$& $\bullet$& $\bullet$&  &50332& $\bullet$& $\bullet$& $\bullet$& $\bullet$&50438& $\bullet$& $\bullet$& $\bullet$&  &50608& $\bullet$& $\bullet$& $\bullet$&  &50776& $\bullet$& $\bullet$& $\bullet$&  &51018& $\bullet$& $\bullet$&  &  &51112& $\bullet$& $\bullet$&  &  &51228& $\bullet$&  &  &  &51346& $\bullet$& $\bullet$&  &  \\
50037& $\bullet$& $\bullet$&  &  &50198& $\bullet$& $\bullet$& $\bullet$&  &50333& $\bullet$& $\bullet$& $\bullet$& $\bullet$&50439& $\bullet$& $\bullet$&  &  &50612& $\bullet$& $\bullet$& $\bullet$&  &50777& $\bullet$& $\bullet$&  &  &51019& $\bullet$& $\bullet$&  &  &51113& $\bullet$& $\bullet$&  &  &51229& $\bullet$& $\bullet$& $\bullet$& $\bullet$&51347& $\bullet$& $\bullet$&  &  \\
50038& $\bullet$&  &  &  &50199& $\bullet$& $\bullet$& $\bullet$&  &50334& $\bullet$& $\bullet$& $\bullet$& $\bullet$&50440& $\bullet$& $\bullet$&  &  &50613& $\bullet$& $\bullet$& $\bullet$&  &50778& $\bullet$&  &  &  &51020& $\bullet$& $\bullet$&  &  &51114& $\bullet$& $\bullet$&  &  &51230& $\bullet$&  &  &  &51349& $\bullet$& $\bullet$& $\bullet$&  \\
50039& $\bullet$&  &  &  &50200& $\bullet$& $\bullet$& $\bullet$&  &50335& $\bullet$& $\bullet$& $\bullet$& $\bullet$&50441& $\bullet$& $\bullet$&  &  &50614& $\bullet$&  &  &  &50780& $\bullet$&  &  &  &51021& $\bullet$& $\bullet$&  &  &51116& $\bullet$& $\bullet$&  &  &51231& $\bullet$& $\bullet$& $\bullet$& $\bullet$&51350& $\bullet$& $\bullet$& $\bullet$&  \\
50040& $\bullet$& $\bullet$&  &  &50201& $\bullet$&  &  &  &50336& $\bullet$&  &  &  &50442& $\bullet$& $\bullet$&  &  &50615& $\bullet$&  &  &  &50781& $\bullet$& $\bullet$& $\bullet$&  &51023& $\bullet$& $\bullet$&  &  &51117& $\bullet$& $\bullet$&  &  &51234& $\bullet$& $\bullet$& $\bullet$& $\bullet$&51351& $\bullet$& $\bullet$& $\bullet$&  \\
50041& $\bullet$& $\bullet$&  &  &50202& $\bullet$& $\bullet$& $\bullet$&  &50337& $\bullet$& $\bullet$& $\bullet$& $\bullet$&50443& $\bullet$& $\bullet$& $\bullet$&  &50616& $\bullet$& $\bullet$& $\bullet$&  &50782& $\bullet$& $\bullet$& $\bullet$&  &51024& $\bullet$& $\bullet$&  &  &51118& $\bullet$& $\bullet$&  &  &51235& $\bullet$& $\bullet$& $\bullet$& $\bullet$&51354& $\bullet$& $\bullet$& $\bullet$&  \\
50042& $\bullet$& $\bullet$&  &  &50203& $\bullet$& $\bullet$& $\bullet$&  &50338& $\bullet$&  &  &  &50444& $\bullet$& $\bullet$&  &  &50619& $\bullet$&  &  &  &50783& $\bullet$&  &  &  &51025& $\bullet$& $\bullet$&  &  &51122& $\bullet$& $\bullet$& $\bullet$& $\bullet$&51236& $\bullet$& $\bullet$& $\bullet$& $\bullet$&51356& $\bullet$& $\bullet$& $\bullet$&  \\
50043& $\bullet$& $\bullet$& $\bullet$&  &50204& $\bullet$&  &  &  &50339& $\bullet$& $\bullet$& $\bullet$& $\bullet$&50445& $\bullet$& $\bullet$&  &  &50620& $\bullet$&  &  &  &50786& $\bullet$& $\bullet$&  &  &51026& $\bullet$& $\bullet$&  &  &51123& $\bullet$& $\bullet$& $\bullet$& $\bullet$&51237& $\bullet$& $\bullet$& $\bullet$& $\bullet$&51357& $\bullet$& $\bullet$& $\bullet$&  \\
50044& $\bullet$& $\bullet$& $\bullet$&  &50205& $\bullet$&  &  &  &50340& $\bullet$&  &  &  &50446& $\bullet$& $\bullet$&  &  &50621& $\bullet$&  &  &  &50790& $\bullet$&  &  &  &51027& $\bullet$& $\bullet$&  &  &51124& $\bullet$& $\bullet$& $\bullet$& $\bullet$&51239& $\bullet$& $\bullet$& $\bullet$& $\bullet$&51359& $\bullet$&  &  &  \\
50045& $\bullet$&  &  &  &50206& $\bullet$&  &  &  &50341& $\bullet$&  &  &  &50447& $\bullet$& $\bullet$& $\bullet$&  &50622& $\bullet$& $\bullet$& $\bullet$&  &50791& $\bullet$&  &  &  &51028& $\bullet$& $\bullet$&  &  &51126& $\bullet$& $\bullet$& $\bullet$& $\bullet$&51240& $\bullet$& $\bullet$& $\bullet$& $\bullet$&51360& $\bullet$& $\bullet$&  &  \\
50046& $\bullet$& $\bullet$&  &  &50208& $\bullet$&  &  &  &50342& $\bullet$& $\bullet$& $\bullet$& $\bullet$&50448& $\bullet$&  &  &  &50623& $\bullet$&  &  &  &50792& $\bullet$&  &  &  &51029& $\bullet$& $\bullet$&  &  &51127& $\bullet$& $\bullet$& $\bullet$& $\bullet$&51241& $\bullet$& $\bullet$& $\bullet$& $\bullet$&51361& $\bullet$& $\bullet$& $\bullet$&  \\
50047& $\bullet$& $\bullet$& $\bullet$&  &50210& $\bullet$&  &  &  &50343& $\bullet$&  &  &  &50449& $\bullet$& $\bullet$&  &  &50624& $\bullet$&  &  &  &50796& $\bullet$&  &  &  &51030& $\bullet$& $\bullet$&  &  &51128& $\bullet$& $\bullet$& $\bullet$& $\bullet$&51248& $\bullet$& $\bullet$& $\bullet$& $\bullet$&51362& $\bullet$& $\bullet$&  &  \\
50048& $\bullet$& $\bullet$& $\bullet$&  &50213& $\bullet$&  &  &  &50344& $\bullet$& $\bullet$&  &  &50453& $\bullet$&  &  &  &50625& $\bullet$& $\bullet$&  &  &50797& $\bullet$& $\bullet$& $\bullet$&  &51032& $\bullet$& $\bullet$&  &  &51129& $\bullet$& $\bullet$& $\bullet$& $\bullet$&51249& $\bullet$& $\bullet$&  &  &51363& $\bullet$& $\bullet$&  &  \\
50049& $\bullet$&  &  &  &50214& $\bullet$& $\bullet$& $\bullet$&  &50345& $\bullet$& $\bullet$& $\bullet$& $\bullet$&50463& $\bullet$& $\bullet$&  &  &50626& $\bullet$&  &  &  &50798& $\bullet$&  &  &  &51033& $\bullet$& $\bullet$& $\bullet$& $\bullet$&51130& $\bullet$& $\bullet$&  &  &51250& $\bullet$& $\bullet$& $\bullet$& $\bullet$&51364& $\bullet$& $\bullet$&  &  \\
50050& $\bullet$& $\bullet$& $\bullet$&  &50215& $\bullet$&  &  &  &50346& $\bullet$& $\bullet$&  &  &50466& $\bullet$& $\bullet$&  &  &50627& $\bullet$&  &  &  &50799& $\bullet$& $\bullet$& $\bullet$&  &51034& $\bullet$& $\bullet$& $\bullet$& $\bullet$&51131& $\bullet$& $\bullet$&  &  &51251& $\bullet$& $\bullet$& $\bullet$& $\bullet$&51365& $\bullet$& $\bullet$&  &  \\
50051& $\bullet$& $\bullet$& $\bullet$&  &50217& $\bullet$& $\bullet$& $\bullet$&  &50347& $\bullet$& $\bullet$& $\bullet$& $\bullet$&50467& $\bullet$& $\bullet$&  &  &50628& $\bullet$&  &  &  &50800& $\bullet$&  &  &  &51035& $\bullet$& $\bullet$& $\bullet$& $\bullet$&51132& $\bullet$& $\bullet$&  &  &51252& $\bullet$& $\bullet$& $\bullet$& $\bullet$&51369& $\bullet$& $\bullet$&  &  \\
50052& $\bullet$&  &  &  &50232& $\bullet$& $\bullet$&  &  &50348& $\bullet$&  &  &  &50468& $\bullet$& $\bullet$&  &  &50642&  &  &  &  &50801& $\bullet$& $\bullet$& $\bullet$&  &51036& $\bullet$&  &  &  &51133& $\bullet$& $\bullet$& $\bullet$&  &51253& $\bullet$& $\bullet$& $\bullet$& $\bullet$&51370& $\bullet$&  &  &  \\
50053& $\bullet$& $\bullet$& $\bullet$&  &50233& $\bullet$& $\bullet$&  &  &50349& $\bullet$& $\bullet$&  &  &50469& $\bullet$& $\bullet$&  &  &50644&  &  &  &  &50803& $\bullet$& $\bullet$&  &  &51038& $\bullet$&  &  &  &51134& $\bullet$& $\bullet$& $\bullet$&  &51254& $\bullet$& $\bullet$& $\bullet$& $\bullet$&51373& $\bullet$& $\bullet$&  &  \\
50054& $\bullet$& $\bullet$& $\bullet$&  &50234& $\bullet$& $\bullet$& $\bullet$&  &50350& $\bullet$& $\bullet$& $\bullet$& $\bullet$&50470& $\bullet$& $\bullet$& $\bullet$&  &50647&  &  &  &  &50807& $\bullet$& $\bullet$& $\bullet$&  &51039& $\bullet$&  &  &  &51135& $\bullet$& $\bullet$& $\bullet$&  &51255& $\bullet$& $\bullet$& $\bullet$& $\bullet$&51461&  &  &  &  \\
50056& $\bullet$& $\bullet$& $\bullet$&  &50236& $\bullet$& $\bullet$&  &  &50351& $\bullet$& $\bullet$& $\bullet$& $\bullet$&50477& $\bullet$& $\bullet$&  &  &50648&  &  &  &  &50812& $\bullet$&  &  &  &51040& $\bullet$&  &  &  &51136& $\bullet$&  &  &  &51256& $\bullet$& $\bullet$& $\bullet$& $\bullet$&51463&  &  &  &  \\
50057& $\bullet$&  &  &  &50237& $\bullet$& $\bullet$&  &  &50352& $\bullet$& $\bullet$& $\bullet$& $\bullet$&50480& $\bullet$& $\bullet$&  &  &50649&  &  &  &  &50814& $\bullet$& $\bullet$&  &  &51041& $\bullet$&  &  &  &51137& $\bullet$& $\bullet$& $\bullet$&  &51257& $\bullet$& $\bullet$& $\bullet$& $\bullet$&51464&  &  &  &  \\
50059& $\bullet$&  &  &  &50239& $\bullet$& $\bullet$& $\bullet$&  &50353& $\bullet$&  &  &  &50481& $\bullet$&  &  &  &50654&  &  &  &  &50816& $\bullet$& $\bullet$& $\bullet$&  &51042& $\bullet$&  &  &  &51138& $\bullet$& $\bullet$& $\bullet$&  &51260& $\bullet$& $\bullet$& $\bullet$& $\bullet$&51465&  &  &  &  \\
50060& $\bullet$& $\bullet$&  &  &50240& $\bullet$& $\bullet$& $\bullet$&  &50354& $\bullet$&  &  &  &50482& $\bullet$& $\bullet$&  &  &50656&  &  &  &  &50817& $\bullet$& $\bullet$& $\bullet$&  &51044& $\bullet$&  &  &  &51139& $\bullet$& $\bullet$&  &  &51261& $\bullet$& $\bullet$& $\bullet$& $\bullet$&51473&  &  &  &  \\
50102& $\bullet$& $\bullet$& $\bullet$&  &50241& $\bullet$& $\bullet$& $\bullet$&  &50355& $\bullet$& $\bullet$& $\bullet$& $\bullet$&50483& $\bullet$& $\bullet$&  &  &50659&  &  &  &  &50818& $\bullet$& $\bullet$& $\bullet$&  &51045& $\bullet$&  &  &  &51140& $\bullet$& $\bullet$& $\bullet$&  &51262& $\bullet$& $\bullet$& $\bullet$& $\bullet$&51477&  &  &  &  \\
50103& $\bullet$&  &  &  &50243& $\bullet$& $\bullet$& $\bullet$&  &50356& $\bullet$&  &  &  &50485& $\bullet$& $\bullet$&  &  &50664&  &  &  &  &50820& $\bullet$&  &  &  &51046& $\bullet$&  &  &  &51141& $\bullet$& $\bullet$& $\bullet$&  &51264& $\bullet$&  &  &  &51479&  &  &  &  \\
50104& $\bullet$& $\bullet$& $\bullet$&  &50245& $\bullet$&  &  &  &50357& $\bullet$&  &  &  &50486& $\bullet$& $\bullet$& $\bullet$&  &50665&  &  &  &  &50821& $\bullet$& $\bullet$& $\bullet$&  &51047& $\bullet$&  &  &  &51142& $\bullet$& $\bullet$& $\bullet$&  &51265& $\bullet$& $\bullet$& $\bullet$& $\bullet$&51480&  &  &  &  \\
50105& $\bullet$& $\bullet$& $\bullet$&  &50247& $\bullet$& $\bullet$& $\bullet$&  &50358& $\bullet$& $\bullet$& $\bullet$& $\bullet$&50487& $\bullet$& $\bullet$&  &  &50669&  &  &  &  &50822& $\bullet$& $\bullet$& $\bullet$&  &51048& $\bullet$&  &  &  &51146& $\bullet$& $\bullet$&  &  &51266& $\bullet$& $\bullet$& $\bullet$& $\bullet$&51481&  &  &  &  \\
50106& $\bullet$& $\bullet$& $\bullet$&  &50248& $\bullet$& $\bullet$& $\bullet$&  &50359& $\bullet$& $\bullet$& $\bullet$& $\bullet$&50488& $\bullet$& $\bullet$&  &  &50682& $\bullet$& $\bullet$&  &  &50823& $\bullet$& $\bullet$& $\bullet$&  &51049& $\bullet$&  &  &  &51147& $\bullet$& $\bullet$&  &  &51267& $\bullet$&  &  &  &51482&  &  &  &  \\
50107& $\bullet$& $\bullet$&  &  &50249& $\bullet$& $\bullet$& $\bullet$&  &50360& $\bullet$& $\bullet$& $\bullet$& $\bullet$&50490& $\bullet$& $\bullet$&  &  &50683& $\bullet$& $\bullet$&  &  &50824& $\bullet$& $\bullet$& $\bullet$&  &51050& $\bullet$&  &  &  &51148& $\bullet$& $\bullet$&  &  &51268& $\bullet$& $\bullet$& $\bullet$& $\bullet$&51484&  &  &  &  \\
50109& $\bullet$& $\bullet$&  &  &50250& $\bullet$& $\bullet$& $\bullet$&  &50361& $\bullet$&  &  &  &50491& $\bullet$& $\bullet$&  &  &50685& $\bullet$& $\bullet$&  &  &50952& $\bullet$&  &  &  &51051& $\bullet$&  &  &  &51149& $\bullet$& $\bullet$&  &  &51269& $\bullet$& $\bullet$& $\bullet$& $\bullet$&51487&  &  &  &  \\
50111& $\bullet$& $\bullet$& $\bullet$&  &50251& $\bullet$& $\bullet$& $\bullet$&  &50362& $\bullet$& $\bullet$& $\bullet$& $\bullet$&50492& $\bullet$& $\bullet$&  &  &50686& $\bullet$& $\bullet$&  &  &50954& $\bullet$&  &  &  &51052& $\bullet$&  &  &  &51150& $\bullet$& $\bullet$&  &  &51271& $\bullet$& $\bullet$& $\bullet$& $\bullet$&51488&  &  &  &  \\
50112& $\bullet$& $\bullet$& $\bullet$&  &50252& $\bullet$& $\bullet$& $\bullet$&  &50363& $\bullet$& $\bullet$& $\bullet$& $\bullet$&50493& $\bullet$& $\bullet$&  &  &50687& $\bullet$& $\bullet$&  &  &50955& $\bullet$&  &  &  &51053& $\bullet$&  &  &  &51151& $\bullet$& $\bullet$&  &  &51272& $\bullet$& $\bullet$& $\bullet$& $\bullet$&51491&  &  &  &  \\
50113& $\bullet$&  &  &  &50253& $\bullet$& $\bullet$&  &  &50364& $\bullet$& $\bullet$& $\bullet$& $\bullet$&50494& $\bullet$&  &  &  &50688& $\bullet$& $\bullet$&  &  &50956& $\bullet$&  &  &  &51054& $\bullet$&  &  &  &51152& $\bullet$& $\bullet$&  &  &51273& $\bullet$& $\bullet$& $\bullet$& $\bullet$&51493&  &  &  &  \\
50114& $\bullet$&  &  &  &50254& $\bullet$& $\bullet$&  &  &50365& $\bullet$& $\bullet$& $\bullet$& $\bullet$&50496& $\bullet$& $\bullet$&  &  &50689& $\bullet$& $\bullet$&  &  &50957& $\bullet$&  &  &  &51055& $\bullet$&  &  &  &51153& $\bullet$& $\bullet$&  &  &51275& $\bullet$& $\bullet$& $\bullet$& $\bullet$&51556&  &  &  &  \\
50115& $\bullet$& $\bullet$&  &  &50255& $\bullet$& $\bullet$& $\bullet$&  &50366& $\bullet$& $\bullet$&  &  &50497& $\bullet$& $\bullet$&  &  &50690& $\bullet$& $\bullet$&  &  &50958& $\bullet$&  &  &  &51056& $\bullet$&  &  &  &51154& $\bullet$& $\bullet$&  &  &51276& $\bullet$& $\bullet$& $\bullet$& $\bullet$&51557&  &  &  &  \\
50116& $\bullet$& $\bullet$& $\bullet$&  &50257& $\bullet$& $\bullet$& $\bullet$&  &50367& $\bullet$& $\bullet$& $\bullet$& $\bullet$&50498& $\bullet$& $\bullet$&  &  &50691& $\bullet$&  &  &  &50959& $\bullet$&  &  &  &51057& $\bullet$&  &  &  &51155& $\bullet$& $\bullet$&  &  &51277& $\bullet$& $\bullet$& $\bullet$& $\bullet$&51558&  &  &  &  \\
50117& $\bullet$& $\bullet$&  &  &50259& $\bullet$& $\bullet$&  &  &50368& $\bullet$& $\bullet$& $\bullet$& $\bullet$&50499& $\bullet$& $\bullet$&  &  &50692& $\bullet$& $\bullet$&  &  &50960& $\bullet$&  &  &  &51058& $\bullet$&  &  &  &51156& $\bullet$& $\bullet$&  &  &51278& $\bullet$& $\bullet$& $\bullet$& $\bullet$&51559&  &  &  &  \\
50118& $\bullet$& $\bullet$& $\bullet$&  &50260& $\bullet$& $\bullet$&  &  &50369& $\bullet$&  &  &  &50500& $\bullet$& $\bullet$& $\bullet$&  &50693& $\bullet$& $\bullet$&  &  &50961& $\bullet$&  &  &  &51059& $\bullet$&  &  &  &51159& $\bullet$& $\bullet$& $\bullet$& $\bullet$&51279& $\bullet$&  &  &  &51560&  &  &  &  \\
50119& $\bullet$&  &  &  &50261& $\bullet$& $\bullet$&  &  &50370& $\bullet$& $\bullet$& $\bullet$& $\bullet$&50501& $\bullet$& $\bullet$& $\bullet$&  &50694& $\bullet$& $\bullet$&  &  &50962& $\bullet$&  &  &  &51060& $\bullet$&  &  &  &51161&  & $\bullet$& $\bullet$&  &51280& $\bullet$& $\bullet$& $\bullet$& $\bullet$&51562&  &  &  &  \\
50121& $\bullet$& $\bullet$&  &  &50262& $\bullet$& $\bullet$&  &  &50372& $\bullet$& $\bullet$& $\bullet$& $\bullet$&50502& $\bullet$& $\bullet$&  &  &50695& $\bullet$& $\bullet$&  &  &50964& $\bullet$& $\bullet$& $\bullet$& $\bullet$&51061& $\bullet$&  &  &  &51162&  &  &  &  &51281& $\bullet$& $\bullet$& $\bullet$& $\bullet$&51563&  &  &  &  \\
50123& $\bullet$&  &  &  &50263& $\bullet$& $\bullet$&  &  &50373& $\bullet$&  &  &  &50503& $\bullet$& $\bullet$&  &  &50696& $\bullet$&  &  &  &50965& $\bullet$& $\bullet$& $\bullet$& $\bullet$&51062& $\bullet$&  &  &  &51163&  & $\bullet$& $\bullet$&  &51291& $\bullet$& $\bullet$& $\bullet$& $\bullet$&51564&  &  &  &  \\
50124& $\bullet$& $\bullet$& $\bullet$&  &50264& $\bullet$& $\bullet$&  &  &50374& $\bullet$&  &  &  &50504& $\bullet$& $\bullet$& $\bullet$&  &50697& $\bullet$& $\bullet$&  &  &50966& $\bullet$& $\bullet$& $\bullet$& $\bullet$&51063& $\bullet$&  &  &  &51164&  &  &  &  &51292& $\bullet$&  &  &  &51565&  &  &  &  \\
50125& $\bullet$&  &  &  &50265& $\bullet$& $\bullet$& $\bullet$&  &50375& $\bullet$&  &  &  &50507& $\bullet$& $\bullet$&  &  &50698& $\bullet$& $\bullet$&  &  &50967& $\bullet$&  &  &  &51064& $\bullet$& $\bullet$&  &  &51168&  & $\bullet$& $\bullet$&  &51293& $\bullet$& $\bullet$& $\bullet$& $\bullet$&51566&  &  &  &  \\
50127& $\bullet$&  &  &  &50266& $\bullet$& $\bullet$& $\bullet$&  &50376& $\bullet$&  &  &  &50509& $\bullet$&  &  &  &50699& $\bullet$& $\bullet$&  &  &50968& $\bullet$& $\bullet$& $\bullet$& $\bullet$&51065& $\bullet$& $\bullet$& $\bullet$& $\bullet$&51169&  & $\bullet$& $\bullet$&  &51294& $\bullet$& $\bullet$& $\bullet$& $\bullet$&51567&  &  &  &  \\
50128& $\bullet$& $\bullet$& $\bullet$&  &50267& $\bullet$& $\bullet$& $\bullet$&  &50377& $\bullet$& $\bullet$& $\bullet$& $\bullet$&50510& $\bullet$& $\bullet$& $\bullet$&  &50700& $\bullet$& $\bullet$&  &  &50969& $\bullet$& $\bullet$&  &  &51066& $\bullet$& $\bullet$&  &  &51170&  &  &  &  &51295& $\bullet$& $\bullet$& $\bullet$& $\bullet$&51568&  &  &  &  \\
50129& $\bullet$& $\bullet$& $\bullet$&  &50268& $\bullet$& $\bullet$& $\bullet$&  &50379& $\bullet$&  &  &  &50514& $\bullet$& $\bullet$&  &  &50701& $\bullet$& $\bullet$& $\bullet$&  &50970& $\bullet$& $\bullet$&  &  &51067& $\bullet$& $\bullet$&  &  &51171&  &  &  &  &51297& $\bullet$& $\bullet$& $\bullet$& $\bullet$&51569&  &  &  &  \\
50130& $\bullet$& $\bullet$& $\bullet$&  &50269& $\bullet$& $\bullet$& $\bullet$&  &50380& $\bullet$&  &  &  &50515& $\bullet$& $\bullet$& $\bullet$&  &50702& $\bullet$& $\bullet$& $\bullet$&  &50972& $\bullet$& $\bullet$& $\bullet$& $\bullet$&51068& $\bullet$& $\bullet$&  &  &51173&  & $\bullet$&  &  &51298& $\bullet$& $\bullet$& $\bullet$& $\bullet$&51570&  &  &  &  \\
50131& $\bullet$&  &  &  &50270& $\bullet$& $\bullet$&  &  &50381& $\bullet$& $\bullet$& $\bullet$& $\bullet$&50516& $\bullet$& $\bullet$& $\bullet$&  &50703& $\bullet$& $\bullet$&  &  &50973& $\bullet$& $\bullet$& $\bullet$& $\bullet$&51069& $\bullet$& $\bullet$&  &  &51177&  & $\bullet$& $\bullet$&  &51299& $\bullet$& $\bullet$& $\bullet$& $\bullet$&51572&  &  &  &  \\
50132& $\bullet$& $\bullet$& $\bullet$&  &50271& $\bullet$& $\bullet$&  &  &50382& $\bullet$& $\bullet$&  &  &50518& $\bullet$&  &  &  &50704& $\bullet$& $\bullet$&  &  &50974& $\bullet$& $\bullet$& $\bullet$& $\bullet$&51070& $\bullet$&  &  &  &51178&  & $\bullet$& $\bullet$&  &51300& $\bullet$&  &  &  &51573&  &  &  &  \\
50134& $\bullet$& $\bullet$& $\bullet$&  &50272& $\bullet$& $\bullet$& $\bullet$& $\bullet$&50383& $\bullet$&  &  &  &50519& $\bullet$& $\bullet$& $\bullet$&  &50705& $\bullet$& $\bullet$&  &  &50976& $\bullet$& $\bullet$& $\bullet$& $\bullet$&51072& $\bullet$& $\bullet$& $\bullet$& $\bullet$&51179&  & $\bullet$& $\bullet$&  &51301& $\bullet$& $\bullet$& $\bullet$& $\bullet$&51574&  &  &  &  \\
50135& $\bullet$&  &  &  &50273& $\bullet$& $\bullet$& $\bullet$& $\bullet$&50385& $\bullet$& $\bullet$& $\bullet$& $\bullet$&50520& $\bullet$&  &  &  &50706& $\bullet$& $\bullet$&  &  &50977& $\bullet$&  &  &  &51073& $\bullet$& $\bullet$& $\bullet$& $\bullet$&51180&  &  &  &  &51302& $\bullet$& $\bullet$& $\bullet$& $\bullet$&51576&  &  &  &  \\
50142&  & $\bullet$& $\bullet$&  &50274& $\bullet$& $\bullet$& $\bullet$& $\bullet$&50386& $\bullet$& $\bullet$& $\bullet$& $\bullet$&50521& $\bullet$& $\bullet$&  &  &50707& $\bullet$& $\bullet$&  &  &50978& $\bullet$&  &  &  &51074& $\bullet$& $\bullet$& $\bullet$& $\bullet$&51181&  & $\bullet$& $\bullet$&  &51303& $\bullet$&  &  &  &51577&  &  &  &  \\
50143&  & $\bullet$& $\bullet$&  &50275& $\bullet$& $\bullet$& $\bullet$& $\bullet$&50387& $\bullet$& $\bullet$& $\bullet$& $\bullet$&50523& $\bullet$& $\bullet$& $\bullet$&  &50708& $\bullet$& $\bullet$&  &  &50979& $\bullet$& $\bullet$& $\bullet$& $\bullet$&51075& $\bullet$& $\bullet$& $\bullet$& $\bullet$&51182&  & $\bullet$& $\bullet$&  &51304& $\bullet$& $\bullet$& $\bullet$& $\bullet$&51578&  &  &  &  \\
50144&  & $\bullet$& $\bullet$&  &50276& $\bullet$&  &  &  &50388& $\bullet$& $\bullet$&  &  &50524& $\bullet$& $\bullet$& $\bullet$&  &50709& $\bullet$& $\bullet$&  &  &50981& $\bullet$& $\bullet$& $\bullet$& $\bullet$&51076& $\bullet$& $\bullet$& $\bullet$& $\bullet$&51183&  & $\bullet$&  &  &51305& $\bullet$&  &  &  &51579&  &  &  &  \\
50145&  & $\bullet$& $\bullet$&  &50278& $\bullet$&  &  &  &50390& $\bullet$& $\bullet$& $\bullet$& $\bullet$&50525& $\bullet$& $\bullet$&  &  &50711& $\bullet$& $\bullet$&  &  &50982& $\bullet$& $\bullet$& $\bullet$& $\bullet$&51077& $\bullet$& $\bullet$& $\bullet$& $\bullet$&51184&  &  &  &  &51306& $\bullet$& $\bullet$& $\bullet$& $\bullet$&51580&  &  &  &  \\
50146&  & $\bullet$&  &  &50282& $\bullet$& $\bullet$& $\bullet$& $\bullet$&50391& $\bullet$& $\bullet$& $\bullet$& $\bullet$&50526& $\bullet$& $\bullet$&  &  &50722& $\bullet$&  &  &  &50983& $\bullet$& $\bullet$& $\bullet$& $\bullet$&51078& $\bullet$& $\bullet$&  &  &51185&  & $\bullet$&  &  &51307& $\bullet$& $\bullet$& $\bullet$& $\bullet$&51582&  &  &  &  \\
50147&  & $\bullet$& $\bullet$&  &50284& $\bullet$&  &  &  &50397& $\bullet$& $\bullet$& $\bullet$& $\bullet$&50527& $\bullet$& $\bullet$&  &  &50723& $\bullet$&  &  &  &50984& $\bullet$& $\bullet$& $\bullet$& $\bullet$&51079& $\bullet$& $\bullet$&  &  &51187&  & $\bullet$&  &  &51308& $\bullet$& $\bullet$& $\bullet$& $\bullet$&51583&  &  &  &  \\
50148&  &  &  &  &50285& $\bullet$&  &  &  &50399& $\bullet$& $\bullet$& $\bullet$& $\bullet$&50528& $\bullet$& $\bullet$& $\bullet$&  &50724& $\bullet$& $\bullet$& $\bullet$&  &50985& $\bullet$& $\bullet$& $\bullet$& $\bullet$&51080& $\bullet$& $\bullet$&  &  &51188&  &  &  &  &51309& $\bullet$& $\bullet$& $\bullet$& $\bullet$&51584&  &  &  &  \\
50149&  & $\bullet$& $\bullet$&  &50287& $\bullet$& $\bullet$& $\bullet$& $\bullet$&50402& $\bullet$& $\bullet$& $\bullet$& $\bullet$&50529& $\bullet$& $\bullet$&  &  &50725& $\bullet$& $\bullet$& $\bullet$&  &50986& $\bullet$& $\bullet$&  &  &51081& $\bullet$& $\bullet$&  &  &51189&  & $\bullet$& $\bullet$&  &51311& $\bullet$& $\bullet$& $\bullet$& $\bullet$&51585&  &  &  &  \\
50150&  &  &  &  &50289& $\bullet$& $\bullet$& $\bullet$& $\bullet$&50403& $\bullet$& $\bullet$& $\bullet$& $\bullet$&50530& $\bullet$& $\bullet$&  &  &50726& $\bullet$& $\bullet$& $\bullet$&  &50987& $\bullet$& $\bullet$&  &  &51082& $\bullet$& $\bullet$&  &  &51192&  &  &  &  &51313& $\bullet$& $\bullet$& $\bullet$& $\bullet$&51606& $\bullet$&  &  &  \\
\end{tabular}

    \caption{\textbf{Subjects used in subsample \#1.} Membership to other
    subsamples is indicated by bullets.}
    \label{tab:subject_ids}
\end{table*}

\end{document}